# The Text Classification Pipeline: Starting Shallow, going Deeper

## From Foundations to GPT in Text Classification: A Comprehensive Survey on Current Approaches and Future Trends

**Marco Siino**
University of Catania
marco.siino@unict.it

**Ilenia Tinnirello**
University of Palermo
ilenia.tinnirello@unipa.it

**Marco La Cascia**
University of Palermo
marco.lacascia@unipa.it

**now**

the essence of knowledge

Boston — Delft

# Contents

# The Text Classification Pipeline: Starting Shallow, going Deeper

Marco Siino[1,2], Ilenia Tinnirello[2] and Marco La Cascia[2]

[1] *University of Catania, Catania, Italy; marco.siino@unict.it*
[2] *University of Palermo, Palermo, Italy; marco.siino@unipa.it,*
*ilenia.tinnirello@unipa.it, marco.lacascia@unipa.it*

ABSTRACT

Text classification stands as a cornerstone within the realm of Natural Language Processing (NLP), particularly when viewed through computer science and engineering. The past decade has seen deep learning revolutionize text classification, propelling advancements in text retrieval, categorization, information extraction, and summarization. The scholarly literature includes datasets, models, and evaluation criteria, with English being the predominant language of focus, despite studies involving Arabic, Chinese, Hindi, and others. The efficacy of text classification models relies heavily on their ability to capture intricate textual relationships and non-linear correlations, necessitating a comprehensive examination of the entire text classification pipeline.

In the NLP domain, a plethora of text representation techniques and model architectures have emerged, with Large Language Models (LLMs) and Generative Pre-trained Transformers (GPTs) at the forefront. These models are adept at transforming extensive textual data into meaningful vector representations encapsulating semantic information. The multidisciplinary nature of text classification, encompassing data mining, linguistics, and information retrieval, highlights the importance of collaborative research to advance the field. This work integrates traditional and contemporary text mining methodologies, fostering a holistic understanding of text classification.

This monograph provides an in-depth exploration of the text classification pipeline, with a particular emphasis on evaluating the impact of each component on the overall performance of text classification models. The pipeline includes state-of-the-art datasets, text preprocessing techniques, text representation methods, classification models, evaluation metrics, and future trends. Each chapter examines these stages, presenting technical innovations and recent findings. The work assesses various classification strategies, offering comparative analyses, examples and case studies. These contributions extend beyond a typical survey, providing a detailed and insightful exploration of the field.

# 1

## Introduction

In several Natural Language Processing (NLP) applications like news categorization, sentiment analysis, and subject labelling, text classification is a crucial and relevant task Garrido-Merchan *et al.*, 2023; Fields *et al.*, 2024b; Emanuel *et al.*, 2024. The goal is to tag or label textual components like sentences, questions, paragraphs, and documents. In this era of massive information dissemination, manually processing and categorizing huge amounts of text data takes a relevant quantity of effort and time. Text information can be found on social media, websites, chat rooms, emails, questions and answers from customer service representatives, insurance claims and user reviews. Furthermore, human factors such as skills and fatigue can influence the effectiveness of text classification by hand. It is preferable to automate the text classification pipeline involving machine learning models to get objective outcomes. Furthermore, to reduce the problem of information overloading, the improvement of information retrieval effectiveness can help in finding the necessary information for a certain task. Figure 1 illustrates a flowchart of the steps involved in text classification, under the light of traditional and most recent machine learning models. A critical first stage is the preprocessing of the text to be provided as input to the model. Classical approaches usually employ AI methods to collect relevant features, which are then classified using machine learning techniques. Next, the text representation approach can severely impact the outcomes, involving a series of transformations to map a source text to predicted labels. Deep learning, as opposed to traditional models, incorporates feature engineering into the training process. Up until 2010, classical text classification models were the most used and popular. Some of them are *logistic regressor*, *Naïve Bayes*, *Support Vector Machine* (SVM) and *K-Nearest Neighbour* (KNN). These methods can outperform past rule-based techniques in consistency and accuracy (Mitra *et al.*, 2007; Atmadja and Purwarianti, 2015). However, they still require feature engineering and are usually more time-consuming. Additionally, it is hard to understand the semantics of the words since they frequently neglect the context or natural sequential arrangement of textual material. In text classification, deep learning algorithms gradually replaced traditional techniques by the 2010s. Deep learning techniques for text mining automatically construct semantically pertinent representations without human intervention to define rules and features. Consequently, the majority of modern text classification activities are based on deep neural networks.

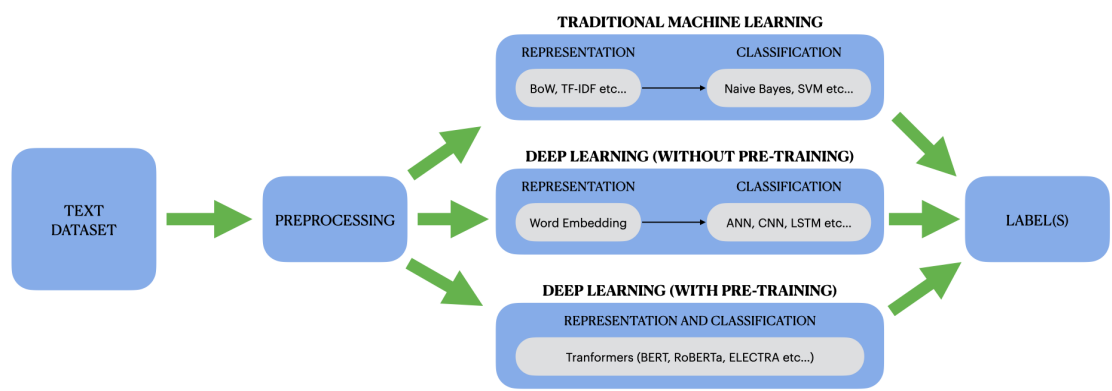

**Figure 1.1:** Overview of the text classification pipeline, illustrating the progression from text datasets to preprocessing, feature representations (e.g., Bag of Words, word embeddings), and final label predictions, encompassing traditional and modern approaches.

Most conventional machine learning models use a two-step procedure. First, the documents are stripped of manually added features

(or any other textual unit). In the following, a classifier receives these features to provide a prediction. The Bag of Words (BoW) feature and extensions are frequently created by hand. Hidden Markov Models, Naive Bayes, SVM, Random Forests and Gradient Boosting, are common classification algorithms employed in the second step. Numerous disadvantages exist with this two-step approach. For instance, using handcrafted features and expecting acceptable performance requires time-consuming feature engineering and analysis. Due to the strategy's heavy reliance on domain expertise for feature generation, it is difficult to adapt it to new applications. Last, because of the specific features domain, these models cannot fully benefit from the vast volumes of training data available. To address the issues related to handcrafted features, the use of neural approaches has increased. The main component of these approaches is an embedding space, where text is encoded as a low-dimensional continuous feature vector without the need for traditional feature representation strategies. The *Latent Semantic Analysis* (LSA) proposed in Landauer and Dumais, 1997 is one of the earliest studies on embedding models. The proposed architecture is trained on 200K words and has fewer than 1 million parameters. In Bengio *et al.*, 2000, the first neural language model was proposed. It consisted of an artificial neural network trained on over 10 million words. When progressively larger embedding models were constructed with significantly more training data, a paradigm change occurred. Several *Word2Vec* models that Google created in 2013 (Mikolov *et al.*, 2013b) were trained using billions of words and quickly gained popularity for numerous NLP applications. As the basis for their contextual embedding model, the researchers from Ai2[1] and the University of Washington created a Bidirectional-Long Short-Term Memory (BiLSTM) network using 93 million hyperparameters and a training performed on billions of words in 2017. A novel model named Embedding from Language Models (ELMo) (Peters *et al.*, 2018) captures contextual information and performs significantly better than Word2Vec. This subsequent development results in the construction of embedding models using Google's new neural architecture, the *Transformer* (Vaswani *et al.*, 2017). The Transformer architecture is based on attention modules, which boosts the effectiveness of extensive model training on the Tensor Processing Unit (TPU). In the same year, Google created the Bidirectional Encoder Representations from Transformers (BERT) (Devlin *et al.*, 2019). BERT has 340M parameters and was trained on 3.3 billion words. More training data and larger models are proposed in the literature every day. The most recent OpenAI GPT model has more than 170 billion parameters Dale, 2021 and it is based on Transformers. Some academics contend that despite the enormous models' remarkable performance on different NLP tasks, they do not truly grasp language and are insufficient for many domains that are mission-critical (Jin *et al.*, 2020; Marcus and Davis, 2019). Recently, there has been a rise of interest toward neuro-symbolic hybrid models to solve significant flaws of neural models like interpretability, inability to use symbolic thinking and lack of grounding (Schlag *et al.*, 2019; Gao *et al.*, 2020).

---

[1] https://allenai.org/allennlp/software/elmo

Although there are many excellent reviews and textbooks on text classification techniques and applications, this work provides a thorough analysis of all the phases that go into creating a text classification pipeline with several contributions, including traditional and deep models to explore the impact on the performance of each stage of the pipeline. Even if specific languages are considered in the related works, from the standpoint of computer science, English is the language that is most frequently used and referred to in the present literature regarding text classification. Furthermore, most of the Large Language Models (LLMs) and pre-trained word embeddings are originally developed focusing on English, partially neglecting the other languages. Nowadays, modern LLMs are multilingual so they can be fed and can produce output also in other languages other than English (Rathje *et al.*, 2024). The rest of this work primarily uses English as the reference language for many of the examples and cases presented and discussed.

Starting with a discussion on some of the more contemporary tasks — such as author profiling, topic classification, news classification, and sentiment analysis — we then present classification models and the most recent and relevant findings. We also cover the most recent deep neural network architectures, which are divided into several types based on their functioning, including Transformers (LLMs and GPTs), Convolutional

Neural Networks (CNNs), Capsule Nets and Recurrent Neural Networks (RNNs).

This monograph is organized as follows: Chapter 2 presents the most common datasets used and available in the literature. In Chapter 3, the preprocessing technique to prepare raw text are presented and discussed. In Chapter 4 the methods to represent text in a numerical way understandable by a computer are reported. In this chapter we also show and analyse a word embedding space trained from scratch. In Chapter 5, traditional and modern classifiers commonly employed for text classification are discussed, including a discussion on modern LLMs and GPTs. In Chapter 6 generic and linguistic-specific metrics to evaluate the performance on text classification tasks are discussed. In Chapter 7 the conclusions and the future perspectives are presented. The contributions and a summary for each chapter of this work are reported in what follows.

## 1.1 Overview and contributions

Several works have investigated text classification techniques from a general standpoint. We specifically mention the work in Li *et al.*, 2020, which offers a thorough analysis of model architectures, from traditional to modern deep learning-based ones. The survey by Kowsari *et al.*, 2019 offers a great examination of preprocessing procedures, including feature extraction and dimensionality reduction. Despite including quantitative outcomes of conventional approaches, Minaee *et al.*, 2021 mainly focuses on deep learning models. By providing a view of each stage required to design a text classification model, this monograph seeks to enhance the landscape of text classification from a general point of view. As a result, we give a thorough explanation of the key data preparation procedures used along with classification models. We provide model descriptions from traditional to deep learning-based ones, in contrast to prior surveys. The design of the classifier and feature extraction are highlighted for the traditional models. A specific overview of each chapter of this work is reported to conclude this section.

**Overview of Chapter 2: Challenges and datasets**

In the early history of machine learning, information retrieval systems primarily used text classification algorithms. But as technology has developed over time, text classification and document categorization have become widely employed in several fields, including law, engineering, social sciences, healthcare, psychology, and medicine. We highlight some domains that use text classification algorithms in this section. Some text classification tasks are discussed in this chapter, including three new datasets related to emerging author profiling tasks. The datasets available in the literature and related to these tasks and usually employed as benchmark, are also reported and presented in this chapter.

**Overview of Chapter 3: Text preprocessing**

In this chapter we collect, report and discuss the text preprocessing techniques found in the literature and their possible and most recent variants, proposing a standard nomenclature based on acronyms. We also provide the reader with useful information for self-study of the techniques presented along with advice on how to operate educated choices to select the preprocessing technique (or combination of techniques) given a specific task, model, and dataset. According to recent related works, we also discuss if simple classifiers' performance is comparable to the ones obtained by Transformer-based models when text preprocessing is performed according to the specific model and dataset used.

**Overview of Chapter 4: Text representation**

Before moving to the classification stage, it is necessary to convert unstructured data, especially free-running text data, into organized numerical data. To do this, a document representation model must be used to employ a subsequent classification system following the text preprocessing stage. Text representation models convert text data into a numerical vector space, which has a substantial impact on how well subsequent learning tasks can perform. In the history of NLP, word representation has always been a topic of interest. It is crucial to properly represent such text data since it contains a wealth of information and may be applied broadly across a variety of applications. This chapter examines the expressive potential of several word representation models, ranging from the traditional to the contemporary word representation approaches provided by LLMs. The chapter discusses numerous representation methods that are frequently employed in the literature. Before discussing well-known representation learning and pre-trained language models, we first discuss various statistical models. Then we move to attention-based representation and, in the last section of this chapter, to a case study about the analysis of a trained word embedding for a specific text classification task. Thanks to a Principal Component Analysis (PCA) tool, it shows and discusses the effect of CNN training on a 3D visualization of a word embedding space. In this way we can motivate some implicit choices operated during the training of a deep learning model, to assign specific word vectors to certain keywords belonging to one of the two class labels used for the discussed task.

**Overview of Chapter 5: Text classification models**

In Chapter 5 are reported both the traditional classification models for text classification and the most modern ones based on deep learning. Models discussed in this chapter belong to three different groups. The non-deep learning deterministic models, the foundational deep learning models and the large pre-trained language models known as Transformers. The term "earlier approaches" refers to all techniques used before the advent of deep neural networks, when the prediction was based on manually created features. Neural networks with only a few hidden layers are also included in this category, and these are so-called "shallow" networks. These methods replace several rule-based ones, which they usually outperform in terms of accuracy. The most recent deep learning models, which have an impact on all artificial intelligence domains, including text classification, are also discussed. These techniques have become popular because they can simulate intricate features without requiring manual engineering, which reduces the need for subject expertise. Finally, we discuss Transformers (LLMs and GPTs) and the recent and emerging discipline of *Prompt Engineering*. We discuss several prompting techniques, and then we move to some ethical considerations on the use of generative AI.

## Overview of Chapter 6: Evaluation metrics

This chapter focuses on how to evaluate the performance of deep learning models in the context of text classification tasks, introducing the most used metrics in the literature. We discuss various metrics such as accuracy, precision, recall, and F1 score, emphasizing the importance of selecting the right metric based on the specific goals. In addition, we explore the limitations of traditional evaluation metrics and highlight the necessity for more sophisticated approaches, particularly in scenarios involving imbalanced datasets. The use of confusion matrices and *ROC-AUC* scores were recommended to provide a more comprehensive evaluation of model performance, along with metrics as *ROUGE* and *BLEU* for tasks involving text generation and summarization. Moreover, we propose the integration of human evaluation methods to supplement quantitative metrics, recognizing that the nuances of language often elude numerical representation.

## Overview of Chapter 7: Conclusions and future perspectives

In the last chapter of this work, we report the final conclusions and future perspectives on the matter.

# 2

---

# Tasks and datasets

---

The process of organizing texts, such as tweets, news articles, and customer reviews into distinct categories can be broadly considered a form of text classification. Common tasks include topic classification, news categorization, and sentiment analysis. Recent research has shown that by enabling text classifiers to process pairs of texts as inputs, various natural language understanding tasks—such as natural language inference and extractive question answering—can be effectively framed as text classification problems. However, these tasks often do not operate within a finite and predefined set of labels, making them less typical of traditional text classification. The initial section of this chapter introduces several popular text classification tasks from the literature.

The availability of labelled datasets has been a significant driver in the rapid advancement of the field. The datasets presented in this chapter are frequently utilized as benchmarks in related research. In this introduction, we list the domain-specific properties of these datasets and provide an overview in Table 2.1 that lists the task description, the overall sample count, the number of target classes, and articles presenting the corresponding dataset.

The text classification tasks presented here are:

- Author profiling

- Topic classification

- News classification

- Sentiment analysis

Text classification serves as a foundational framework for various NLP tasks by mapping textual inputs to predefined categories. The following tasks can be reformulated as text classification problems, where textual inputs are mapped to predefined categories. Named Entity Recognition (NER) can be framed as a sequence classification task, where each token is assigned a label corresponding to entity types (Lample *et al.*, 2016). Similarly, co-reference resolution can be approached as a classification problem, determining whether two mentions refer to the same entity (Lee *et al.*, 2017). Relation extraction is often modelled as a multi-class classification task, where predefined relationship labels are assigned to entity pairs within a text (Zeng *et al.*, 2014). Paraphrase identification, which assesses whether two sentences express the same meaning, can also be formulated as a binary classification problem (Dolan and Brockett, 2005). Additionally, textual entailment—determining whether one sentence logically follows from another—can be treated as a binary or multi-class classification task (Bowman *et al.*, 2015).

## 2.1 Research areas

### 2.1.1 Author profiling

One of the three main areas of automatic authorship identification, alongside authorship attribution and authorship verification, is author profiling. The development of this field began to take shape at the turn of the 20th century. Initially, the approach was applied to the writings of Francis Bacon, William Shakespeare, and Christopher Marlowe by an American self-taught physicist and meteorologist (i.e., Thomas Corwin Mendenhall). Mendenhall analysed the word lengths of these authors to identify quantitative stylistic variations.

**Table 2.1:** Dataset characterization and stats.

| Dataset | Task | #Total documents | #Number of classes | Reference |
|---------|------|------------------|--------------------|-----------|
| FNS | Author profiling | 500 | 2 | Pardo *et al.*, 2020 |
| HSS | Author profiling | 600 | 2 | Rangel *et al.*, 2021 |
| ISS | Author profiling | 600 | 2 | Bevendorff *et al.*, 2022 |
| MR | Sentiment analysis | 10,662 | 2 | Pang *et al.*, 2002 |
| SST1 | Sentiment analysis | 11,855 | 5 | Socher *et al.*, 2013 |
| SST2 | Sentiment analysis | 9,613 | 2 | Socher *et al.*, 2013 |
| MPQA | Sentiment analysis | 10,606 | 2 | Deng and Wiebe, 2015 |
| IMDB | Sentiment analysis | 50,000 | 2 | Maas *et al.*, 2011 |
| Yelp2 | Sentiment analysis | 290,000 | 2 | Zhang *et al.*, 2015 |
| Yelp5 | Sentiment analysis | 700,000 | 5 | Zhang *et al.*, 2015 |
| Amazon2 | Sentiment analysis | 4,000,000 | 2 | Zhang *et al.*, 2015 |
| Amazon5 | Sentiment analysis | 3,650,000 | 5 | Zhang *et al.*, 2015 |
| Google News | News classification | 190,000 | 2 | Das *et al.*, 2007 |
| Reuters news | News classification | 10,788 | 90 | URL[1] |
| 20NG | News classification | 376,420 | 20 | URL[2] |
| AG News | News classification | 127,600 | 4 | URL[3] |
| Sogou | News classification | 2,909,551 | 5 | URL[4] |
| PCL | Topic classification | 10,637 | 2 | Pérez-Almendros *et al.*, 2022 |
| DBpedia | Topic classification | 630,000 | 14 | Lehmann *et al.*, 2015 |
| Ohsumed | Topic classification | 7,400 | 23 | URL[5] |
| ISTO | Topic classification | 44,898 | 2 | URL[6] |
| EUR-Lex | Topic classification | 19,314 | 3,956 | Loza Mencía and Fürnkranz, 2008 |
| Yahoo! | Topic classification | 1,460,000 | 10 | Zhang *et al.*, 2015 |
| WOS | Topic classification | 46,985 | 134 | Kowsari *et al.*, 2017 |

Author profiling involves the analysis of a corpus of texts to determine the author's identity or to identify distinct traits of the author based on stylistic and content-based factors. Commonly analysed factors include age and gender, but recent research has also explored additional aspects such as personality traits and occupation (Wiegmann *et al.*, 2020). Author profiling is valuable in various sectors, particularly forensics and marketing, where identifying specific traits of a text's author is crucial. The task of author profiling can vary depending on the application, the traits to be identified, the number of authors studied, and the volume of texts available for analysis. While traditionally focused on written works such as literary texts, the scope has expanded to include online texts with the advent of computers and the Internet.

Despite significant advancements in the 21st century, author profiling remains a challenging and not fully resolved process. Below are some well-known author profiling datasets that have been featured in recent literature.

- **Fake News Spreaders (FNS)**. The FNS dataset is presented

and discussed in Pardo *et al.*, 2020 and available under request[7]. The dataset was used for the international shared task at PAN[8]. The organizers of the task aim to determine whether it is feasible to differentiate between authors who have previously disseminated fake news and those who have not. The dataset comprises tweets in both Spanish and English. Each author in the dataset is represented by one hundred tweets, and a corresponding class label indicating whether the author has shared fake news in the past (labelled as 1) or not (labelled as 0). The training set includes 150 authors per label, while the test set includes 100 authors per label. In total, the dataset consists of 500 authors, amounting to 50,000 tweets. The results of the participants in the Fake News Spreader (FNS) challenge are publicly available[9].

- **Hate Speech Spreaders (HSS)**.The HSS dataset is presented and discussed in Rangel *et al.*, 2021. As an initial step in curbing the spread of hate speech among online users, the task's organizers aim to identify potential Twitter users who disseminate hate speech. The dataset includes tweets in both Spanish and English. Each author in the dataset is represented by two hundred tweets with a corresponding class label indicating whether the author has shared hate speech in the past (labelled as 1) or not (labelled as 0). For each language, the training set includes 100 authors per class, while the test set includes 50 authors per class. In total, the dataset comprises 600 authors, amounting to 120,000 tweets. The results of the participants in the Hate Speech Spreader (HSS) task are publicly available[10].

- **Irony and Stereotype Spreaders (ISS)**.The ISS dataset is presented and discussed in Bueno *et al.*, 2022; Bevendorff *et al.*, 2022 and available under request[11]. The dataset was used for

---

the international shared task at PAN[12]. The task's organizers want to focus on irony. Especially when words are used subtly and figuratively to indicate the opposite of what is expressed. A more violent version of irony, sarcasm aims to mock or ridicule a target without necessarily restricting the possibility of hurting it. The objective is to profile users whose tweets can be labelled as sarcastic. A group of 600 Twitter authors make up the dataset that the PAN organizers have created. Two hundred tweets are provided for each author. Each author is represented by a unique XML file with 200 tweets. Four hundred and twenty authors made up the organizers' labelled train set. In the test set, there are 180 further authors. The train set's authors are identified by the letters "I" (ISS) or "NI" (nISS). The results of the participants in the task are available online[13].

### 2.1.2 Topic classification

Topic classification, often referred to as *topic analysis*, aims to identify the main theme or themes of a text (for example, determining whether a product review pertains to "ease of use" or "customer assistance"). In topic analysis, the intricate textual theme is defined to ascertain the text's meaning. A crucial aspect of this method is topic labelling, which involves assigning themes to documents to streamline the topic analysis process. Below, we list several state-of-the-art datasets used in this domain.

- **Patronizing and Condescending Language (PCL)**. Described in detail in Pérez-Almendros *et al.*, 2022, the dataset originates from the detecting PCL task hosted at SemEval-2022. The task is an emerging one about detecting PCL (Pérez-Almendros *et al.*, 2020). PCL occurs when language implies superiority over others, talks down to them, or portrays them or their circumstances in a kind but belittling manner, often evoking feelings of pity or compassion. PCL is typically involuntary and unconscious, often stemming from good intentions. A classifier must ascertain whether PCL is present in a given text to fulfil the task. The dataset is available on GitHub[14].

---

[12]https://pan.webis.de
[13]https://pan.webis.de/clef22/pan22-web/author-profiling.html

- **DBpedia**. Wikipedia's most frequently used info boxes were used to create the DBpedia (Lehmann *et al.*, 2015), a sizable multilingual knowledge library. Every month, it is released a new edition of DBpedia which adds or removes classes and attributes. The most widely used version of DBpedia comprises 14 classes, 560,000 and 70,000 records, for training and testing respectively.

- **Ohsumed**. The Ohsumed[15] has a MEDLINE database affiliation. There are 23 categories for cardiovascular diseases and 7,400 texts overall. All texts are classified into one or more classes and are abstracts of medical information.

- **ISTO Fake News**. The dataset[16] contains two types of articles: fake and real news. This dataset is collected from real-world sources; the truthful articles were obtained by crawling articles from *Reuters.com*. As for the fake news articles, they were collected from different sources. The fake news articles were collected from unreliable websites and flagged by Politifact (a fact-checking organization in the USA) and Wikipedia. The dataset contains different types of articles on different topics, however, the majority of articles focus on political and World news topics.

- **EUR-Lex**. The EUR-Lex dataset (Loza Mencía and Fürnkranz, 2008) consists of several document categories that are indexed by orthogonal categorization systems to enable a variety of search functions. With 19,314 documents and 3,956 categories, the most widely used variant of the dataset is based on various parts of EU laws.

- **Yahoo! Answer**. The Yahoo! Answer[17] dataset (Zhang *et al.*,

---

2015) concerns topic labelling with 10 different classes. There are 6,000 and 140,000 samples to test and train respectively. Three components, referred to as question titles, question contexts, and best responses, are included in every sentence.

- **Web Of Science (WoS)**. The WoS dataset Kowsari *et al.*, 2017 is a set of information and meta-information about articles and is available via WoS, the most reputable global citation database, regardless of the publisher. There are three variants of WOS: WOS-46985, WOS-11967, and WOS-5736. The full dataset name is WOS-46985. WOS-46985 has two subsets: WOS-11967 and WOS-5736. The WOS-46985 dataset consists of research papers categorized into multiple scientific disciplines. Its two subsets, WOS-11967 and WOS-5736, focus on different aspects of text classification. WOS-11967 includes papers from broader scientific domains, covering research topics across natural sciences, engineering, and social sciences. This subset is often used for multi-class text classification tasks. WOS-5736 contains a more specialized set of papers, typically focusing on hierarchical classification, where documents are assigned to categories at different levels of granularity within a structured taxonomy. These subsets help researchers study various classification challenges, including multi-class and hierarchical text classification in scientific literature.

### 2.1.3 News classification

News classification involves the automated categorization of news articles into predefined tags based on their content, with the model's accuracy derived from training on labelled news records. News items can be categorized into various domains such as business, entertainment, politics, sports, technology, and more. News classification systems help users efficiently find articles of interest, saving time and reducing information overload. The task of categorizing news items by topic or user interest is crucial. By leveraging user preferences, identifying emerging news topics, or recommending relevant material, a news classification model assists individuals in obtaining real-time information tailored to their needs. Here, we delve into the details of several commonly used datasets in this domain.

- **Google News**. The Google News dataset presented in Das *et al.*, [2007](#) is made up of two datasets. The first consists of a subset of clicks received on the Google News website over a certain period, from the top 5,000 users (top sorted by the number of clicks). There are about 40,000 unique items that are part of this dataset and about 370,000 clicks. The second dataset is similar to the previous one (in fact a superset) and just contains more records: 500,000 users, 190,000 unique items and about 10,000,000 clicks. In order to have uniformity in comparisons, authors binarized the first dataset as follows: if the rating for an item, by a user, is larger than the average rating by this user (average computed over her set of ratings) they assign it a binary rating of 1, 0 otherwise.

- **Reuters news**. The Reuters-21578 dataset[18] is often used for text categorization. It was gathered by the Reuters Economic press release service in 1987. A version of Reuters-21578 with multiple classes containing 10,788 documents is named *ModApte*. A total of 90 lessons, 7,769 training samples, and 3,019 test samples are included. R8, R52, RCV1, and RCV1-v2 are additional datasets generated from a portion of the Reuters dataset.

- **20 Newsgroup (20NG)**. The 20NG dataset[19] consists of newsgroup documents that were posted on 20 various themes. For text categorization, text clustering, and other tasks, different variations of this dataset are employed. One of the most often used versions has 18,821 papers, evenly distributed among all topics.

- **AG News**. The AG News dataset[20] consists of news articles compiled by academic news search engine *ComeToMyHead* from more than 2,000 news sources. It takes advantage of each news story's title and description fields. A total of 120,000 training texts and 7,600 test texts are included in AG. Each sample consists of a brief sentence that has a four-class label.

---

[18]https://martin-thoma.com/nlp-reuters
[19]http://qwone.com/~jason/20Newsgroups/
[20]http://groups.di.unipi.it/~gulli/AG_corpus_of_news_articles.html

- **Sogou**. The *SogouCS* and *SogouCA* news sets are included in the *Sogou*[21] dataset, which combines both of them. The name of the domains within the URL serves as the labels for each text. So, as the classification labels for the news, the domain names in their URLs are used. For illustration, the news at http://sports.sohu.com is classed under the sports category.

### 2.1.4 Sentiment analysis

Sentiment analysis, often referred to as *opinion mining* or *emotion AI*, involves the systematic identification, extraction, quantification, and study of affective states and subjective information using NLP, text analysis, computational linguistics, and biometrics. This technique is widely applied in marketing, customer service, and clinical medical settings. It is employed to analyse voice of the customer materials, including reviews and survey responses, as well as content from the internet and social media, and healthcare documents.

This category of tasks involves identifying the polarity and perspective of users' opinions in text, such as tweets, movie reviews, or product reviews. Unlike traditional text classification, which focuses on the objective content of the text, sentiment analysis aims to determine whether the text supports a particular viewpoint. It may also involve understanding the emotional states and subjective information conveyed in the text, often categorized by the emotions evoked. The task can be modelled as a binary problem, classifying texts into negative and positive categories, or a multi-label task, grouping texts into multiple sentiment labels. Here, we present details of some of the most commonly used datasets in the literature, which serve as benchmarks for sentiment analysis.

- **Movie Review (MR)**. The MR dataset (Pang *et al.*, 2002) is a set of film reviews that was created with the goal of identifying the

---

[21]https://huggingface.co/DSs/sogou_news/blob/main/README.md

sentiment attached to each user review and deciding whether it is positive or negative. There is a sentence for each review. There are 5,331 positive samples and 5,331 negative samples in the corpus.

- **Stanford Sentiment Treebank (SST)**. The SST dataset (Socher *et al.*, 2013) extends MR. It has two categories: one with binary labels and the other with fine-grained (five-class) labels. Namely, SST-1 and SST-2, respectively. There are 8,544/1,101/2,210 samples, in the train/dev/test set respectively for a total of 11,855 movie reviews in SST-1. SST-2 is divided into train, dev and test sets, with respective sizes of 6,920, 872, and 1,821.

- **Multi-Perspective Question Answering (MPQA)**. The MPQA is an opinion dataset (Deng and Wiebe, 2015). It also has two class labels and an MPQA dataset of opinion polarity detecting sub-tasks. In total, 10,606 phrases from news stories from various news sources are included in MPQA. There are 7,293 negative texts and 3,311 positive texts, all without text labels.

- **Internet Movie Database (IMDB)**. A dataset for binary sentiment classification is first described in Maas *et al.*, 2011 as the IMDB dataset. It comprises 25,000 reviews of highly divisive movies for testing and 25,000 for training. Additional unlabelled data is also available for use. The collection includes binary sentiment polarity labels for the movie reviews that go along with them. The total of 50,000 reviews are divided into 25,000 reviews each for training and testing, and make up the core dataset. The reviews are balanced for the two classes (i.e., 25,000 are positives and 25,000 are negatives). For unsupervised learning, an additional 50,000 unlabelled documents are included. The IMDB dataset is available online[22].

- **Yelp**. The Yelp reviews dataset (Zhang *et al.*, 2015) comes from the 2015 Yelp dataset Challenge. 1,569,264 of the samples in this dataset include review texts. From this dataset, two classification tasks are created: one predicts the total amount of stars that a buyer has provided, and the other predicts whether a star's polarity is positive or negative. The first dataset has 650,000 and 50,000 samples for train and test respectively, and 280,000 training samples and 10,000 test samples for each polarity in the polarity dataset.

---

[22]https://ai.stanford.edu/~amaas/data/sentiment/

- **Amazon**. A well-known corpus known as the Amazon dataset was created by gathering product reviews from the Amazon website (Zhang *et al.*, 2015). There are two categories in this dataset. There are 3,600,000 and 400,000 samples in the train and in the test sets in the Amazon-2 with two labels. For training and testing purposes, Amazon-5, which has five classes, has 3,000,000 and 650,000 comments.

## 2.2 Conclusion

In this chapter, we have examined some of the most relevant datasets in the field of text classification across various domains, emphasizing their role in advancing research on document categorization and emerging classification tasks such as author profiling. We have also provided an overview of widely adopted benchmark datasets that serve as critical resources for evaluating text classification approaches.

It is worth mentioning that Siino *et al.*, 2022a have analysed linguistic corpora and datasets to identify key properties that may improve text classification performance, particularly in tasks such as fake news spreader detection. Additionally, several studies have explored data augmentation techniques to enhance dataset quality and expand training samples. A notable example is in Siino *et al.*, 2024b, which increases dataset size by generating backtranslated versions in multiple languages beyond English, thereby enriching linguistic diversity and potentially improving classification robustness.

These studies underscore the importance of high-quality datasets and augmentation strategies in supporting advancements in text classification research, facilitating the development of more effective classification methodologies across various application areas.

# 3

## Preprocessing

Tasks related to NLP, typically involve lexical tokenization, preprocessing, probabilistic tokenization, and classification stages. The preprocessing step includes operations such as lowercasing, stemming, lemmatization, stop word removal, and other techniques discussed in this chapter. Here, we use the term *preprocessing* to refer to any modifications made to the input text after lexical tokenization and before probabilistic tokenization.

Specifically, preprocessing can involve deleting unnecessary content for certain tasks (e.g., removing stop words and non-alphabetic characters), merging semantically similar words to enhance prediction accuracy and reduce data sparsity (using stemming, lemmatization, character casing conversion, expanding abbreviations, correcting misspellings), and increasing the amount of semantic information available (e.g., *Part of Speech* tagging, managing negation words). However, preprocessing can also inadvertently delete important data (such as relevant stop words) or introduce errors (e.g., conflating semantically distinct words through stemming, which can alter the outcomes of a classification model). In this chapter, preprocessing involves transforming the text before determining which text units to use as tokens during the probabilistic tokenization stage.

Despite its importance, the text preprocessing stage is often overlooked in many text mining studies. However, unstructured texts available on the internet contain a substantial amount of noise. In some cases, the noise level can be so high that it misleads machine learning algorithms. Noise can be caused by users frequently using slang, acronyms, and making spelling and grammar mistakes. Users may also overuse punctuation marks to emphasize emotions, such as typing multiple exclamation marks instead of a single one. In this context, noise refers to any useless information that remains after preprocessing a dataset, which can affect subsequent text-based tasks. As discussed in Siino *et al.*, 2024c, an incorrect choice during text preprocessing can lead to a significant difference in classification performance, potentially reducing accuracy by over 25% using the same model and dataset.

Preprocessing can be summarized as the process of cleaning and preparing texts for subsequent operations. Effective data cleaning and normalization are crucial because the performance of models employed after preprocessing depends significantly on the quality of the data. The role of preprocessing before and during feature selection is of prominent importance, although past research has provided conflicting recommendations due to variations in datasets, techniques, and models evaluated.

There is no standard convention for preprocessing in the literature, with each study testing different techniques. This chapter reports and discusses various preprocessing techniques and the results available in the literature. The aim is to improve the text preparation stage, resolve inconsistencies in preprocessing advice, and offer guidelines for future studies. We investigate how preprocessing choices affect performance using both deep (pre-trained or not) and non-deep learning models. A well-designed preprocessing stage can remove noise, highlight important features, and reduce the time required for training and testing a model. It is essential to make an educated and context-dependent choice about which preprocessing methods (or combinations) to employ and in what order.

In this chapter - partly based on one of our previous studies (Siino *et al.*, 2024d) - we collect, report, and discuss text preprocessing techniques found in the literature, including their recent variants, and propose a uniform nomenclature based on acronyms. We provide useful information for self-study and in-depth understanding of these techniques, offering advice on making educated choices for selecting preprocessing techniques given a specific task, model, and dataset.

We discuss how text preprocessing affects the performance of modern pre-trained architectures based on attention (i.e., Transformers) and determine if the performance of simple classifiers is comparable to that of Transformer-based models when text preprocessing is tailored to the specific model and/or dataset.

This chapter on text preprocessing is structured as follows: the next two sections discuss the gaps in the literature and related work on the impact of preprocessing techniques. Section 3.3 provides a complete discussion of the collected preprocessing techniques.

## 3.1   Gaps in the literature

In this subsection, we briefly introduce some of the most referenced and comprehensive surveys reported in the literature on text preprocessing. A more detailed discussion, including the most recent and relevant studies, is provided in the section dedicated to related work. We conclude this subsection by highlighting the gaps found in the literature.

In Singh and Kumari, 2016, the authors examine the effects of pre-processing on Twitter data, emphasizing the significant improvement in classifier performance. They removed URLs, user mentions, stop words, hashtags, and punctuation, and then used n-grams to replace slang words with their standard equivalents. This preprocessing method links slang to existing words to better understand their meaning and sentiment. The authors used an SVM classifier and concluded by questioning how effectively the proposed system would work with different classifiers on other types of text.

The authors in Symeonidis *et al.*, 2018 studied how various pre-processing techniques affect model performance using four traditional classifiers and a neural network. They represented words using only TF-IDF (unigram). The study found that while removing punctuation does not enhance classification performance, other preprocessing steps like removing digits, expanding contractions to base words, and lemmatization do. Additionally, the study showed how different preprocessing strategies interact and identified those that work best when combined. However, the authors suggested that future studies could test these preprocessing techniques on datasets from different domains, such as news articles and product or movie reviews.

In Naseem *et al.*, 2021, the authors analysed twelve different preprocessing techniques on three Twitter datasets focused on hate speech detection. They observed the impact of these techniques on the classification tasks. However, they did not explore all possible combinations of the proposed preprocessing techniques but considered a subset after an inference process. The authors suggested that future research should examine the impact of these and other preprocessing strategies in various domains, as well as other combinations and their interactions.

## 3.2 Literature review

In this section, we report the results of some of the most relevant and recent studies that employ text preprocessing techniques to evaluate their effects. These studies not only use preprocessing techniques but also conduct comparative evaluations using one or more models and/or datasets. For a detailed discussion on the preprocessing techniques and the corresponding related work, please refer to Section 3.3.

Recently, the authors in Kurniasih and Manik, 2022 used various deep neural architectures, excluding Transformers, to examine the impact of preprocessing on a pre-trained BERT model when fine-tuning it as the first embedding layer. They found that text preprocessing had a negligible influence on most of the models tested. The study was conducted on a single Indonesian dataset containing 3,217 instances from the Water Resources Agency of Jakarta, classifying textual reports into five categories. The authors used an Indonesian pre-trained version of BERT for the embedding. Given the substantial changes in performance outcomes between models with and without text preprocessing, the authors suggest that future studies should examine the impact of each text preprocessing step.

To investigate the effects of different preprocessing techniques, the authors in Hair Zaki *et al.*, 2022 applied fourteen text preprocessing approaches to datasets from Twitter, Facebook, and YouTube. They used text preprocessing algorithms in a specific order and employed an SVM to assess the variation in accuracy for sentiment classification. The results showed that consistently using all the preprocessing approaches could achieve an accuracy of 82.57% using unigram representations. Although the proposed preprocessing strategy proved effective on the selected dataset, an in-depth investigation using deep learning models is lacking.

The performance of an SVM classifier was also evaluated in Bao *et al.*, 2014 on a Twitter dataset for sentiment classification. The authors explored combinations of preprocessing techniques and found that reserving URL features, normalizing repeated letters, and transforming negations increased the accuracy of sentiment classification. Conversely, accuracy decreased when stemming and lemmatization were used. Adding bigrams and emotion features to the initial feature space resulted in superior outcomes.

In Garg and Sharma, 2022, the authors employed traditional models like Naive Bayes, SVM, K-means, and Fuzzy logic algorithms. Specifically, on a Twitter dataset, they explored three basic preprocessing methods: tokenization, removing stop words, and stemming. The findings indicated that preprocessing had a significant impact on reducing data dimensionality, leading to higher performance in sentiment analysis classification tasks.

For unstructured product review data, the authors in Arief and Deris, 2021 demonstrated that the correctness of classifier predictions depends on a suitable text preprocessing sequence. The dataset used for training consisted of product reviews from Amazon, with ratings of one or two stars collapsed into negative reviews and ratings of four or five stars classified as positive. The authors employed traditional models, including Naive Bayes, Decision Tree, and SVM.

Four traditional classifiers (Naive Bayes, Logistic Regression, SVM, and Random Forest) were also employed in Jianqiang and Xiaolin, 2017, where the authors explored the impact of six preprocessing techniques using five different Twitter datasets. They discovered that extending acronyms and substituting negations, as opposed to removing URLs,

**Table 3.1:** Acronyms for preprocessing techniques and real case examples, raw and preprocessed.

| Acronym | Technique | Raw | Preprocessed |
|---------|-----------|-----|--------------|
| DON | Do Nothing | *'Like a Rolling Stone'* | *'Like a Rolling Stone'* |
| RNS | Replace Noise and Pseudonimization | *'@Obama 0x10FFFF tells #metoo! bit.ly/~'* | *'USER tells HASHTAG! URL'* |
| RSA | Replace Slang/Abbreviations | *'omg you are so nice!'* | *'Oh my God you are so nice!'* |
| RCT | Replace Contraction | *'wedon't like butterflies.'* | *'wedo not like butterflies.'* |
| RRP | Remove Repeated Punctuation | *'welike her!!!'* | *'welike her multiExclamation'* |
| RPT | Removing Punctuation | *'You. are. cool.'* | *'You are cool'* |
| RNB | Remove Numbers | *'You are gr8.'* | *'You are gr.'* |
| LOW | Lowercasing | *'You Rock! YEAH!'* | *'you rock! yeah!'* |
| RSW | Remove Stop Words | *'This is nice'* | *'is nice'* |
| SCO | Spelling Correction | *'Ilenia is so kind!'* | *'Ilenia is so kind!'* |
| POS | Part-of-Speech Tagging | *'Kim likes you'* | *'Kim (PN) likes (VB) you (N)'* |
| LEM | Lemmatization | *'webe go to shopping'* | *'weam go to shop'* |
| STM | Stemming | *'Girl's shirt with different colors'* | *'Girl shirt with differ color'* |
| ECR | Remove Elongation | *'You are cooool!'* | *'You are cool!'* |
| EMO | Emoticon HaTMLCing | *':)'* | *'happy'* |
| NEG | Negation HaTMLCing | *'weam not happy today!'* | *'weam sad today!'* |
| WSG | Word Segmentation | *'#sometrendingtopic'* | *'some+trending+topic'* |

numerals, or stop words, enhanced classification results in terms of F1-measure and accuracy.

Transformers were used in Cunha *et al.*, 2021, where the authors removed stop words and kept only features appearing in at least two documents before applying TF-IDF. The experimental findings showed that in smaller datasets, shallow and straightforward non-neural methods achieved some of the best results. Conversely, Transformers performed better in terms of classification accuracy in larger datasets. However, the study only marginally focused on the impact of text preprocessing.

Regarding a Twitter-related task on irony detection, the authors in González *et al.*, 2020 performed a case-folding preprocess of tweets before tokenizing with the TokTokTokenizer from NLTK. They replaced hashtags, user mentions, and URLs with generic labels and shortened elongated words. While the authors employed BERT as a classification model, they only used the preprocessing strategy discussed above.

The authors in Cunha *et al.*, 2020 introduced and applied a new preprocessing strategy based on three steps: lowering dimensionality, increasing sparseness, and reducing the number of training samples. These steps proved to improve performance and/or reduce execution time. A significant finding reported in the study is that proper data preprocessing is more crucial than the classification algorithm itself, especially for achieving the best performance at the lowest possible cost.

### 3.3 Preprocessing techniques

This section presents the preprocessing techniques found in the literature, using a systematic methodology. A recent comparative survey by Symeonidis *et al.*, 2018 evaluates various text preprocessing techniques on two Twitter datasets designed for sentiment analysis. This article served as the foundation for our work due to its comprehensive coverage of available techniques, as shown in Table 3.2.

To compile the list of related works on preprocessing techniques, we included all studies cited by or citing Symeonidis *et al.*, 2018 that discussed at least three different preprocessing techniques. Techniques not covered in Symeonidis *et al.*, 2018 were added as columns to Table 3.2 and discussed accordingly. Studies focusing on fewer than three techniques are not included in the table but are briefly discussed in Section 3.3 if they offer novel or deeper insights into specific techniques.

For each study added to the reference list, we included papers cited by or citing each work in Table 3.2, provided they discussed at least three different preprocessing techniques. This approach ensures that, to the best of our knowledge, the most frequently cited preprocessing techniques in the literature are included in this chapter.

The preprocessing techniques discussed here represent the initial stage for any text classification task following lexical tokenization. As defined in Jurafsky and Martin, 2009, tokenization involves separating a continuous text into words. Various preprocessing techniques can then be applied to these words. The subsequent step after text preprocessing is splitting the text into n-grams (probabilistic tokenization). Before feeding the preprocessed text into a model, it must be tokenized into a numerical form that a computer can process.

While some studies present tokenization (lexical or probabilistic) as a preprocessing technique, we do not include tokenization among the techniques discussed here. The techniques in this chapter are characterized by their ability to alter the syntactic and semantic content of a text after lexical tokenization. Tokenization, whether lexical or probabilistic, is a necessary procedure to fragment text for subsequent processing stages. However, since tokenization is often considered part of preprocessing, we introduce and discuss it in the remainder of this section.

**Table 3.2:** Techniques discussed in related work that proposes at least three different preprocessing methods.

| Article | RNS | RSA | RCT | RRP | RPT | RNB | LOW | RSW | SCO | POS | LEM | STM | ECR | EMO | NEG | WSG |
|---|---|---|---|---|---|---|---|---|---|---|---|---|---|---|---|---|
| Alam (2019)Alam and Yao, 2019 | × | | | | | | | × | | | | × | | × | | |
| Albalawawe(2021)Albalawi et al., 2021 | × | | | | × | × | | × | | | | × | × | | | |
| Alzahranwe(2021)Alzahrani and Jololian, 2021 | × | | | | × | × | | × | | | | | × | | | |
| Anandarajan (2019)Anandarajan et al., 2019 | × | | | | × | × | × | × | | × | × | × | | | × | |
| Angianwe(2016)Angiani et al., 2016 | × | | | | | | × | × | × | | | × | | | | |
| Araslanov (2020)Araslanov et al., 2020 | × | | | | | × | × | × | × | × | × | × | | | × | |
| Babanejad (2020)Babanejad et al., 2020 | × | | | | × | × | × | × | | | | × | | | × | |
| Bao (2014)Bao et al., 2014 | × | | | | | | | × | | | | × | | | × | |
| Denny (2018)Denny and Spirling, 2018 | | | | | × | × | × | × | | | × | | × | | | |
| Duong (2021)Duong and Nguyen-Thi, 2021 | | × | | | × | × | × | × | × | × | | | × | × | × | |
| Hacohen (2020)HaCohen-Kerner et al., 2020 | | | | | | | × | × | × | | | | × | | | |
| Haddiwe(2013)Haddi et al., 2013 | | × | | | | | | × | | | | × | | | × | |
| Hickman (2022)Hickman et al., 2022 | × | × | | | | | × | × | × | | × | × | × | × | × | |
| Jianqiang (2017)Jianqiang and Xiaolin, 2017 | × | × | × | | | × | | × | | | | × | | | | |
| Kadhim (2018)Kadhim, 2018 | × | | | | | | | × | | | | × | | | | |
| Kathuria (2021)Kathuria et al., 2021 | × | × | | | × | × | × | × | | × | × | × | × | | × | |
| Koopman (2020)Koopman and Wilhelm, 2020 | | | | | × | | × | × | | | × | × | | | | |
| Kowsarwe(2019)Kowsari et al., 2019 | × | × | | | × | | | × | | | × | × | | | | |
| Kumar (2019)Kumar and Dhinesh Babu, 2019 | × | × | | | × | | | × | × | | × | × | × | × | | |
| Kunilovskaya (2021)Kunilovskaya and Plum, 2021 | | | | | × | | × | × | | × | × | | × | | | |
| Lison (2017)Lison and Kutuzov, 2017 | | | | | | | | × | | × | × | | | | | |
| Mohammad (2018)Mohammad, 2018 | × | | × | | × | | × | × | | | × | × | | | | |
| Naseem (2021)Naseem et al., 2021 | × | × | × | | × | × | × | × | × | | × | | × | × | | × |
| Petrovic (2019)Petrović and Stanković, 2019 | | | | | × | | × | × | | | | × | | | | |
| Pradha (2019)Pradha et al., 2019 | × | | | | × | | × | × | × | | × | × | | | | |
| Rosid (2020)Rosid et al., 2020 | | | | | | | × | × | × | | | × | | | | |
| Singh (2016)Singh and Kumari, 2016 | × | × | | | × | | | × | × | | × | × | × | | | |
| Smelyakov (2020)Smelyakov et al., 2020 | × | | | | | × | × | × | | | × | × | | | | |
| Symeonidis (2018)Symeonidis et al., 2018 | × | × | × | × | × | × | × | × | × | × | × | × | × | | × | |
| Toman (2006)Toman et al., 2006 | | | | | | | | × | | | | × | | | | |
| Uysal (2014)Uysal and Gunal, 2014 | | × | | | | | | × | × | × | × | × | | | | |
| Zong (2021)Zong et al., 2021 | × | | | | | | | × | | | | × | | | | |

Lexical tokenization, as discussed in Hassler and Fliedl, 2006; Mc-Namee and Mayfield, 2004; Vijayarani and Janani, 2016; A. Mullen *et al.*, 2018, typically involves splitting text into words. Probabilistic tokenization, on the other hand, can segment text into smaller or larger units called tokens. While common tokenization methods operate at the word level, various sub-word tokenization strategies are also explored in the literature (Sennrich *et al.*, 2016; Kudo, 2018; Schuster and Nakajima, 2012). Regardless of the tokenization window size, the process generally involves segmenting text. Usually, only alphanumeric or alphabetic characters separated by non-alphanumeric characters (e.g., whitespace, tabs, punctuation) are considered during segmentation.

The goal of probabilistic tokenization is to produce single units of information—the tokens—that can be mapped into numerical representations. The token list serves as the foundation for further processing, such as text mining, parsing, or classification. Both linguistics (where tokenization segments text into words) and computer science (where probabilistic tokenization maps tokens into numbers) benefit from this process. However, the complexity of tokenization can vary depending on the language's syntax. For instance, in languages like Italian and English, most words are delimited by whitespace. In contrast, languages like Chinese do not have obvious word boundaries, making the process more challenging and requiring techniques known as word segmentation.

When applying multiple preprocessing techniques in combination, the order can be crucial. While some techniques, such as removing stop words and punctuation, can be applied independently, others require careful consideration of their sequence to ensure consistent results. For example, Part-Of-Speech (POS) tagging should be applied before stemming, and negation handling should be done before removing stop words to ensure the tagger functions correctly. As noted in Babanejad *et al.*, 2020, it is not always necessary to perform preprocessing on both the training and test sets.

Given the methodology outlined earlier and throughout this chapter, the histogram in Figure 3.1 displays a list of preprocessing techniques documented in the literature. The histogram also indicates the frequency with which these techniques have been used in related works.

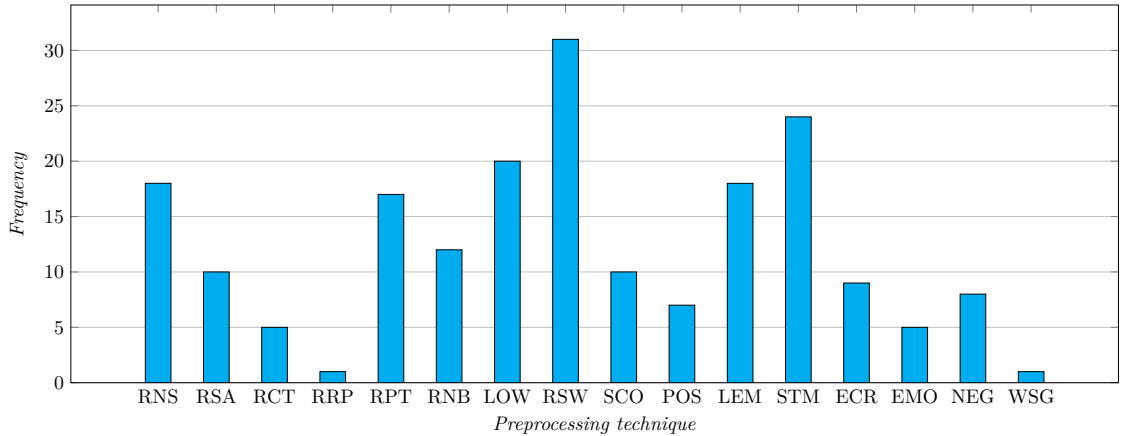

**Figure 3.1:** Number of times that the techniques discussed in this article are found in related work. In Table 1 are reported the expanded acronyms under the bars. The works related to the Figure are the ones listed in the Table 3.2. Each bar in the Figure actually shows the counts of the X in the table for each column.

### 3.3.1 Replace noise and pseudonimization

The definition of noise varies significantly according to the literature, with regard to removing and/or replacing noise. Usually noise replacement consists in replacing or removing unwanted strings and Unicode characters, which are regarded as crawling by-products, that can add further noise to the data. For this reason, some authors employ regular expressions to eliminate Unicode strings and non-English words. The authors in Babanejad *et al.*, 2020 do not explicitly mention noise removal. However, they apply a few text preprocessing techniques at the beginning of their evaluation. These techniques involve removing HTML tags and special characters from text, such as *"%*=()/"*. Furthermore, not all datasets are provided as plain text.

Especially in the context of sentiment analysis, another form of noise replacement is *pseudonimization*. User-posted tweets may include URLs, user mentions or hashtags (such as @*username* or #*music*), or both. In this way, users can link their tweet to a certain subject or user, and these strings of characters, depending on the task, can be treated as noise replacing them with specific tags. In the literature are described a number of methods to deal with this additional data supplied by users. In Agarwal *et al.*, 2011, authors replace all the URLs with a tag *U*, and replace user mentions (e.g. @*brucespringsteen*) with the tag *T*. The majority of academics believe that URLs don't reveal anything about the sentiment of a tweet (Ketsbaia *et al.*, 2020; Indra *et al.*, 2016; Aljebreen *et al.*, 2021; Resyanto *et al.*, 2019). Other scholars expand URLs from Twitter into full URLs before tokenization (Borra and Rieder, 2014; Benzarti and Faiz, 2015). The tweet text is then refined by removing any URLs that match the tokens. In conclusion, no general rules apply in definition and managing of noise. Definition and operations can vary significantly from a study to another.

### 3.3.2   Replace slang and abbreviation

Considering the character count restrictions in social networks (e.g., Twitter), abbreviations, acronyms, informal writing styles, short words and slang are frequently used (Tan *et al.*, 2015). These words have to be managed (e.g., replacing *OMG* with *Oh My God*). By haTMLCing these informal words in the text and changing them to reflect their actual meaning, an automated classifier may perform better while preserving information. These words and sentences can be managed in order to impute their meaning accurately. In Kouloumpis *et al.*, 2011 slangs and abbreviations are converted into word meanings that can be comprehended by utilizing conventional text analysis methods. In Symeonidis *et al.*, 2018 authors manually compile a lookup database with these words, phrases, and their replacements. However, it is worth noting that word embedding-based models could eventually manage slang and abbreviation as-is, understanding from the context, during the training phase, their original meaning.

### 3.3.3   Replace contraction

Contractions are short-form words that are used by users to reduce the number of characters in a tweet/post (Sagolla, 2009). An apostrophe is used in contractions to replace one or more missing letters. One preprocessing method consists of performing contraction replacement (e.g., *can't* be replaced by *cannot*).

Expanding contractions could or could not be a beneficial preprocessing technique before performing probabilistic tokenization. In a word embedding layer which splits words at a space character, further meaning could be provided, keeping the word *can't* instead of *cannot*. This way, a single word can incorporate what is expressed by the two single consecutive words *can* and *not*. However, words like *not* could be of prominent importance for subsequent stages coming later, like the ones that replace negations with antonyms. Otherwise, if the splitting of the words is performed at punctuation, tokenization would create the tokens *can* and *'t*. In this last example, as it matches other negative forms in the text, this tokenization could not be all that helpful. It is worth mentioning that, even if the main referenced language of this thesis is English, some interesting considerations could be made concerning other languages. For example, French has a contraction phenomenon which consists of truncating many words (for example, *manif* for *manifestation*), and Italian often presents articles with an apostrophe (e.g., *L'arte della guerra*, 'The art of war'), which should likewise be managed when focusing with these languages.

### 3.3.4 Remove repeated punctuation

In Symeonidis *et al.*, 2018, authors distinguish three punctuation signs: stop marks, question, and exclamation. These punctuation marks, according to authors, indicate the presence of emotion in the text considered. Because of this, authors substitute a representative tag in its place. For instance, "multiQuestionMark" is used in place of the token "???". This procedure is performed before deleting punctuation. However, in the not pre-trained models evaluated in this PhD thesis, if there is not any space between repeated punctuation marks, a separated word is created in the dictionary. As an example, given the sentence: "*Are you sure???*", three different words will be considered as separated tokens (i.e., *Are, you* and *sure???*). In the case of a single and/or multiple spaces (i.e., "*Are you sure ???*"), four words/tokens will be added to the dictionary (i.e., *Are, you, sure* and *???*). Of course, these different splitting strategies would lead to different behaviors of a subsequent classifier.

### 3.3.5   Remove punctuation

In written texts, punctuation can be used to express sentiment and emotion (Thelwall, 2017) (e.g., "*You are late! Hurry up!*"). Even if this punctuation use can be easily understood by humans, it could not be so for an automatic classification tool. Furthermore, punctuation can be useless when dealing with certain text classification tasks. For this reason, punctuation removal is often applied in many preprocessing tasks for automated text classification. However, punctuation symbols can also denote sentiment. In Balahur, 2013, authors detect punctuation signs like "*!!!*" and replace them with the label "*multiexclamation*". An application where punctuation is removed can also be found in Lin and He, 2009. In the study presented in Siino *et al.*, 2021, the authors do not remove punctuation during preprocessing. In fact, they consider as separate entries in the dictionary the words *up* and *up!*. In this way, the word embedding layer, trained from scratch in the study, at the end of the training phase is able to differentiate the meanings of the two entries in the dictionary assigning different word vectors in the embedding space. These behaviours could be, eventually, able to get the intended meaning of the version with the exclamation mark, to invoke someone for moving faster. Removing punctuation from the sentence and replacing it with a single space (i.e., "*You are late Hurry up*"), would result in the change of some latent information, maybe of interest for certain text classification tasks (e.g., author profiling as in the study of Siino *et al.*, 2021).

### 3.3.6   Remove numbers

Despite the fact that numbers can offer helpful data to obtain a performance gain of a classifier, it is usual to delete them during the preprocessing stage (Lin and He, 2009; Anandarajan *et al.*, 2019). Such a practice could be due to historical reasons, where computational power and traditional machine learning classifiers required a stricter preprocessing phase to lighten datasets. However, other scholars (Denny and Spirling, 2018; Siino *et al.*, 2021) argue that numbers are useful, indeed they do not remove them from the original source text.

  In fact, the sentence: "*we won 2 dollars on bets.*" compared to: "*we*

*won 2,000,000 dollars on bets.*" will become: "*we won dollars on bets.*". However, the resulting sentence has lost the intended meaning of the user who pronounced it. Such a meaning could be considered differently by an attention based model or even by a shallow neural network to provide the correct prediction. Even in the case of author profiling tasks, the use of numbers could characterize a user based on the quantity expressed by the numbers in text. Removing numbers could lead to another type of information loss. For instance, the removal of *4* from the sentence: "*we did it 4 you*" (i.e., "*we did it you*") would alter the original true meaning of the sentence even for a human classifier. Finally, removing the number *8* from the word *w8*, again, could lead to a loss of information and to a deterioration in performance as well as in the previous example.

### 3.3.7 Lowercasing

Among others, lowercasing (i.e., converting uppercase to lowercase letters) is one of the most common techniques to perform preprocessing on a source text before further steps.

Lowercasing is discussed in Camacho-Collados and Pilehvar, 2018 and consists in converting to lowercase each character of a text (e.g., "*Your band sounds like Rolling Stones*" — "*your band sounds like rolling stones*"). Before the classification step, authors in Uysal and Gunal, 2014 change capital letters from uppercase to lowercase. According to authors, the classification's performance has improved. Lowercasing has been a common method in many deep and non-deep architectures presented in the literature due to its simplicity. Lowercasing may have undesirable effects on system performance since it increases ambiguity despite the fact that it reduces vocabulary size and sparsity (Djuric *et al.*, 2015). In the example reported above — regarding the rock band The Rolling Stones — lowercasing could produce for a non-human classifier an ambiguity, comparing the sound of a band to a set of stones rolling[1] instead of comparing the same sound to the popular rock band.

Lowercasing, on the other hand, conflates multiple spellings of words that are based on case. The diversity of capitalization in the dataset may interfere with classification and degrade performance. This could be the case of a single misspelled word in a dataset (e.g., "*houSe*"). In this case, a word embedding layer trained from scratch could assign a new embedding vector instead of using the most properly semantic-related word "*house*".

---

[1] ...and in this case, maybe, you should look for a new drummer.

Differences in experimental results across various works in the literature can be simply explained based on the domains considered.

### 3.3.8   Remove stop words

The removal of stop words, according to this study, is the most often employed preprocessing method found in the literature. Stop words are typically frequent terms in a language and are assumed to be the least informative (Gerlach *et al.*, 2019) (i.e., stop words alone do not provide meaning to document). Stop words are language-specific and cannot be considered as keywords in text mining applications, so they could be useless in information retrieval. Stop words often appear in writings without being related to a specific subject (e.g., prepositions, articles, conjunctions, pronouns etc.). Before performing the text classification task, stop words are typically removed. The size of a dataset is actually decreased after removing stop words from it. Example of stop words are: "of", "a", "the", "in", "an", "with", "and", "to". Depending on the list used, there are usually more than 400 stop words in the English language (Dolamic and Savoy, 2010; Flood, 1999).

The first study considering stop words is conducted in Luhn, 1960. There, the author makes the suggestion that words in written texts can be split into terms considered as keyword or non-keyword using a stop list. In Saif *et al.*, 2014, the authors employ data from six different Twitter datasets to use different stop word detection algorithms and examine how eliminating stop words impacts the effectiveness of two popular supervised sentiment classification techniques. By tracking changes in the classification performance, in the amount of data sparsity and in the size of the feature space of the classifier, the authors evaluate the effects of eliminating stop words. Authors compare results between static stop word removal techniques (e.g., based on pre-compiled lists) versus dynamic stop word removal techniques (Makrehchi and Kamel, 2008) (e.g., based on dynamic detection of stop words in a document). The results demonstrate that the performance is adversely affected by the usage of pre-compiled stop words list. Otherwise, the best strategy to retain significant performance while lowering data sparsity and significantly condensing the space of the features appears to be the dynamic creation of stop word lists by deleting those uncommon words appearing rarely in the dataset. Researchers have found that a word's relevance can be inferred from its frequency in a data collection. This discovery led to the exploration of various well-liked stop word removal techniques in the literature. While some approaches consider both the top and the bottom-ranked words to be stop words, others make the assumption that stop words correspond to the most frequently occurring words. Another well-liked alternative to using the raw frequency of terms has also been discussed in the literature: Inverse Document Frequency (IDF). To conclude this section, four different stop word removal techniques are now described.

- *The traditional approach.* The traditional approach (Rijsbergen, 1979) relies on removing stop words gleaned from pre-compiled lists.

- *Approaches based on Zipf's law.* Three approaches for creating stop words that are moved by Zipf's law exist, besides the conventional stop words list (Courseault Trumbach and Payne, 2007; Makrehchi and Kamel, 2008). Among these are the words that are most frequently used and words that only appear once, or singletons. Additionally, terms having a low inverse document frequency are thought to be removed (IDF).

- *The mutual information method.* A notion of how informative a term can be about a certain class is supplied by a supervised technique that determines the amount of information that each word and document class share (Cover and Thomas, 2001). A lower mutual information means that the word has a weak ability for helping in discrimination, hence it needs to be dropped.

- *Random sampling of data chunks.* It was initially suggested in Lo *et al.*, 2005 to use this technique to manually identify stop

words in web publications. This approach operates by repeatedly processing different, randomly chosen, data chunks. It then uses the Kullback-Leibler divergence (Joyce, 2011) metric to order the terms in each chunk according to how informative they are.

### 3.3.9   Spelling correction

It is common that texts shared online by users contain spelling errors. For instance, tweets frequently contain typos as well as grammatical errors. These errors might make classification tasks more problematic. The unintended consequence of having the same term transcribed differently is lessened by correcting spelling and grammar errors. Examples of misspelled words are: *absense*, *decieve*, *noticable*. After a spelling correction step, the mentioned words would be substituted respectively by: *absence*, *deceive*, *noticeable*. In Mullen and Malouf, 2006 it is proven that correcting spelling errors can improve classification effectiveness. Although other type of errors could be introduced after performing a spelling correction, this step generally improves performance.

Eventually, an interesting way to perform spell-checking is presented in Virmani and Taneja, 2019 where a spell checker is employed to improve stemming, while synonyms of related tokens are combined.

### 3.3.10   Part-of-Speech tagging

The word class is identified via POS tagging, which takes into account the word's placement in the sentence (Manning *et al.*, 2002). A POS tag is then given to any word in a sentence. Noun (NN), proper plural noun (NNPS), verb (VB), adverb (RB), superlative adverb (RBS), third-person verb (VBZ), and other tags are examples of tags[2]. It has been demonstrated that four POS classes—namely, nouns, adjectives, verbs, and adverbs—are more informative than other classes. Several purposes of POS tagging in preprocessing are discussed in related work. In Symeonidis *et al.*, 2018 the use of POS tagging allows some parts of speech to be excluded since they do not express the suitable sentiment for the purpose at hand. Only verbs, adverbs, and nouns were kept in the study. In Barbosa and Feng, 2010, in order to tag opinion statements with sentiments, the authors employ POS tags as pointers. In the literature, exist dozens of different tag sets, defined in the context of different theoretical frameworks and also designed to represent morphologically different languages. The above-mentioned tag set is the one related to a popular project of the last century for the construction of a treebank of English language (i.e., the *Penn Treebank*). The tag set is still used today, but has been superseded by others more suited to represent not only the English language. One of the most relevant is the tag set project of Universal Dependencies[3].

---

[2] An example from Twitter is the case of a retweet replaced by the tag *RT*

Some popular libraries and tools that use rule-based approaches to perform POS tagging are the NLTK library's *pos_tag()*[4] and the *TextBlob*[5] Python library. Other libraries based on statistical models are the *spaCy library's POS tagger*[6] that is trained on the OntoNotes 5 corpus and the *Averaged Perceptron Tagger in NLTK*[7] that is based on the above-mentioned tag set project of the Universal Dependencies.

Specially in deep learning-based models, this process of assigning POS to each term is helpful to increase semantic informativeness in text. However, due to its impact on diminishing accuracy, some authors choose to omit POS tagging for certain tasks (Boiy *et al.*, 2007), while others found POS tagging useful (Anandarajan *et al.*, 2019).

### 3.3.11 Lemmatization

Lemmatization is used to replace a word with its corresponding lemma, or dictionary form. By analysing a word's location in a sentence and removing its inflectional ending, this technique creates the lemma as it appears in a dictionary (e.g., *Performance is greatly improved*, replaced by *Performance be greatly improve*). In Guzman and Maalej, 2014, lemmatization reduces various word forms to the same lemma to enhance user sentiment extraction effectiveness. Lemmatization is discussed in Camacho-Collados and Pilehvar, 2018 and, in the context of an SVM model, in Leopold and Kindermann, 2002. In Kuznetsov and Gurevych, 2018 authors address the issue of ambiguity after lemmatization. Authors use lemmatization in combination with POS disambiguation to alleviate the problem.

---

[3]https://universaldependencies.org/
[4]https://www.nltk.org/api/nltk.tag.pos_tag.html
[5]https://textblob.readthedocs.io/en/dev/quickstart.html
[6]https://spacy.io/api/tagger
[7]https://www.nltk.org/api/nltk.tag.perceptron.html

Lemmatization has long been a common preprocessing step for traditional models. Since deep learning models started to be employed, lemmatization has rarely been used as a preprocessing stage. Lemmatization's major goal is to reduce sparsity because a dataset may contain various inflected versions of the same lemma. Furthermore, in the context of author profiling tasks, lemmatization can lead to ignore relevant writing style details (Hernández Farías *et al.*, 2019). Eventually, it is worth reporting that in inflexionless language (e.g., Chinese), words are only in one form. For inflexionless languages, techniques like lemmatization or stemming, does not provide any change to the text.

### 3.3.12   Stemming

To obtain stem versions of derived words, a process known as stemming is used. For instance, stemming techniques can reduce word variations like *easy*, *easily*, *easier*, *easiest* to the word *easy*. The dimensionality of dictionaries is decreased, since many words are collapsed to the same one. This procedure reduces entropy and raises the significance of the concept behind a word like the one from the previous example (i.e., *easy*). In the end, stemming enables the same consideration of nouns, verbs, and adverbs that share the same stem. Word frequencies are commonly calculated after stemming, since derived words share semantic similarities with their root forms.

The first known stemming algorithm was presented in 1968 and discussed in Lovins, 1968. Going forward, the algorithm for stemming introduced in Porter, 1980 has been often employed by a multitude of scholars. It is likely the most popular and effective stemming technique for the English language.

Stemming is applied in Srividhya and Anitha, 2010 and also discussed in Vijayarani *et al.*, 2015. The goal of stemming in both studies is to find, for any derived word, its corresponding stem. As discussed in Gemci and Peker, 2013, the stemming algorithm depends on the language considered (i.e., Turkish in this case). The library commonly used for Turkish language is discussed in Akın and Akın, 2007. For the same language, the fixed-prefix approach described in Can *et al.*, 2008 is a computationally straightforward yet highly efficient stemming tool. The performance and efficacy of stemming in applications like spelling checkers across languages are examined by authors in Gupta and Lehal, 2011. Although advanced algorithm employ morphological understanding creating a stem from the words, a typical simple stemming technique would involve deleting suffixes using a list of frequently occurring suffixes. The study provides a comprehensive overview of known stemmers for the Indian language, as well as popular stemming strategies.

Truncating approaches, statistical methods, and mixed methods are typically used to apply stemmed algorithms. The mechanism used by each of these divisions to determine the word variations' stems is different. Below is a discussion of a few of these techniques. For further discussion on stemming techniques, a deep overview is presented in Moral *et al.*, 2014.

- *Truncating techniques* involve removing a word's prefixes or suffixes, referred to as affixes. Truncating a word at the n-th character, is the simplest basic stemmer (i.e., it consists in keeping *n* letters and removing the remaining). Words that are shorter than *n* are left untouched using this strategy. When the word length is short, there is a greater chance of over stemming.

- *Porter stemmer* is one of the most well-known stemming algorithms developed in 1980 (Porter, 1980). On the fundamental algorithm, numerous alterations, improvements, and suggestions have been proposed. The original algorithm is based on the fact that in the English language, suffixes are usually composed of groupings of simple and small suffixes. The algorithm is performed along five steps. Each stage applies the rules until one of them satisfies the criteria. If a match is found, the suffix is then removed and the subsequent action is evaluated. At the end of the last stage, the resultant stem is returned. A stemming framework named *Snowball* was created by Porter. The primary goal of the framework is to give developers the freedom to create custom stemmers for different languages or character sets.

- *Lovins stemmer* was proposed in 1968 (Lovins, 1968). The Lovins stemmer eliminates a word's longest suffix. Each word is altered, checking a different table that performs numerous alterations to turn these stems into acceptable words after the ending has been deleted. Due to the fact that it is a one pass method, it can never remove more than one suffix from a word. This algorithm has the following benefits: 1) it is extremely quick; 2) it can haTMLCe changing letters doubled for words as *getting* into *get* and 3) it can haTMLCe plurals that are irregular (e.g., "mouse" and "mouses", "die" and "dice" etc.). It is worth reporting that the Lovins stemmer, although being a heavier stemmer, results in superior data reduction. With its extensive suffix collection, the Lovins method only requires two significant stages to delete a suffix. The algorithm by Lovins is quicker than the Porter one, based on five iterations. Due to its extremely long endings list, it is larger than the Porter method.

- *Paice/Husk Stemmer* is introduced in Paice, 1990 and is an ongoing method using one database that has more than one hundred rules and uses the final character of a suffix as the index. It tries to determine the relevant rule based on the final character of a word. Rules detail the substitution or deletion of a word ending. If any rule does not match, the algorithm ends. The algorithm ends also if the first character of a word is a vowel and no more than two or three letters remain in the word. If not, the rule is followed and the procedure is repeated. The benefit is that both deletion and replacement as per the rule are applied at every iteration. However, because of the weight of this stemmer, over-stemming can happen.

The two primary categories of stemming issues are over- and under-stemming. If two words having different stems are replaced by the same root, then a case of over-stemming occurs. Another term for this is a false positive. On the other hand, the act of giving two words that ought to share the same root a different root is called under-stemming.

This is also known as a false negative.

### 3.3.13 Removing elongation

A character that is repeated once or more times can be found in elongated words (e.g. *cooooool*, *greeeeeeat*, *goooood* etc.). Tweets and other social media posts frequently contain words with repeated letters that can be managed to better mine sentiment (Bakliwal *et al.*, 2012). Character repetitions are employed by users to emphasize and express their sentiments. The preprocess step of removing elongation consists of replacing elongated words with their source words, so they can be considered as the same entity. Repeated characters are reduced to a single one to prevent the learner from considering lengthened words differently from their basic form. If not, a classifier could interpret them as distinct words, and the longer words are likely to be underestimated because of their lower frequency in the text.

### 3.3.14 Emoticon and Emoji Handling

On the internet and in social networks, emotional icons are frequently used to denote users' sentiment (Hogenboom *et al.*, 2013). Users use different emoticons (e.g., *:)*, *:(* etc.), to express opinions too. Not to be confused with emoticons, emojis are pictographs of objects, faces, and symbols. However, in a generic preprocessing step, the same operations used for emoticons can be applied to emojis too. Depending on the considered task, it could also be important to capture information provided by emoticons or emojis to perform text classification.

In Wang and Castanon, 2015 authors study and evaluate the impact of emoticons on sentiments of tweets. The authors demonstrate the value of emotional icons in conveying messages on social media. In Pecar *et al.*, 2018, the usefulness of processing emoticons on user-generated content is highlighted by the authors.

Emoticons could also be replaced with scores that express a score against a polarity, but they can also be translated into text in the corresponding word. For example, for a specific sentiment classification task, the words *pos* and *neg* can be used in place of the positive and negative icons, respectively. In other studies, emoticons are substituted with the words that best describe them, such as *sad* in place of *:-(*. However, for instance, the irony in the usage of a sad emoticon while texting something positive, can revert the original meaning of a sentence.

In Agarwal *et al.*, 2011 authors employ emoticons as features and associate words to a value of pleasantness from one to three. Emoticons are scored similarly to other words and are broken down into the following classes: extremely negative, negative, neutral, positive and extremely positive.

Keeping as-is emoticons in any text, for word-embedding-based models, leads to the generation of a word vector with an associated semantic as for any other word in the dataset.

### 3.3.15   Negation Handling

As stated in Babanejad *et al.*, 2020, one of the best preprocessing methods for tackling tasks involving sentiment analysis is negation handling. A crucial stage in sentiment analysis is dealing with negations, such as "not nice". One of the most relevant causes of misclassification is the omission of negation words, which can affect the tone of all the surrounding words. One way to perform negation handling is removing negative forms in text to reduce ambiguities of the classified sentences. Specifically, when facing sentiment analysis tasks, negation is significant because, in many circumstances, the polarity of words or sentences can be affected by negation words, which can cause the polarity to invert. The most typical method of handling negation is to look for terms that are similar to "not" in each sentence, then see if the next word has an antonym. The word "sad" will be used in place of phrases like "not happy" for instance. To perform the replacement of words with the corresponding antonyms, it is generally used *WordNet*, presented in Miller, 1995.

In Babanejad *et al.*, 2020 authors handle negation by performing the following steps. At first, they compile an antonym dictionary using the WordNet dataset. In their work, the authors explain how to manage the three possible cases when looking for antonyms (i.e., a single antonym, multiple antonyms or no antonyms). The word's antonym is then randomly selected from the antonym dictionary considered. Eventually, the negation terms in tokenized text are identified by the authors. If is discovered a negation word, the token that follows it (i.e., the word to be negated) is selected, and the antonym of that word is searched in the dictionary of the antonyms. The negated word and the negation word are swapped out if an antonym is found. In their work, the authors provide a running example where the sentence *"I am not happy today"* is replaced by the sentence *"I am sad today"*.

Handling negations can generally improve performance for sentiment analysis-related tasks based on sentence classification. However, a comprehensive study on the effect of handling negations for author profiling tasks (i.e., classifying a whole dataset related to an author instead of performing classification of single sentences) is still missing.

Negation handling, mentioned here, usually solves the problem considering the presence of particles or adverbs of denial. Indeed, to treat negations effectively also on a larger portion of text (instead of single words), parsing strategies apply.

### 3.3.16 Word segmentation

It is quite common to find different words merged together in online texts. Such a case can be due both to a typing error or a deliberate choice. In the first case, a user could wrongly type the word *"Beyoncelemonade"* instead of the two different words *"Beyoncé Lemonade"*. The merged word represents noise and could likely be the only token in the dataset. In a tweet like: *"welike beyoncelemonade"* a model could not understand the topic (i.e., *music*) of the sentence. Considering the same merged word, a user could deliberately write *#beyoncelemonade* as a hashtag within the shared post. In this case, word segmentation would change the desired usage of the author, as reported in Naseem *et al.*, 2021. Nevertheless, segmenting merged words has proved to be helpful in understanding and better classifying the contents of tweets and postsPalmer, 1997Yamaguchi and Tanaka-Ishii, 2012.

In other cases, a model could benefit from processing words grouped. It is the case of words like *"United States"*, where splitting single words as different tokens could make it harder for a model to catch the underlying concept of the single word *"UnitedStates"*. In the second case, word embedding-based architectures could get the meaning of a whole sentence, understanding the reference to the specific country (i.e. the United States of America).

### 3.3.17  Conclusion

In this chapter, we have compiled and presented the most widely used preprocessing techniques from the literature. We then performed an evaluation and comparison of the three most common techniques in four datasets from various domains. To assess the impact of different combinations of preprocessing, the study in Siino *et al.*, 2024c conducted extensive tests using nine machine learning models. The study not only lists the best and worst performance strategies for each dataset and model but also suggests techniques that, whether used alone or in combination, consistently deliver superior performance. The results highlight the variability in performance based on the algorithm used, underscoring the importance of selecting an appropriate learning algorithm for the task to enhance the performance of text classification. The best preprocessing strategies, individually or in combination, were identified through rigorous testing and observation of the interactions between preprocessing methods. Our analysis emphasizes the critical role of data preparation in ensuring consistency when comparing different learning models. Furthermore, the research demonstrates that the choice of preprocessing method significantly affects the results, even with modern Transformers. These findings should encourage researchers to carefully select and document their preprocessing choices when evaluating or comparing models. According to the study, while techniques such as removing stop words and lowercase often perform better, the study indicates that to completely skip preprocessing is rarely optimal. The recent advancements in model capabilities, particularly with Transformers, have shifted focus from data preparation to model development. However, our findings underscore the importance of source data and preprocessing, which should not be overlooked. Effective preprocessing can enhance both the performance and understanding of the latest Transformer-based models, such as ChatGPT.

It is also worth mentioning that studies based on more recent

Transformer-based architectures, attempt to highlight their robustness while varying the preprocessing technique employed or perturbing the input text. However, in these studies, there is always a slight performance degradation that confirms how the optimal robustness to input variation is not reachable (Singh *et al.*, 2024; Peters and Martins, 2024; Aliakbarzadeh *et al.*, 2025). In conclusion, despite the impressive performance of modern Transformers, there is a tendency to overlook the real impact of preprocessing methods. Insights from this area may lead to more effective and consciously designed models, potentially revealing interesting mechanisms, especially in deep learning.

# 4

---

# Representation

---

Before advancing to the classification stage, it is essential to transform unstructured data, particularly free-running text, into organized numerical data. This transformation requires a document representation model to facilitate subsequent classification tasks following the text preprocessing stage. Text representation models convert text data into a numerical vector space, significantly influencing the performance of subsequent learning tasks. Throughout the history of NLP, word representation has been a critical area of interest, as it involves capturing the rich information embedded in text data for various applications.

This chapter explores the expressive capabilities of several word representation models, from traditional methods to contemporary language models. Various model designs, including language models, have been examined, along with a range of text representation techniques. These models can convert large volumes of text into useful vector representations that effectively capture relevant semantic information. Different machine learning models can leverage these representations for a variety of NLP tasks. Effective text representation, which captures intrinsic data properties, is likely to enhance performance.

In the following sections, we briefly discuss the drawbacks of the pro-vided representation models. Specifically, after preprocessing raw text, the next stage involves probabilistic tokenization based on a splitting strategy. Probabilistic tokenization separates text units and converts them into numerical representations. In automatic text classification, a single word is commonly used as the unit from the text. In this context, a single n-gram refers to a single word.

Although not strictly a text representation method, n-grams can be employed as features to represent units of text. A representation that uses single words (1-gram), regardless of order, is known as a Bag of Words (BoW). This approach is straightforward to implement and represents text as a vector, typically manageable in size. The terms 2-gram and 3-gram are frequently used. When two or more *grams* are used in place of a single gram (i.e., word) the term *n-gram* can be used. An illustration of a 2-gram is given in the following clause:

- *"Once upon a time you dressed so fine."*

In the proposed example, the tokens would be:

- {*"Once upon", "upon a", "a time", "time you", "you dressed" "dressed so", "so fine"*}

An Example of 3-Gram:

- *"Once upon a time you dressed so fine."*

In the proposed example, the tokens would be:

- { *"Once upon a", "upon a time", "a time you", "time you dressed", "you dressed so", "dressed so fine"*}

It is worth mentioning that also split strategies at the character level have been reported in the literature, as in Zhang *et al.*, 2015, where the authors show that a character-level CNN achieves interesting performance. Comparisons are made between deep models like word-based ConvNets and RNN and more conventional models like BoW, n-grams, and their TF-IDF variations. In this case, considering a sentence like:

- *"Purple Haze"*

The tokens are as follows:

- $\{$ *"P"*, *"u"*, *"r"*, *"p"*, *"l"*, *"e"*, *"H"*, *"a"*, *"z"*, *"e"* $\}$

The remaining part of this section covers various representation models that are frequently utilized. Over time, numerous researchers have proposed different solutions to address the problem of maintaining the syntactic and semantic connections of words within the selected representation. These methods are reviewed alongside relevant literature. We begin by discussing statistical methods, followed by an exploration of significant representation learning techniques and pre-trained language models.

## 4.1   Text representation models

### 4.1.1   Statistical models

The earliest and most straightforward methods for representing textual data are statistical word representation techniques. Early models for information retrieval, and NLP heavily relied on these word representation models due to their ease of design and application across various tasks. However, despite their simplicity, these models have several notable drawbacks:

- They do not consider the order of words.

- They overlook the relationships between words.

- The size of the input vector is proportional to the vocabulary size, making them computationally expensive and potentially leading to suboptimal performance.

This section presents these models, which were frequently used in the past for text classification. These word representation approaches are based on word frequency, converting text into a vector form that quantifies a word's usage frequency within a text. The following sections briefly describe common statistical techniques that are frequently employed in the literature.

*"Like a rolling stone"*

**Figure 4.1:** One-hot encoding example

## One-hot encoding

A fundamental method for representing text is one-hot encoding. In this approach, each categorical value is converted into a new categorical column, and a binary value of 1 or 0 is assigned to these columns. The dimensionality of one-hot encoding is equal to the number of terms in the vocabulary. Each vocabulary term is represented as a vector of binary values (0 or 1). After mapping each token to an integer value, a binary vector is used to represent this integer value, where all values are zero except for the index corresponding to the word in question, which is marked with a 1. Each unique word has its dimension, indicated by a single 1 in that dimension and 0s in all other dimensions. Consequently, with one-hot encoding, all words in the dictionary are orthogonal to each other.

Considering the following sentence:

- *"Like a rolling stone"*

The one-hot encoding representation is depicted in Figure 4.1.

## Bag of Words (BoW)

The Bag-of-Words (BoW) model is another method for representing documents. BoW creates a vector representation of a document by counting the frequency of terms within the text, a technique also known as a *vector space model*. This approach simplifies complex texts by treating them as unordered collections of words, effectively disregarding the semantic and structural connections between phrases. Despite these limitations, BoW has proven effective for various classification tasks.

The core idea behind BoW models is that each word is represented as a one-hot-encoded vector with a size equal to the vocabulary. Consequently, BoW-based methods are often combined with feature extraction techniques that consider word diversity, allowing a single vector to represent an entire document rather than individual words. However, as the vocabulary size can grow to hundreds of thousands of terms, this approach may introduce significant high-dimensionality challenges.

BoW is utilized in various fields, including machine learning for computer vision, Bayesian spam filters, and document categorization. In BoW, a body of text, such as a sentence or document, is viewed as a collection of words without considering their order or grammatical structure. The BoW process generates lists of words, ignoring their semantic relationships since the words are not structured into sentences. The meaning of a sentence can often be inferred from its constituent words, and the main topics of corpora can be determined by counting word frequencies rather than relying on grammar or word order.

However, the BoW representation has several limitations. These include high dimensionality, loss of correlation with adjacent words, and the inability to capture semantic relationships among terms in a document. Additionally, BoW models struggle with scalability due to the potentially vast vocabulary size, leading to issues such as identical vector representations for different phrases (e.g., "John loves Jane" and "Jane loves John"). Consequently, the size and scalability of BoW models present significant challenges for computer scientists and data scientists.

A BoW representation example is depicted in Figure 4.2.

## Term Frequency-Inverse Document Frequency (TF-IDF)

Term Frequency (TF), commonly paired with the BoW model, is another method for representing text. This approach assigns the feature space based on the number of tokens in each document. TF is a straightforward way to weigh words by mapping each word to a number that indicates how often it appears across the entire corpus.

*"As Long As You Love Me"*

**Figure 4.2:** BoW encoding example

Word frequency can be used as a boolean value or scaled logarithmically in methods that build upon TF. In these techniques, word frequencies in each document are converted into a vector. While this method is simple, it has limitations, as it can be dominated by frequently used words in the language.

For a corpus of texts, the relative frequency of a word in a single document compared to other documents is often used instead of the raw count. Notably, common terms tend to have less value in large corpora. To address this, TF is often weighted by Inverse Document Frequency (IDF). IDF reduces the impact of popular terms and boosts the significance of rarer words. The combination of TF and IDF is known as Term Frequency-Inverse Document Frequency (TF-IDF). The mathematical representations of TF, IDF, and TF-IDF are provided in Equations 4.1, 4.2, and 4.3.

$$tf_{ij} = \frac{n_{ij}}{|D_j|} \tag{4.1}$$

$$idf_i = \log_{10} \frac{|D|}{|d_i|} \tag{4.2}$$

$$tf - idf = tf_{ij} \times idf_i \tag{4.3}$$

Here $n_{ij}$ is the number of occurrences of the term $i$ in the document $j$. The number of terms in the document $D_j$ is $|D_j|$. Looking at Equation 4.2, $|D|$ is the total number of documents and $|d_i|$ is the number of documents containing the term $i$.

TF-IDF representations can become quite large, depending on the size of the vocabulary. To mitigate issues with memory usage and time complexity, one can limit the number of features included in the vectors. Alternatively, dimensionality reduction techniques can be applied to the full-sized representations.

Despite TF-IDF's efforts to handle common terminology, it has certain limitations. Since each word is treated as a separate index, TF-IDF cannot capture similarities between words. However, recent advancements in complex models have led to new approaches, such as word embeddings, which can account for word similarity and POS tagging.

### 4.1.2   Word embedding models

Statistical word representation methods struggle with the high dimensionality of dictionaries and fail to capture the semantic and syntactic meanings of words. To address these limitations, researchers developed techniques to represent words in low-dimensional spaces. Traditional statistical approaches fall short in modelling semantic meanings, even though they capture some syntactic relationships. For instance, synonyms, which are semantically similar, are treated as entirely distinct entities in these models, leading to orthogonal representations in the feature space.

Models like BoW ignore word meanings, treating semantically similar words (e.g., "auto," "car," "automobile") as orthogonal vectors. This issue hampers the model's ability to understand sentences, as it disregards word order. N-grams do not resolve this problem, necessitating methods that automatically learn representations for tasks like classification. These techniques, known as feature learning or representation learning, are crucial because machine learning models heavily depend on how input data is represented.

Deep learning models have largely replaced traditional feature learning approaches, as they can automatically learn critical features through both supervised and unsupervised methods. In NLP, unsupervised text representation techniques like word embeddings have become prevalent. These methods map text components, typically words, to n-dimensional vectors of continuous values, which can be processed by computers and capture semantic meanings (Siino, 2024c; Siino, 2024a). Relying on artificial neural networks, these techniques infer word meanings from their context within a text.

Word embeddings have significantly enhanced the performance of various downstream tasks due to their strong representation learning capabilities. Models like *Word2Vec*, *GloVe*, and *FastText* have improved classification outcomes by capturing more semantic and syntactic information than traditional linguistic features. However, these "static" embeddings, which assign a single vector to each word regardless of context, struggle with polysemy—where a word has multiple meanings. For example, the word "*sound*" has different meanings as a noun and an adjective, and a single embedding cannot effectively represent all its senses.

Additionally, models like Word2Vec and GloVe cannot handle out-of-vocabulary (OOV) terms, a problem addressed by FastText, which breaks words into n-grams. These limitations, along with poor performance on low-quality text, affect the effectiveness of text classification.

The following sections introduce Word2Vec, GloVe, and FastText, popular word embedding techniques successfully applied in deep learning. Subsequently, context-based representation techniques will be discussed.

**Word2Vec**

The authors in Mikolov *et al.*, 2013a introduced one of the earliest and most renowned word embedding frameworks, utilizing shallow neural networks to generate high-dimensional vectors for each word. Initially, Word2Vec included two models: the Continuous Skip-gram and the Continuous Bag-of-Words (CBOW). The CBOW model learns word representations by predicting a central word based on its surrounding context. Conversely, the Skip-gram model reverses this task by predicting a word's neighbouring words. These models tackle complex problems, aiming not to accurately predict words but to create meaningful mappings between words and their embeddings.

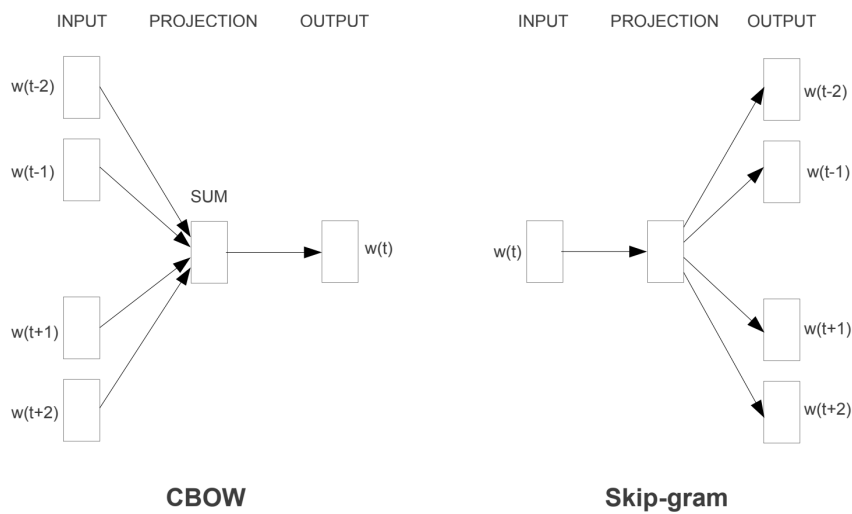

**Figure 4.3:** The original picture from the work on CBOW and Skip-gram models presented in Mikolov *et al.*, 2013a.

Figure 4.3 illustrates the original concept from Mikolov *et al.*, 2013a, showcasing a basic CBOW model. This method is a powerful tool for identifying relationships and word similarities within corpora. For example, the embedding can capture the proximity of words like "large" and "bigger" in the vector space.

**Continuous BoW Model**. For a specific word, the Continuous Bag-of-Words (CBOW) model uses multiple surrounding words as its representation. For instance, for the target word "air-force," context words might include "airplane" and "military." This involves creating multiple connections from the input to the hidden layer, with the number of connections equal to the number of context words. The first step is to create a vocabulary, which is a list of all unique words in the corpus. The shallow neural network's task is to predict the target word given its context. The number of context words used depends on the window size setting, which typically ranges from 4 to 5 words.

**Continuous Skip-Gram Model**. This architecture closely resembles CBOW but aims to maximize the classification of a word based on the preceding word in the same phrase, rather than predicting the next word from its context. Both the Continuous Bag-of-Words (CBOW) and Skip-gram models help preserve the syntactic and semantic content of sentences for machine learning algorithms.

**Global Vectors for Word Representation (GloVe)**

Another notable word embedding approach is GloVe (Global Vectors for Word Representations) (Pennington *et al.*, 2014). Similar to Word2Vec, GloVe differs fundamentally by using a count-based model rather than Word2Vec's predictive architecture. While predictive models like Word2Vec define word vectors by minimizing the loss between the target and prediction based on context words and their vector representations, count-based models like GloVe determine semantic relatedness by analyzing the statistical co-occurrence of words within the corpus.

Unlike Word2Vec, which relies solely on local context information, GloVe embeddings are trained using global co-occurrence data. However, the large word co-occurrence matrix used by GloVe necessitates a dimensionality reduction phase. This technique is well-suited for parallelization, making it easier to train on larger datasets. Although compressing representations might make them more robust, the ability to handle larger datasets offsets this potential drawback.

GloVe embeddings used in various studies are built from a vocabulary of over four hundred thousand words, trained on corpora such as Gigaword 5 and Wikipedia 2014, with 50 dimensions for word representation. Additionally, GloVe offers pre-trained embeddings with different dimensions (e.g., 100, 200, or 300), developed using even larger corpora like Twitter data.

**FastText**

One of the leading methods for static word embeddings is FastText, developed by Bojanowski *et al.*, 2017 at the Facebook AI Research lab. FastText addresses a key limitation of its predecessors by incorporating word morphology, which earlier models overlooked. Instead of assigning a distinct vector to each word, FastText represents each word using a bag-of-characters n-gram approach. For example, the word "house" with n = 3 would be represented as the sequences "ho", "hou", "ous", "use", and "se", along with the entire word.

FastText embeddings are trained using the skip-gram architecture. The final vector for a word is composed of the sum of its character n-grams. This approach allows FastText to create effective word embeddings for rare words by leveraging shared n-grams from more common words. Importantly, FastText can handle out-of-vocabulary (OOV) words as long as it has encountered the constituent n-grams during training, a capability lacking in both GloVe and Word2Vec.

Facebook has released pre-trained word vectors using FastText on Wikipedia, available in 294 languages.

### Generic Context word representation (Context2Vec)

This representation technique, introduced in Melamud *et al.*, 2016, is illustrated in Figure 4.4 in comparison to Word2Vec. The model employs a BiLSTM neural network to enhance word representations within a given context window. By training on a large text corpus, the neural network embeds words and their sentence contexts into the same low-dimensional space. This approach refines the model to capture the interactions between target words and their entire sentential context, providing a more robust and contextually aware representation.

### Contextualized word representations Vectors (CoVe)

Based on Context2Vec, the CoVe model was introduced in McCann *et al.*, 2017. Unlike GloVe (which uses matrix factorization) or Word2Vec (which employs skip-gram or CBOW), CoVe was developed using machine translation techniques. The authors began with GloVe word vectors and pre-trained a two-layer BiLSTM for an attention-based sequence-to-sequence translation task. They then combined this with GloVe vectors to create CoVe, using the output of the sequence encoder. This combined model was employed in downstream tasks using transfer learning. The authors demonstrated that incorporating these context vectors (CoVe) improved performance across various typical tasks, outperforming the use of unsupervised word and character vectors alone, as shown in tasks like SQuAD.

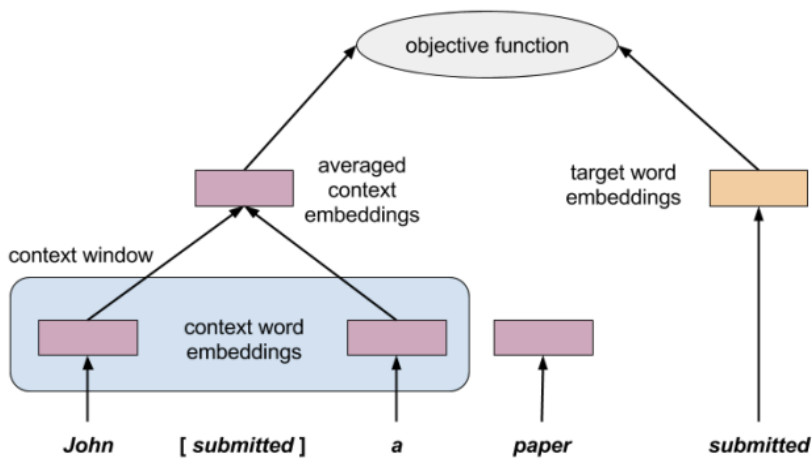

(a) word2vec *CBOW*

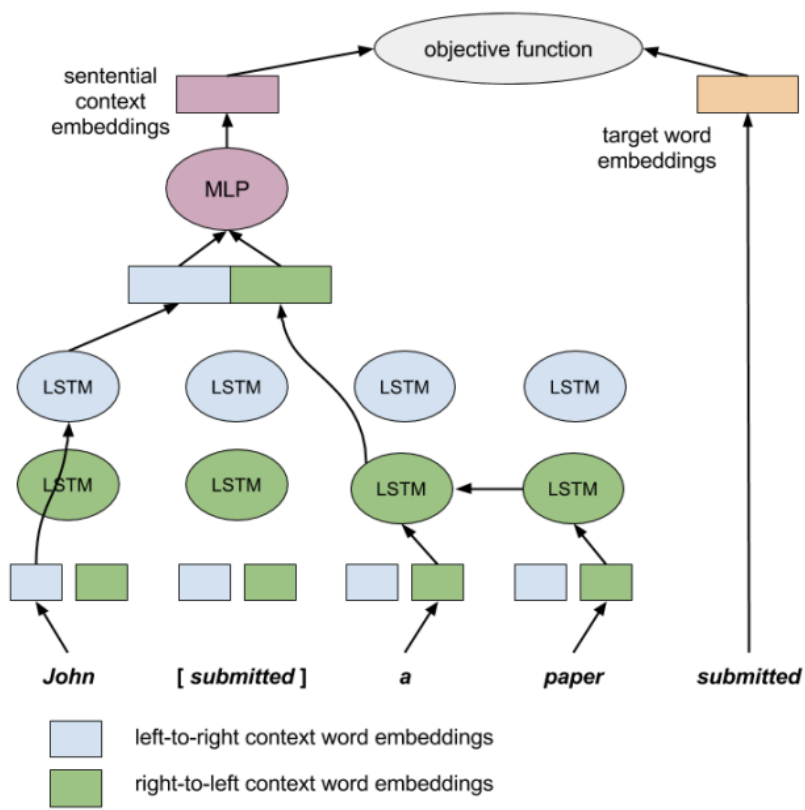

(b) *context2vec*

**Figure 4.4:** The original picture from the work on Context2Vec presented in Melamud *et al.*, 2016.

**Embedding from Language Models (ELMo)**

In Peters *et al.*, 1802, the authors introduce ELMo (Embeddings from Language Models), a novel contextual word representation method that captures both the complex aspects of word use, such as semantics and syntax and how these uses vary with the linguistic context (i.e., modelling polysemy). ELMo addresses the challenges of representing the flexible nature of word use in grammar and semantics, and how these uses adapt to different linguistic environments.

ELMo learns word embeddings from a bidirectional language model, processing text both forward and backwards. Unlike other contextual word representations that use only the final layer, ELMo concatenates the representations learned from all layers of the bidirectional language model. This allows ELMo to provide multiple embeddings for the same word in different contexts. Both the forward and backward language models in ELMo are trained using the log-likelihood of sentences. The final vector is computed by concatenating the hidden representations obtained from both directions. By incorporating ELMo, the authors achieve new state-of-the-art results across various tasks, with relative error reductions ranging from 6% to 20% over strong baseline models.

## 4.2   Analysis of a word embedding space

In this section, we present the results of a case study analyzing a word embedding trained from scratch from a previous work of ours (Siino *et al.*, 2022a). The methodology proposed here allows for a deeper investigation into the results and behaviour of a deep model trained on a specific dataset. Our analysis focuses on the FNS dataset to examine the performance and predictions of a simple CNN on the test set after training (Siino *et al.*, 2022a). This additional step can be integrated into the text classification pipeline to enhance model performance and gain a better understanding of its behaviour. However, the CNN-based model must capture more than just frequency differences, as suggested by its results. This section provides a post-hoc analysis of the word embedding layer. While hybrid approaches have been used to explain AI models Kenny *et al.*, 2021, the CNN tested here can be considered a shallow neural model. Therefore, it can be analyzed by mapping the outputs of each layer back to its inputs.

After training, we visualized two distinct clusters in the embedding projector, as shown in Figure 4.2. To understand how these clusters relate to the two classes, we labelled the words in the embedding space. We extracted 3959 keywords using a Bayesian model, specifically selecting the 1980 most frequent tokens from corpus 0 and 1979 most frequent tokens from corpus 1 and labelled them accordingly. We then visualized these keywords in the embedding space of the trained CNN model, as depicted in Figure 4.2b. Notably, we used key tokens retrieved by the Bayesian model rather than those from Sketch Engine because the former shares the same tokenization as the CNN model. We excluded tokens that appeared in both corpora.

Figure 4.2b confirms that the two clusters are closely related to the two task classes, with red dots representing FNS and blue dots representing nFNS. Exploring these clusters, we identified some keywords that were also highlighted using Sketch Engine Keywords. In Figures 4.2a and 4.2b, we highlighted *Unete*[1] as an FNS keyword and *bulos*[2] as an nFNS keyword.

It's important to note that the tokenization used by Sketch Engine differs from that of the CNN model. For instance, Sketch Engine distinguishes between cased and uncased letters, whereas the CNN model does not. Additionally, punctuation is always treated separately in the CNN model.

In the embedding space, we observed that tokens with higher keyness scores are positioned farther from the other cluster (e.g., *Unete* in Figure 4.2a). This suggests that tokens may be located according to their keyness scores within the embedding space.

## 4.3 Conclusion

In this chapter, we have explored various numerical representation methods for text, transitioning from traditional statistical models to more advanced word embedding techniques. We discussed the limitations

---

[1] In English: *join up.*
[2] In English: *hoaxes.*

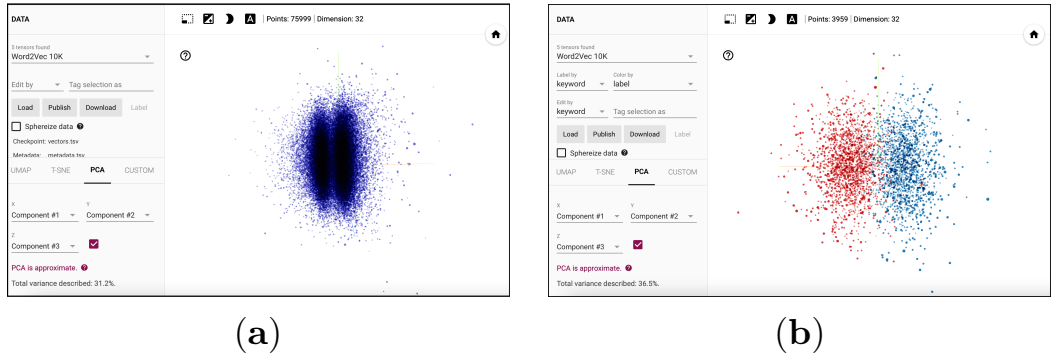

(**a**)                                              (**b**)

**Figure 4.5:** Word embedding as visualized in a 3-dimensional space. (**a**) Unlabeled word embedding space (75,999 points). (**b**) Labelled word embedding space (3959 points).

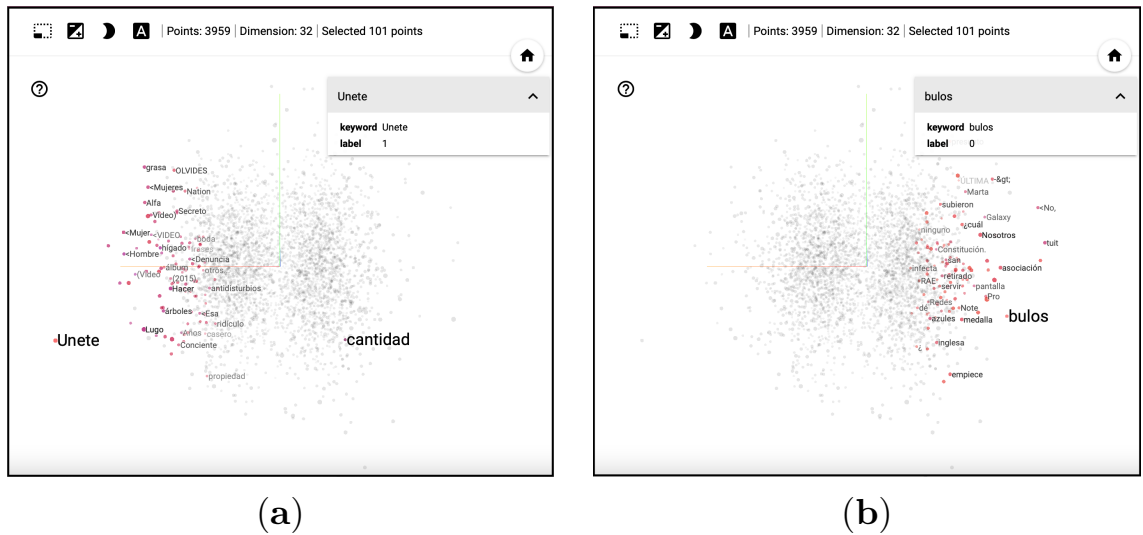

(**a**)                                              (**b**)

**Figure 4.6:** Visualization of FNS and nFNS keywords in the labelled embedding space. (**a**) Label 1. (**b**) Label 0.

of classical approaches, such as BoW and TF-IDF, and highlighted the advantages of word embeddings in capturing semantic relationships and contextual meaning. This progression reflects the evolution of NLP, where deep learning-based models have significantly enhanced text representation and understanding. These advancements lay the groundwork for more sophisticated machine learning and AI applications, enabling more accurate and nuanced language processing capabilities.

Then we reported an analysis from Siino *et al.*, 2022a to show that the deep model involved — a shallow CNN — can separate the vector spaces of word embeddings related to the two labels during the training phase. Notably, this ability of the deep model is highly task-dependent. When authors are strongly characterized by a specific vocabulary, the separability of classes can occur as early as the initial word embedding stage, rather than during convolution in subsequent layers. However, achieving this separability is not always feasible when training a word embedding layer from scratch. As the task varies, authors belonging to a class may not necessarily be characterized by certain keywords, or there may be an overlap between the point clouds in the word embedding space. Therefore, the methodology presented in this section could be valuable for analysing the embedding space after model training. Based on the results, one can evaluate whether it is necessary to introduce additional complexity into the model with successive layers to enhance classification performance.

# 5

# Classification

Text classification involves extracting features from raw text data and categorizing the text based on these features. Over the years, various text classification models have been developed, which can be grouped into three categories: Traditional Machine Learning-based Classifiers (TMLCs) deterministic models, Foundational Deep Learning Models (FDLMs), and Transformers.

Until recently, TMLCs were commonly used for text classification. These models use general-purpose classifiers that are not specifically designed for text interpretation. The text classification pipeline (Figure 1) includes steps to convert text into machine-interpretable features, partially addressing the unique challenges of textual data. One of the earliest models for text classification tasks was the Naive Bayes classifier. Other popular models include K-Nearest Neighbors (KNN), Support Vector Machines (SVM), Logistic Regression, and Random Forest. Recently, there has been debate over the performance of Light Gradient Boosting Machine (LightGBM) and Extreme Gradient Boosting (XGBoost).

For FDLMs, a Convolutional Neural Network (CNN) model was introduced in Kim, 2014 for text classification tasks. Other neural network architectures include artificial neural networks, Recurrent Neural

Networks (RNNs), and bidirectional Long Short-Term Memory networks (LSTMs).

Although not originally designed for text classification, the Bidirectional Encoder Representations from Transformers (BERT) and other Transformer-based architectures have been widely used in text classification models due to their success on various datasets. Other language models have also been employed as classifiers for text classification tasks. Here, we present some of the most common architectures used for text classification.

## 5.1 Traditional Machine Learning-based Classifiers (TMLCs)

Traditional Machine Learning-based Classifiers (TMLCs) speed up the text classification process without requiring initial pre-training, achieving significant results across various text classification tasks. In any TMLC, the first step is to preprocess the input text using techniques such as removing stop words, eliminating noise, and filtering out unwanted characters or strings (see Chapter 3). Following this, a representation model is selected to convert the text data into a numerical format, as discussed in Chapter 4.

This section briefly describes TMLCs. These methods rely on generic classification approaches and emphasize careful data pre-processing and feature engineering to achieve competitive results.

### 5.1.1 Logistic regression

*Logistic regression* (Genkin *et al.*, 2007) is one of the earliest and most notable classification techniques. As a linear classifier, logistic regression aims to predict probabilities over classes by identifying the most distinguishing features. Its basic formulation is particularly effective for binary classification tasks but can be extended to multinomial situations using the softmax function or by building an ensemble of binary classifiers with a one-vs.-rest strategy.

Linear classifiers like logistic regressors are well-suited for large and high-dimensional datasets. Logistic regression has been shown to outperform traditional back-off smoothing methods because it can han-dle unknown terms and avoids overestimating conditional probabilities that are originally zero. Ridge logistic regression is a popular approach for text classification, but its effectiveness for large-scale documents is debatable. To address this, sparse solutions are combined with ridge regression, removing less important features and solving the classical problem of ridge regressors (Pereira *et al.*, 2016).

Logistic regression is widely used in text classification for various tasks (Shah *et al.*, 2020). Despite its name, logistic regression is a linear classification model, also known as maximum-entropy classification, logit regression, or log-linear classifier. Logistic regression uses a logistic function to approximate the likelihood of possible outcomes. It is also employed in ensembles of text classifiers, as reported in Siino *et al.*, 2022c.

An implementation of logistic regression is available online via *sklearn*[1]. A common solver for this implementation is *lbfgs*, discussed in Byrd *et al.*, 1995.

### 5.1.2   Naïve Bayes

Naïve Bayes models are particularly popular due to their straightforward structure and ease of computation. The simplicity of Naïve Bayes comes from its assumption of independence, which posits that no feature influences any other feature. The core idea of the Naïve Bayes method is to use the prior probability of a class, as observed in the training set, to determine its posterior probability given the features.

Naïve Bayes classifiers are derived from Bayes theorem, which states that given the number of documents $n$ to be classified into $z$ classes where $z \in \{x_1, x_2, ...., x_z\}$ the predicted label out is $x \in X$. The Bayes theorem, which asserts that the predicted label out is $x \in X$, is the foundation for Naïve Bayes classifiers. Given the number of documents $n$ to be categorized into $z$ classes, where $z \in \{x_1, x_2, ...., x_z\}$, the expected label out is $x$ *in* $X$. This is how the Naïve Bayes theorem is formulated:

$$P(x|y) = P(x)\frac{P(y|x)}{P(y)} \tag{5.1}$$

---

[1]https://scikit-learn.org/stable/modules/generated/sklearn.linear__model.LogisticRegression.html

Where $y$ stands for a document and $x$ stands for the classes. The Naïve Bayes algorithm will, to put it simply, compute the likelihood that each word in the training data will be classified. Once each word's probability has been determined, the classifier is next instructed to categorize fresh data using the probabilities that had already been determined during the training phase.

The Naïve Bayes approach is straightforward and involves fewer parameters, making it less vulnerable to missing data. It assumes that features are independent of each other. However, Naïve Bayes's performance can decline when the number of features is high or there is a strong correlation between features. The Naïve Bayes method assumes that the conditions between texts are independent once the target value is given. It primarily uses the prior probability to determine the posterior probability. Naïve Bayes is widely used for text classification tasks due to its simplicity. Although the assumption of feature independence is sometimes incorrect, it significantly simplifies calculations and can improve performance.

Naïve Bayes has been widely used for large-scale document classification tasks since the 1950s, as noted by Porter, 1980. The Bayes theorem, developed by Thomas Bayes, serves as the theoretical foundation for the Naïve Bayes classifier approach. This method has garnered significant attention in recent studies (Qu *et al.*, 2018) and is commonly used in information retrieval.

Naïve Bayes for text classification employs generative models, which are the most frequently used approach. In its simplest form, Naïve Bayes counts the words in documents. The Naïve Bayes classifier is also considered a modern text classification application, as it is used in identifying fake news (Granik and Mesyura, 2017) and sentiment analysis (Mubarok *et al.*, 2017). Three popular Naïve Bayes methods for text classification are Bernoulli Naïve Bayes, Gaussian Naïve Bayes, and Multinomial Naïve Bayes.

As reported in McCallum and Nigam, 1998 and demonstrated experimentally over time through various text classification tasks Raschka, 2014, Naïve Bayes is one of the most effective models for classification. A popular multinomial Naïve Bayes classifier from *sklearn* is the Multi-nomialNB implementation[2]. When dealing with multinomial distributed data, MultinomialNB implements the Naïve Bayes method. Data are commonly expressed as word vector counts.

### 5.1.3   K-NN-Based Classification

Text classification using K-Nearest Neighbors (k-NN) algorithms (Cover and Hart, 1967) approaches the problem by locating the k-most similar labeled instances and, in its basic form, assigning the most prevalent category to the unlabeled instance being classified.

Unlike methods that use a discriminating class domain to determine the category, k-NN relies on nearby finite neighboring samples. This makes it better suited for datasets with greater class overlap or inter-mixing. The k-NN algorithm identifies the k documents in the training set that are closest to a test document $x$, and then ranks the category choices based on the classifications of these k neighbors. The category score of the neighbor documents may depend on how closely $x$ resembles each neighboring document. If multiple k-NN documents fall under the same category, the similarity score of that class with respect to the test document $x$ is calculated by summing these scores. The test document $x$ is then assigned to the class with the highest score.

However, the k-NN approach can be time-consuming on large-scale datasets due to the positive association between model time/space complexity and data volume (Jiang *et al.*, 2012). To address this, scholars in Soucy and Mineau, 2001 propose a k-NN technique without feature weighting to reduce the number of selected features. By employing feature selection, this method can identify relevant features and create word interdependencies.

k-NN typically classifies samples better when there is more data, but it can struggle with extremely asymmetric data distributions. To enhance classification performance on unbalanced corpora, the Neighbor-Weighted K-Nearest Neighbor (NWKNN) (Tan, 2005) is introduced. This method assigns larger weights to neighbors in narrow classes and smaller weights to neighbors in broader classes.

---

[2]https://scikit-learn.org/stable/modules/generated/sklearn.NaÃŕve_bayes.MultinomialNB.html

### 5.1.4 Decision tree

*Decision trees* were introduced in Quinlan, 1986 and further detailed in Magerman, 1995. They are one of the oldest classification models for text and data mining, successfully used in various fields. The primary motivation behind decision trees is to build tree-based attributes for data points, with the key question being which feature should be at the child level and which should be the parent feature.

The decision tree consists of a root node, decision nodes, and leaf nodes, which represent the dataset, execute computations, and perform classification, respectively. During the training phase, the classifier learns the decisions needed to divide labeled groups. To classify an unlabeled instance, the data is processed through the tree. At each decision node, a specific property of the incoming text is compared to a threshold learned during training. The choice is based on whether the selected feature is more or less prominent than the threshold, dividing the tree into two parts. The text traverses these decision nodes until it reaches a leaf node, which describes the class to which it is assigned.

The benefits of the decision trees include minimal hyperparameter tuning, simplicity in description, and ease of understanding its visualizations. However, it has significant drawbacks, such as the risk of overfitting, sensitivity to small changes in the data, and difficulties with predictions outside the training samples. The decision trees produce simple classification rules, and pruning techniques (Rastogi and Shim, 2000) can help mitigate the impact of noise. However, its fundamental weakness is its inability to handle rapidly growing datasets effectively. The Iterative Dichotomiser 3 (ID3) algorithm (Quinlan, 1986) uses information gain as the attribute selection criterion for each node, choosing the attribute with the highest information gain value as the discriminant for the current node.

In Johnson *et al.*, 2002, the author proposes a decision tree-based symbolic rule system. This approach converts each text into a vector based on word frequency and generates rules from the training data. Additional data, similar to the training data, is classified using these learned rules. The Fast Decision Tree (FDT) (Vateekul and Kubat, 2009) employs a two-pronged approach to reduce the computational costs of decision tree algorithms: pre-selecting a feature set and training multiple decision trees on various data subsets. For imbalanced classes, the results from different decision trees are integrated using a data-fusion technique.

### 5.1.5   Random forest

*Random forest*, also known as an ensemble learning methodology, combines the outcomes of multiple trained models to create a more robust classifier with better performance than a single model.

A random forest, described in Ho, 1998, is easy to learn and produces improved classification outcomes. Each tree in the random forest is trained on a bootstrapped subset of the training text. At each decision node, a random subset of features is selected, and the model considers only a portion of these attributes.

The main issue with using a single decision tree is its high variability, which makes it sensitive to the organization of the training data and feature arrangements. Although the random forest is quick to train on textual data, Bansal *et al.*, 2018 noted that it can be slow to make predictions after training. Random forest performs well with both categorical and continuous data, can handle missing values automatically, is robust to outliers, and is less affected by noise. However, training numerous trees can be computationally expensive, time-consuming, and memory-intensive.

### 5.1.6   Support Vector Machines (SVMs)

Authors in Cortes and Vapnik, 1995 introduced the Support Vector Machine (SVM) for binary classification in pattern recognition. For the first time, authors in Joachims, 1998 represented each text as a vector and applied the SVM algorithm for text classification. SVM-based methods divide text classification challenges into numerous binary classification tasks. By maximizing the distance between the hyperplane and the two categories of training sets, SVM creates an optimal hyperplane in the input space or feature space, resulting in the best generalization ability.

The objective is to maximize the perpendicular distance along the category boundary, which minimizes the classification error rate. The problem of building an optimal hyperplane can be formulated as a quadratic programming problem to achieve a globally optimal solution. To enable SVM to handle nonlinear problems and become a reliable nonlinear classifier, selecting the appropriate kernel function is crucial (Leslie *et al.*, 2001; Taira and Haruno, 1999).

To further reduce the labeling effort based on the supervised learning algorithm SVM, active learning (Li and Guo, 2013) and adaptive learning (Peng *et al.*, 2008) methods are employed for text classification. Joachims, 2002 proposes a theoretical learning model that combines the statistical traits with the generalization performance of an SVM, analyzing the features and benefits using a quantitative approach. This analysis examines what the SVM algorithms learn and identifies suitable tasks.

The Transductive Support Vector Machine (TSVM) (Joachims, 1999) introduces a universal decision function that considers a specific test set to reduce misclassifications of particular test collections. It establishes a better framework and learns more quickly by utilizing existing knowledge.

SVMs extend to multidimensional, non-linear classification by projecting their inputs into a higher-dimensional space to better distinguish training categories. This process is known as the kernel trick, where the function mapping to this higher-dimensional space is called a kernel function. The key to achieving good performance is choosing the proper form and parameters for the kernel function.

As reported in Colas and Brazdil, 2006 and in Liu *et al.*, 2010, classifiers based on SVM are well-established methods for text classification tasks. SVM are also employed in ensemble-based text classifiers, as reported in Croce *et al.*, 2022. Thanks to SVM models, classification results compared to other classification methods improved. Based on Chang and Lin, 2011, is available online the *sklearn* SVC implementation[3].

---

[3]https://scikit-learn.org/stable/modules/generated/sklearn.svm.SVC.html

## 5.2   Foundational Deep Learning Models (FDLMs)

The Artificial Neural Networks (ANN) that make up the FDLMs mimic the human brain to automatically learn high-level features from data, outperforming conventional models in speech recognition, picture processing, and text understanding. To categorize the data, input datasets like single-label, multi-label, unsupervised, and imbalanced datasets should be examined. The input word vectors are delivered into the ANN for training following the trait of the dataset up until the termination condition is met. The downstream tasks, such as sentiment categorization, question answering, and event prediction, provide as proof of the training model's effectiveness. In recent decades, a large number of deep learning models for text classification have been suggested. The first two deep learning methods for the text classification task that outperform conventional models are the multilayer perceptron and the recursive neural network. Then, for text categorization, Convolutional Neural Networks (CNNs), Recurrent Neural Networks (RNNs), and attention processes are applied. Many researchers enhance CNN, RNN, and attention, or model fusion and multitask approaches, to improve text classification performance for various tasks. Text categorization and other NLP methods have advanced significantly with the introduction of BERT, which can produce contextualized word vectors. It has been found that text classification models based on BERT perform better than the models mentioned above in a variety of NLP tasks, including text classification. Additionally, Graph Neural Network (GNN)-based text classification technology is being studied by certain academics in order to collect structural information in the text that cannot be captured by alternative techniques.

### 5.2.1   Artificial Neural Network (ANN)

The gap between shallow and deep methodologies is bridged by straightforward structures like Multilayer Perceptrons (MLPs) or ANN. These neural network designs are among the most fundamental, but they serve as the cornerstone for the first word embedding methods and produce great results when used as standalone classifiers. These MLP models often approach input text as an unordered BoW, with each input word being represented by a different feature extraction method (like TF-IDF or word embeddings).

ANN see the text as a collection of BoW. They first use an embedding model, such as Word2Vec (Mikolov *et al.*, 2013a) or Glove (Pennington *et al.*, 2014), to learn a vector representation for each word. They then use the vector sum or average of the embeddings as the representation of the text, pass it through one or more feed-forward layers known as Multi-Layer Perceptrons (MLPs), and perform classification on the representation of the final layer using a classifier, such as The Deep Average Network (DAN) (Iyyer *et al.*, 2015) that is one of these models.

DAN performs better than other more complex models that are intended to explicitly learn the compositionality of texts, despite their simplicity. On datasets with large syntactic variance, DAN, for instance, performs better than syntactic models. A straightforward and effective text classifier named fastText is proposed by the authors in Joulin *et al.*, 2016. FastText sees text as a collection of words, much like DAN. FastText, unlike DAN, uses a bag of n-grams as extra features to record local word order data. In practice, this proves to be quite effective, producing outcomes that are comparable to those obtained by methods that explicitly employ the order of the words (Wang and Manning, 2012).

Additionally, the authors of Le and Mikolov, 2014 propose doc2vec, which uses an unsupervised approach to train fixed-length feature representations of variable-length textual units like sentences, paragraphs, and documents. Doc2vec's architecture resembles that of the CBOW model. The extra paragraph token that is via matrix converted to a paragraph vector is the only difference. To forecast the fourth word in doc2vec, this vector's concatenation or average with a context of three words is employed. The paragraph vector serves as a placeholder for context-missing data and can serve as a reminder of the paragraph's subject. After training, the paragraph vector is sent to a classifier for prediction and utilized as features for the paragraph (for example, in place of or in addition to BoW). When Doc2vec was released, it produced brand-state-of-the-art outcomes on several text classification tasks.

### 5.2.2 Recurrent Neural Networks (RNNs)

RNNs (Pouyanfar *et al.*, 2018)—which are designed to get word relationships and text structures for TC—view text as a series of words. Pure RNN models, on the other hand, frequently perform worse than feed-forward neural networks. Long Short-Term Memory (LSTM) is the most often used RNN variation, since it is intended to better capture long-term dependency. By incorporating a memory cell to retain values over virtually any period and three gates (input gate, output gate, forget gate) to control the flow of data into and out of the cell, LSTM solves the gradient disappearing or exploding issues that plague vanilla RNNs. There have been efforts to make RNNs and LSTM models for text classification better by capturing additional data, such as natural language tree structures, long-span word relations in text, document topics, and so forth. The authors of Nowak *et al.*, 2017 describe how to conduct text classification using LSTM networks and various variations, such as BiLSTM and GRU. Additionally, authors who employ a BiLSTM in Siino *et al.*, 2022b do so with noteworthy outcomes. Two bidirectional LSTM layers make up the model.

The authors in Tai *et al.*, 2015 develop a Tree-LSTM model, a generalization of LSTM to tree-structured network typologies, to learn complicated semantic representations. Because natural language possesses syntactic characteristics that would naturally join words to form phrases, the authors contend that Tree-LSTM is a more effective model for NLP tasks than the chain-structured LSTM. On the two tasks of sentiment classification and predicting the semantic similarity of two sentences, they validate the efficiency of Tree-LSTM. The chain-structured LSTM is also extended to tree structures by the authors of Zhu *et al.*, 2015, using a memory cell to preserve the history of numerous child cells or numerous descendant cells in a recursive process. The new model, they contend, offers a systematic approach to thinking about long-distance communication over hierarchies, such as language or picture parse structures. The LSTM architecture is supplemented in Cheng *et al.*, 2016 with a memory network in place of a single memory cell to model long-span word relations for machine reading. With brain attention, this permits adaptive memory use during recurrence and provides a method for weakly inducing relationships between tokens. In terms of language modelling, sentiment analysis, and NLI, this model yields encouraging results. By capturing important information with various timescales, the Multi-Timescale LSTM (MT-LSTM) neural network, which is described in Liu *et al.*, 2015, is also intended to model extended texts, such as sentences and papers. A typical LSTM model's hidden states are divided into many categories by MT-LSTM. At various times, each group is updated and activated. MT-LSTM can therefore model extremely long documents. On text classification, MT-LSTM is said to perform better than several baselines, including models based on LSTM and RNN. RNNs have trouble remembering long-distance dependencies, but they do a decent job of capturing the local structure of a word sequence. Contrarily, word order is not taken into account by latent topic models, which can only represent the overall semantic structure of a document. The authors of Dieng *et al.*, 2017 suggest a TopicRNN model to combine the advantages of latent topic models and RNNs. It uses latent topics to capture global (semantic) dependencies while employing RNNs to capture local (syntactic) dependencies. According to reports, TopicRNN performs better in sentiment analysis than RNN baselines. Other intriguing RNN-based models exist. The authors of Liu *et al.*, 2016 train RNNs to utilize labelled training data from numerous related tasks by utilizing multitask learning. The authors of Johnson and Zhang, 2016 investigate an LSTM-based text region embedding technique. Authors in Zhou *et al.*, 2016 present a novel architecture that combines a BiLSTM model with two-dimensional max-pooling to capture text features. A bilateral multi-perspective matching model is put out in Wang *et al.*, 2017 inside the "matching-aggregation" framework. A BiLSTM model is used by the authors of Wan *et al.*, 2016 to investigate semantic matching utilizing various positional sentence representations. It is crucial to remember that RNNs are a subset of DNNs. A recursive neural network continually applies the same set of weights over a structural input to create a structured prediction or a vector representation over inputs of varying sizes. Recursive neural networks (RNNs) are recursive neural networks with a linear chain structure input, whereas recursive neural networks with a hierarchical structure input, such as parse trees of English language sentences (Socher *et al.*, 2013), can operate on hierarchical structures by integrating child representations into parent representations. RNNs are the most popular recursive neural networks for text classification because of their effectiveness and ease of use.

### 5.2.3 Convolutional Neural Networks (CNNs)

Computer vision applications are frequently linked with CNNs. CNNs are employed for classifying images using convolving filters that extract picture characteristics. However, they have also been used, especially in the context of NLP and text classification. In Kim, 2014, one of the earliest attempts to use a CNN for sentiment analysis is covered. Figure 5.1 shows the original network structure. The author describes a series of experiments using a CNN trained for sentence-level classification tasks on top of pre-trained word vectors. The author demonstrates that a straightforward CNN with little hyperparameter adjustment and static vectors performs admirably on a variety of benchmarks. Additional performance benefits can be obtained by learning task-specific vectors through fine-tuning. To support the use of both task-specific and static vectors, the author also suggests a straightforward change to the architecture. The CNN models mentioned here outperform the current state of the art on 4 of the 7 tasks, including sentiment analysis and question classification.

The CNN architecture used in Siino *et al.*, 2022a to identify FNS on Twitter is displayed in Figure 5.2. The input text's vectors are first combined into a word embedding matrix. The convolutional layer, which has multiple filters with various dimensions, feds the matrix. The output of the convolutional layers is then passed through the pooling layer and concatenated to create the final vector representation of the text for two additional pairs of conv-pool layers. The last vector predicts the category. To avoid overfitting, certain dropout layers are placed between layers. It is worth noting that in the same study, the authors analyse each layer's behaviour and output, motivating their choices and providing insights related to the network behaviour.

Examining their input, which likewise uses word embeddings, is the simplest way to comprehend these methods. RNNs typically input a sentence's words in order, but CNNs provide sentences as a matrix, with each row representing an embedding of a word (therefore, the number of columns corresponds to the size of the embeddings). Contrary to RNN, CNN can apply convolutions defined by many kernels to numerous chunks of a sequence at once. In contrast, convolutional filters often glide over local portions of an image in two directions in image-based tasks. Instead, filters in text-related tasks are typically made to be as wide as the embedding size, ensuring that this operation only proceeds in ways that make sense from a sentence-level perspective while always taking the full embedding for each word into account. In general, the speed and effectiveness of CNNs' latent representations are considered to be their key benefits. On the other hand, when analyzing text, other features that could be used while working with images, like location invariance and local compositionality, make little sense.

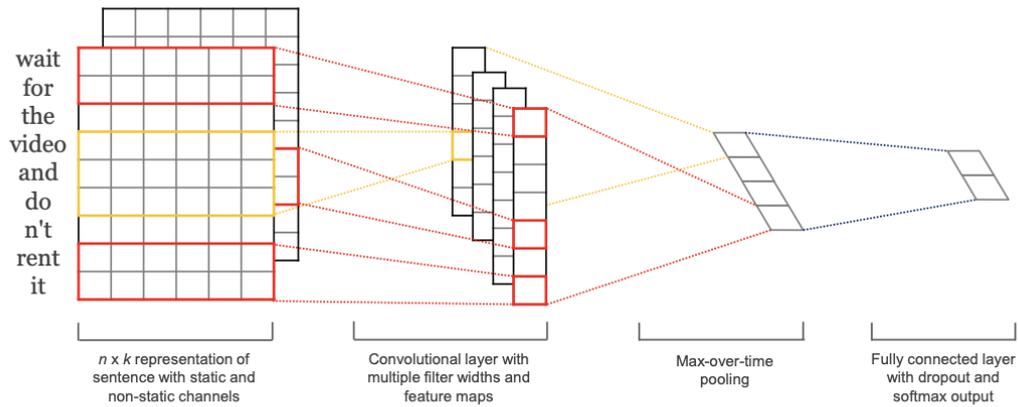

Figure 1: Model architecture with two channels for an example sentence.

**Figure 5.1:** The original image of the CNN architecture proposed in Kim, 2014.

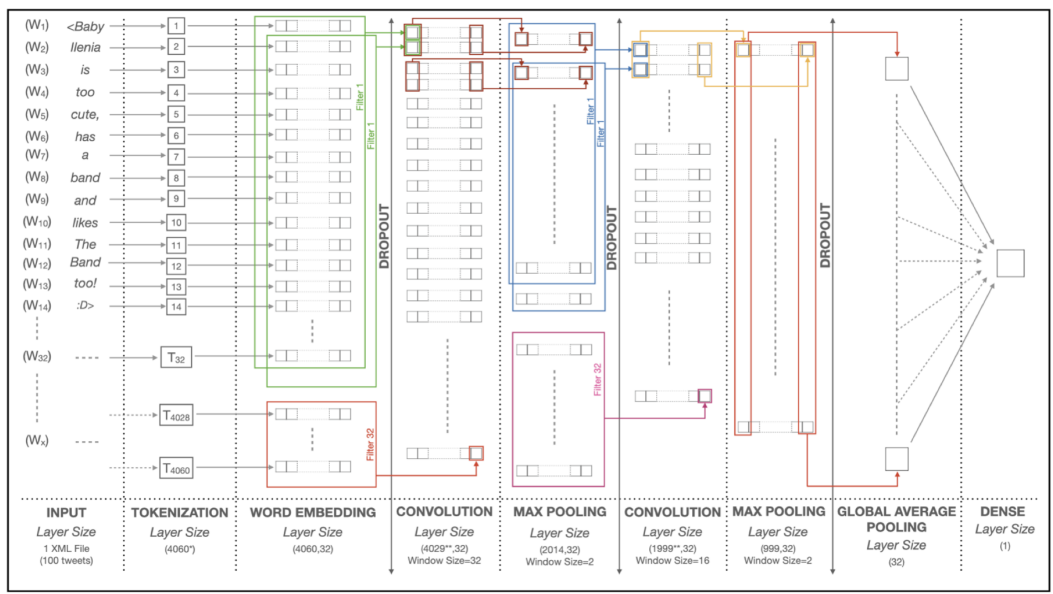

**Figure 5.2:** The architecture of the CNN used proposed in Siino *et al.*, 2022a.

Other interesting applications based on CNN are discussed in Siino *et al.*, 2021 and also used in Mangione *et al.*, 2022. Such CNNs consist essentially of a single convolutional layer. As demonstrated by its results, these CNNs outperforms Transformers and others proposed models as stated in Rangel *et al.*, 2021.

### 5.2.4   Capsule Neural Networks

CNNs employ pooling and multiple layers of convolution to classify images and words. While pooling helps identify key features and simplify computation, convolution can lose spatial relationship information, leading to misclassifications based on orientation or proportion.

To address these pooling issues, Hinton *et al.*, 2011 introduced capsule networks (CapsNets). A capsule is a group of neurons that represents various properties of an entity, such as an object or its components, through an activity vector. The vector's length indicates the likelihood of the entity's existence, and its orientation represents the entity's characteristics.

Unlike CNNs' max-pooling, which selects and discards information, capsules use all network data up to the final layer for classification. This is done by "routing" each lower-layer capsule to its ideal parent capsule in the higher layer. Methods like dynamic routing-by-agreement (Sabour *et al.*, 2017) or the EM algorithm (Hinton *et al.*, 2018) can implement this routing.

Capsule networks have been recently applied to text classification, represent a sentence or document as a vector using capsules. The authors of Yang *et al.*, 2018 propose a text classification model based on a variation of CapsNets. This model consists of four layers: an n-gram convolutional layer, a capsule layer, a convolutional capsule layer, and a fully connected capsule layer.

To stabilize the dynamic routing process and minimize disruption from noise capsules (which contain background data like stop words or irrelevant words), the authors test three methods. They also explore two capsule structures: Capsule-A and Capsule-B. Capsule-A is similar to the CapsNet in Sabour *et al.*, 2017. Capsule-B, on the other hand, uses three parallel networks with filters of different window sizes in the n-gram convolutional layer to learn a more comprehensive text representation. In the experiments, Capsule-B performs better.

### 5.2.5 Graph Neural Networks

Graphs are highly useful in social networks and text classification because they effectively represent relationships and structures within data. In social networks, graphs model user interactions, connections, and influence, enabling community detection and recommendations (Senette *et al.*, 2024; Siino *et al.*, 2020). In text classification, graphs help analyse word co-occurrences, document relationships, and semantic connections, improving tasks like topic modelling, sentiment analysis, and information retrieval. Their ability to capture complex relationships makes them essential for enhancing accuracy and efficiency in these domains.

TextRank (Mihalcea and Tarau, 2004) is one of the earliest graph-based models developed for NLP. It represents a natural language document as a graph with nodes and edges. Nodes can represent various text units, such as words or complete sentences, depending on the application. Edges can capture lexical or semantic relationships, contextual overlap, or other types of relationships between nodes.

Modern Graph Neural Networks (GNNs) extend deep learning meth-

ods for graph data, similar to the text graphs used by *TextRank*. Over the past few years, various Deep Neural Networks (DNNs), including CNNs, RNNs, and autoencoders, have been adapted to handle the complexity of graph data.

For example, to perform graph convolutions, a 2D convolution of CNNs for image processing is generalized by taking the weighted average of a node's neighbourhood information. Convolutional GNNs, such as Graph Convolutional Networks (GCNs) (Kipf and Welling, 2017) and their derivatives, are commonly used due to their effectiveness and ease of integration with other neural networks, achieving state-of-the-art results in many applications. GCNs are an effective CNN variation for graphs, stacking layers of learned first-order spectrum filters and applying a nonlinear activation function to learn graph representations. Text classification is a common application of GNNs in NLP, where the relationships between words or documents are used to infer document labels.

Peng *et al.*, 2018 propose a graph-CNN-based model that first converts text into a graph of words and then uses graph convolution procedures to process the word graph. Their experiments show that CNN models can learn multiple levels of semantics, while the graph-of-words representation captures non-consecutive and long-distance semantics.

Peng *et al.*, 2019 present a text classification model based on hierarchical taxonomy-aware and attentional graph capsule CNNs. A distinctive feature of this model is its use of hierarchical relationships among class labels, which were previously considered independent. The authors introduce a novel weighted margin loss that considers label representation similarity and develop a hierarchical taxonomy embedding approach to train their representations.

A similar Graph CNN (GCNN) model for text classification is proposed in Yao *et al.*, 2019. The authors create a single text graph for a corpus based on word co-occurrence and document-word relations and then train a Text Graph Convolutional Network (Text GCN) for the corpus. The Text GCN learns word and document embeddings jointly, supervised by the known class labels for documents, starting with a one-hot representation of each.

Finally, another interesting application is in Lomonaco *et al.*, 2022 where the introduced model leverages ELECTRA-based document embedding and a text graph processed using a GCN. The goal is to identify harmful tweets (i.e., predict whether a tweet is harmful and why). The authors introduce a novel method capable of handling various types of heterogeneous textual or social information. The authors demonstrate the performance of an initial version of this model on the task, highlighting areas for future improvement.

## 5.3 Transformers

In this section, we present the two major classes of Transformer-based architectures: the Large Language Models (LLMs) and the Generative Pretrained Transformers (GPTs). Both LLMs and GPTs have revolutionized the field of natural language processing by enabling a wide range of sophisticated applications, from text generation to sentiment analysis. We delve into the workings of these architectures, highlighting their unique attributes and shared principles. While the LLMs are designed primarily for understanding and generating not necessarily text as output, they excel in tasks that require contextual comprehension and coherence over longer sequences. On the other hand, GPTs are specifically tailored for generative purposes, leveraging their autoregressive nature to produce human-like text based on given prompts. This distinction is crucial as it determines the choice of model based on the intended application. The first LLMs (e.g., BERT-based) were mainly built, making use of the encoder part of the Transformer architecture. In this way, the output was usually a contextual representation of the input text, capturing semantic nuances and allowing for the effective extraction of text features. On the top of such type of architecture is usually applied a final dense layer that, based on the addressed task, would eventually produce a single class or multiclass response Siino *et al.*, 2022b; Siino and Tinnirello, 2023; Siino *et al.*, 2022a. On the other hand, generative Transformers (e.g., GPT-based) leverage the decoder component, enabling the model to generate coherent and contextually relevant text sequences. This generative capability opens up a wide array of applications, from creative writing and automated content gen-eration to more sophisticated uses in dialogue systems and interactive storytelling, or code synthesis. The encoder part is shown in the left part of the Figure 5.3 while the decoder part is shown on the right side.

### 5.3.1    The architecture

The most fundamental form of language modelling involves predicting the next word in a sentence by estimating the probability of a word given its preceding or following context. Despite predating neural networks, language models have been instrumental in numerous modern deep learning advancements. Early language models included n-gram models, which assign probabilities to word sequences (i.e., sentences). A well-structured sentence typically receives a higher score, although the specific interpretation of this probability depends on the task, such as improved translation.

While the primary goal is to predict the likelihood of the next word, the task is often framed as assigning probabilities to entire sentences. These models typically rely on the Markov assumption, which posits that the likelihood of the next word depends only on the $k$ preceding words. Future advancements in this field are expected to leverage the Transformer architecture (Vaswani *et al.*, 2017), which has proven to be faster and more efficient for language modelling compared to LSTMs or CNNs. Although Transformers will be discussed in more detail later, they are briefly introduced here as language representation models.

Encoder-decoder structures are common in competitive neuronal sequence transduction models. The model is autoregressive at each phase, using the previous symbols as extra input to construct the next. Transformers' encoder converts an input series of symbol representations (x1,..., xn) into an equivalent sequence of continuous representations, z = (z1, ..., zn). Then the decoder produces a sequence (y1,..., ym) of symbols, starting with z. Following its general architecture, the Transformer uses layered self-attention and point-wise, entirely connected layers for the encoder and decoder. The general architecture of a Transformer is depicted in Figure 5.3, as presented in the original work Vaswani *et al.*, 2017.

For downstream tasks, Transformer-based architectures typically

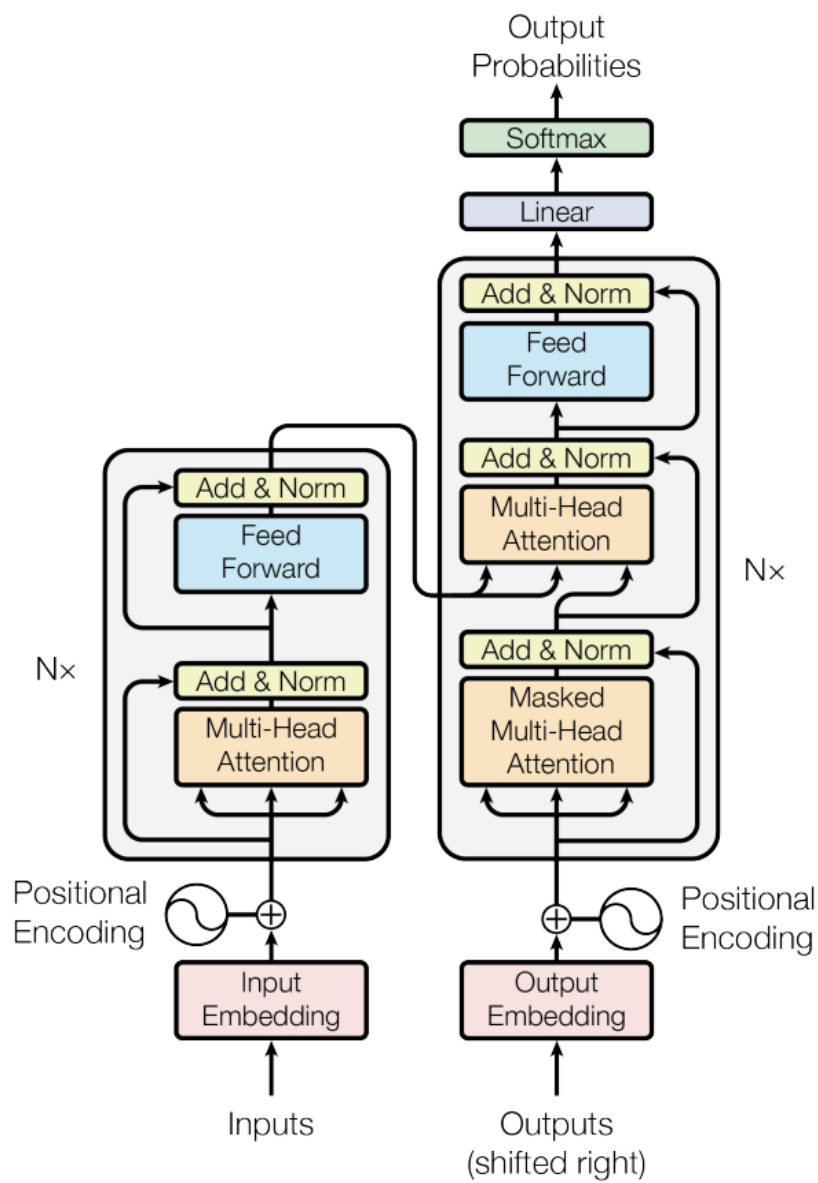

**Figure 5.3:** The original picture of a Transformer from Vaswani *et al.*, 2017.

follow these steps:

1. *General language model pre-training*. This phase involves unsupervised learning on large, unlabelled text datasets, allowing the model to capture broad linguistic patterns.

2. *Target task language model fine-tuning*. The pre-trained language model is then fine-tuned on a specific task using labelled data, adapting it to the nuances of the target task.

The pre-training phase is unsupervised and can leverage vast amounts of unlabeled text data, making it as comprehensive as possible. During the pre-training, the common objective functions used are: The commonly used objective functions during pre-training include Masked Language Modelling (MLM) and Next Sentence Prediction (NSP). MLM enables the model to predict missing words in a sentence, enhancing its understanding of context and semantics. NSP, on the other hand, focuses on predicting the relationship between sentence pairs, which aids in understanding how different sentences relate to each other in a given context. Together, these objectives equip the model with a robust understanding of language structure and meaning. This foundational knowledge is crucial for downstream tasks such as sentiment analysis, text summarization, and question-answering, where a nuanced grasp of language is required. The Transformer-based models discussed here represent the current state-of-the-art, and while incremental improvements are still possible, creating significantly better architectures remains challenging. These models excel at handling context-related problems but are often trained on general domain corpora like Wikipedia, limiting their applicability to specific tasks or domains. There is a hypothesis that domain-specific Transformer-based models could enhance performance in specialized subdomains.

**RNN encoder–decoder**

Sequence transduction methods have traditionally been dominated by networks with RNN-like designs. Researchers began pushing the boundaries of text classification using RNN-based encoder-decoder architectures and recurrent language models, which are advancements over traditional word embedding methods.

To better understand Transformers, consider a translation task where the input sequence is a sentence in a source language, and the output sequence is its translation in another language. In an RNN-based approach, each word in the input sequence is processed sequentially by the encoder. At each time step $t$, the model receives the new input word and the hidden state from the previous time step $t - 1$. Theoretically, RNNs should be able to learn both short- and long-term associations between words due to this step-by-step processing. The encoder's output, known as the "context," is a compressed representation of the input sequence.

Following this, the decoder evaluates the context and generates a new sequence of words (e.g., a translation into a different language), where each word depends on the results of the preceding time step. The context, which contains contextually significant information, is latently recorded during encoding and can later be utilized for tasks like text classification. However, a major drawback of this approach is that the encoder must compress all relevant information into a fixed-length vector.

This compression becomes problematic, especially for longer sentences, as the performance of basic encoder-decoder models rapidly degrades with increasing input sentence length. Additionally, recurrent models have inherent limitations due to their sequential nature. Parallelization is impossible, leading to more complex computations. Longer sentences pose a true bottleneck for RNNs, often causing memory issues due to the network's tendency to forget earlier parts of the sequence (primarily due to the vanishing gradient problem).

The attention mechanism was introduced to address the drawbacks of recurrent architectures. Incorporating attention mechanisms marked a significant turning point in NLP, eventually becoming a fundamental component of the Transformer architecture. Unlike LSTM-based models, which showed little benefit from significant size increases, the depth of Transformer models has proven to be highly advantageous for their performance.

## The attention mechanism

The attention mechanism was initially introduced to enhance the learning process by focusing on the more significant components of input phrases, essentially allowing the model to "pay attention" to crucial elements. Traditionally, encoder-decoder designs based on RNNs have been used to address sequence-to-sequence (seq2seq) problems, employing stacked RNN layers for both the encoder and decoder.

Bahdanau *et al.*, 2015 introduced the concept of attention to tackling issues in neural machine translation tasks. The authors proposed that the decoder could distinguish between input words and identify which are essential for generating the next target word by leveraging knowledge of the entire input sequence. The attention mechanism relies on the encoder's hidden state (also known as "annotation") to enhance the input context for each decoder unit, which contains information about the entire input sequence. This specific technique is referred to as "*additive attention*". While there are various ways to integrate the attention mechanism into seq2seq architectures, the primary goal is to create an alignment score that measures the relative importance of words in the input and output sequences. Beyond NLP, where attention first proved its value, attentive artificial neural networks are now applied in numerous domains.

In the field of text classification, hierarchical attention networks (Miculicich *et al.*, 2018; Yang *et al.*, 2016) serve as innovative examples. These methods operate at two levels: the word level, when encoding document phrases, and the sentence level, when encoding the significance of each sentence relative to the intended sequence. However, attention has evolved from being just an additional augmentation to serving as a foundational component. This evolution is exemplified in the Transformer architecture, which retains the familiar encoder-decoder structure but eschews recursion. Instead, dependencies between input and output are established solely through the attention mechanism. Transformers have demonstrated superior performance and significantly faster processing speeds due to their high degree of parallelization.

**The Transformer architecture**

Vaswani *et al.*, 2017 introduced the Transformer architecture, an advanced encoder-decoder model that processes all input tokens (such as words) simultaneously rather than sequentially. Transformers treat input sequences as a bag of tokens, disregarding the order. To understand the relationships between tokens, the Transformer employs a mechanism called "self-attention." Through a specific encoding phase before the encoder's first layer, the same word appearing in different positions within a sentence will have distinct representations.

Positional encoding is used to preserve information about the relative positions of words, which would otherwise be lost. The self-attention layer, a key component of this architecture, allows the encoder to consider other words in the input sentence as it processes each word. Multiple self-attention layers are stacked to form a multi-head attention layer. The outputs of these heads are concatenated and passed through a linear layer to combine them into a single matrix.

The Transformer's multi-head self-attention layer performs multiple parallel iterations of these processes to expand the range of representation sub-spaces the model can focus on. The outputs of the attention heads are concatenated, passed through a linear layer to form the final representation, which integrates information from all attention heads. This representation is then normalized, added to the residual input, and fed into a feed-forward linear layer.

Transformers significantly enhance text text classification and other NLP tasks by efficiently learning global semantic representations. They often use unsupervised techniques to autonomously extract semantic knowledge and create pre-training targets to help machines understand semantics. Up to date the representation provided by these models not only improves performance on benchmark datasets but also offers insights into the underlying linguistic structures.

### 5.3.2 Large Language Models (LLMs)

Also for text classification, the attention-based techniques are successfully applied. The model can pay varying attention to different inputs thanks to the attention mechanism. It first groups necessary words into sentence vectors, and then groups necessary sentence vectors into text vectors. Through the two levels of attention, it can determine the relative contributions of each word and sentence to the classification judgment, which is useful for applications and analysis. An example of application is show in the Figure 5.4. The task is a binary classification problem. News is provided as input sentence to a BERT model. After obtaining the output (latent word representation of the input text), this is passed to a Dense Layer made of two units, corresponding to the two possible class (i.e., fake news or non-fake news) to detect fake news. It is worth mentioning that at least three different fine-tuning strategies can be applied to this scenario. They are:

- **Fine-tuning the weights of the whole architecture**. In this case, either the already-trained weights of the BERT model and the weights of the added Dense Layer are adjusted to the specific dataset related to the fake news detection task.

- **Fine-tuning the weights of the Dense Layer**. In this case, only the weights of the added Dense Layer are adjusted to the specific dataset related to the fake news detection task. The already trained weights of the BERT model are frozen during the fine-tuning.

- **Fine-tuning chosen weights of the whole architecture**. Freeze only specific layers of the BERT model and/or the Dense Layer.

The popularity of the attention mechanism stems from its potential to enhance text classification performance with interpretability. The remainder of this section introduces a few of the most well-known LLMs that are also employed for several text classification applications.

### Bidirectional Encoder Representations from Transformers (BERT)

BERT (Bidirectional Encoder Representations from Transformers) is a pre-trained language model that enables fine-tuning on specific tasks without requiring task-specific architectures. It is first trained on large amounts of unlabeled text (free text) using unsupervised objectives like Masked Language Modeling (MLM) and Next Sentence Prediction (NSP). After pre-training, BERT can be fine-tuned on individual down-stream tasks (e.g., sentiment analysis, question answering, named entity recognition) with minimal modification to the base architecture—usually just adding a small task-specific output layer. The contextualized word representation language model is presented in Devlin *et al.*, 2019 and uses parallel attention layers rather than sequential recurrence in the transformer. BERT is trained with two tasks in place of the fundamental language task to promote bidirectional prediction and sentence-level comprehension. BERT is trained on two unsupervised objectives: (1) an MLM task, in which 15% of the tokens are randomly masked (i.e., replaced with the "[MASK]" token), and the model is trained to predict the masked tokens; and (2) an NSP task, in which the model is given a pair of sentences and trained to determine when the second one follows the first. The purpose of this second assignment is to gather more practical or long-term data. English Wikipedia text passages and the dataset of Books Corpus are used in BERT training. The BERT-Base and BERT-Large pre-trained models are both available. BERT can be used for unannotated data as well as fine-tuned task-specific data directly from the trained model. Online resources include both the fine-tuning code and the publicly available pre-trained model.

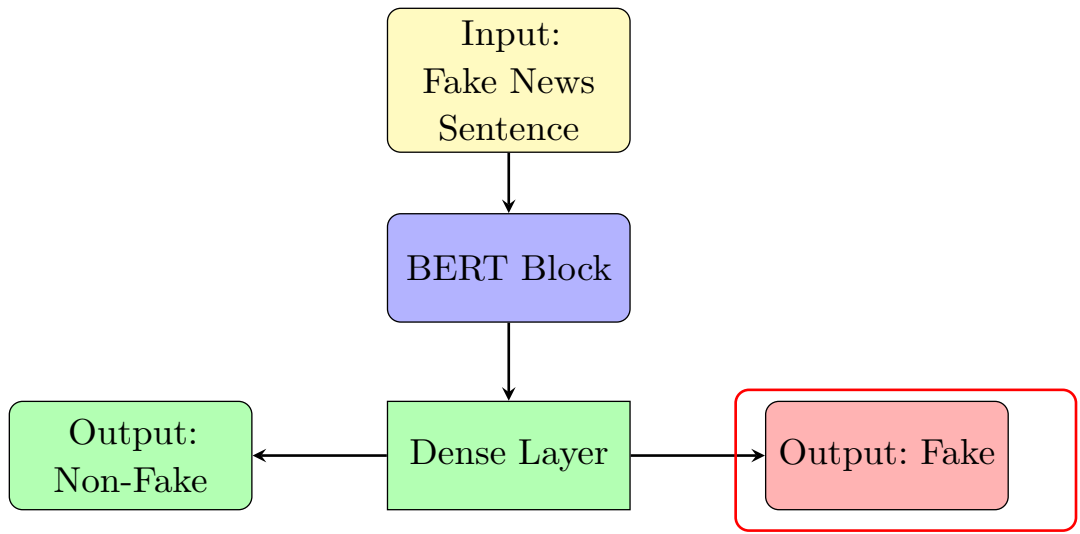

**Figure 5.4:** An LLM-based classifier using a BERT block and a dense layer to classify a sentence as fake or non-fake news. The figure highlights the "Fake" output as the result.

## RoBERTa

Authors in Liu *et al.*, 2019, by offering a replication study on the pre-training of BERT, improve the performance of the BERT model by changing the pre-training stage. These adjustments consist of the following: (1) training the model for more time using a larger batch size; (2) ignoring the objective of predicting the next sentence; (3) using longer sequences for training; (4) altering the pattern for masking dynamically used on the training instances.

## ALBERT

Despite its success, BERT has some drawbacks, such as its enormous amount of parameters, which leads to concerns with pre-training time degradation, memory management challenges and model degradation. These problems are extremely effectively addressed by ALBERT, which Lan proposed in Lan *et al.*, 2020 and updated based on the BERT architecture. ALBERT uses two-parameter reduction techniques to scale pre-trained models, removing the crucial obstacles. The large vocabulary embedding matrix is divided into two smaller matrices using factorized embedding parametrization, NSP loss is replaced with SOP loss, and cross-layer parameter sharing prevents the parameter from increasing with network depth. When compared to BERT, these techniques considerably reduce the amount of parameters utilized while having little to no impact on the model's performance, enhancing parameter efficiency. As BERT large has 18 times fewer parameters and can be trained roughly 1.7 times faster, an ALBERT configuration is the same as that. Despite having fewer parameters than BERT, ALBERT produces novel SOTA outcomes.

## DistilBERT

A lighter version of BERT based on a transformer (i.e., DistilBERT), requires a quicker model to train being a more compact general-purpose language representation model. DistilBERT shrinks the original BERT model by 40% while keeping 97% of its language understanding skills and increasing speed by 60%. If BERT can be seen as the instructor in the process of knowledge distillation, DistilBERT is the pupil. A little model that represents the student is trained to mimic the behaviour of the larger model (i.e., the teacher). Such a compact model is trained with a linear combination of three losses: the *distillation loss* (i.e., $L_{ce}$), the *masked language modelling loss* (i.e., $L_{mlm}$), and the *cosine embedding loss* (i.e., $L_{cos}$). Because of the distilled nature of the model, training and fine-tuning a specific dataset for a specific task is of prominent importance. Refer to Sanh *et al.*, 2019 for a thorough description of DistilBERT.

### XLNet

A generalized autoregressive pretraining strategy is the one suggested in Yang *et al.*, 2019. Optimizing the predicted likelihood across all combinations of the factorization order, it enables learning bidirectional contexts. BERT is surpassed by XLNet, often with a relevant margin, on a number of tasks, including question answering, sentiment analysis, document ranking and NLI. A popular implementation is the pre-trained XLNet using zero-shot (Chen *et al.*, 2021).

### Text-to-Text Transfer Transformer (T5)

By converting the data to text-to-text format and using an encoder-decoder framework, unified NLU and generation is possible. The T5 pre-training corpus has been developed, and it also comprehensively contrasts previously presented methodologies, in terms of pre-training aims, architectures, pre-training datasets, and transfer mechanisms. T5 (Raffel *et al.*, 2020) employs a pre-training for multitasking and a text infilling objective. T5 employs the decoder's token vocabulary as the prediction labels for fine-tuning.

**ELECTRA**

According to what stated in Clark *et al.*, 2020, ELECTRA suggests replacing certain tokens with possible replacements taken from a small generator network, instead of masking the input like in BERT. Then, a discriminative model is trained to predict whether each token in the corrupted input was replaced by a generator sample or not, as opposed to developing a model that predicts the original identities of the corrupted tokens. Along with GNN, ELECTRA can also be employed as an embedding layer, as in Lomonaco *et al.*, 2022.

### 5.3.3 Generative Pretrained Transformers (GPTs)

Generative Pre-trained Transformers (GPTs) represent a significant leap in the development of language models. Unlike previous approaches that employed masked token prediction, GPTs utilize an autoregressive approach, allowing them to generate text that follows a coherent sequence based on the preceding context. This characteristic enables GPTs to excel in various tasks such as text generation, completion, and dialogue systems. The architecture of GPTs is built on the Transformer model, which leverages self-attention mechanisms to capture long-range dependencies within the text. As a result, GPTs can produce more contextually relevant responses and maintain coherence over extended passages. Recent advancements have focused on scaling these models, leading to variants like GPT-3.

The most recent discipline related to the GPT models is *Prompt Engineering* (Siino and Tinnirello, 2024a; Siino and Tinnirello, 2024b; Siino and Tinnirello, 2024c). Prompt engineering involves crafting inputs to effectively guide the model's output, optimizing its performance across specific tasks. By systematically manipulating prompts, researchers have demonstrated significant improvements in task completion, allowing for tailored behaviour based on the user's intent. Furthermore, this field has led to the development of more sophisticated techniques that analyse the interaction between prompts and model outputs, uncovering underlying mechanisms of model behaviour. Techniques such as few-shot and zero-shot prompting have emerged, enabling models to generalize from limited examples and perform well on novel tasks without the need for extensive retraining. This advancement not only enhances the interoperability of GPT models across diverse applications but also emphasizes the importance of understanding context and nuance in prompt design.

In text classification, prompt engineering can be used to distinguish sentiment, detect spam, identify topics, or recognize biases within textual data. Moreover, advancements in chain-of-thought and few-shot prompting techniques enable LLMs to handle complex classification scenarios with improved interpretability and robustness, making them valuable tools in natural language processing applications (Fields *et al.*, 2024a; Yu *et al.*, 2023; Edwards and Camacho-Collados, 2024).

In the rest of this subsection, we discuss prompt engineering and some of the prompting techniques available to date, some modern GPTs, and some limitations and ethical considerations on the use of generative models.

**Prompt Engineering**

As already stated, effective prompt engineering plays a crucial role in maximizing the potential of generative models. By carefully crafting prompts, users can direct the model's outputs more effectively, achieving results that align more closely with their objectives. Additionally, encompassing variations in phrasing, context and examples can significantly influence the model's interpretation and the quality of its responses. This ability to drive the models' output based on the input prompt is sometimes referred in the literature as In-Context Learning (ICL) (Dong *et al.*, 2024). Practitioners need to consider the specific attributes of the generative model they are working with, as different models may respond variably to similar prompts. Just as an example, it is important to mention the recent findings in Liu *et al.*, 2024b. The authors experiment on different prompt lengths to identify the optimal conditions for eliciting coherent and contextually relevant outputs. Their results indicate a notable correlation between prompt length and response quality, emphasizing that both overly concise and excessively verbose prompts can hinder performance. In synthesis, the authors noticed that GPT models tend to pay more attention to the first and the last part of a long prompt, neglecting most of the content in the middle (i.e., "lost in the middle") Liu *et al.*, 2024b. Thanks to this finding, and as an example when designing a prompt, it would be beneficial to introduce the most relevant information at the beginning and the end of a long prompt. In the rest of this section, we discuss some of the most noticeable prompting techniques proposed in the literature.

**Zero-shot prompting**   Zero-shot Prompting (Liu *et al.*, 2024a; Kong *et al.*, 2024), wherein specific examples are not provided to guide the model, has shown potential effectiveness in generating coherent and contextually relevant outputs. This method capitalizes on the vast knowledge encapsulated within the model, allowing for flexibility and adaptability in various contexts. Zero-shot is defined as the model's ability to infer and generate responses based solely on its training data without the need for explicit examples. This approach raises important implications for applications across different domains, especially when rapid response generation is required. A simple example of zero-shot prompting, regarding the automatic labelling of a positive or negative movie review, would be: the model is prompted with a review such as "*The film was a thrilling experience with exceptional performances,*" and it must determine the sentiment without prior examples provided within the prompt. This ability reflects the underlying architecture's transfer learning capabilities, enabling it to understand nuances in language and sentiment.

**Few-shot prompting**   Few-shot prompting (Ye and Durrett, 2022; Siino and Tinnirello, 2024b) involves providing the model with a few examples of the desired output to guide its response. This technique has shown to enhance the performance of language models significantly, as it allows them to better understand the context and the specific requirements of the task at hand. Moreover, few-shot prompting not only aids in providing context but also helps to bridge the gap between zero-shot capabilities and fully supervised learning. By striking a balance between these approaches, we can leverage the strengths of both paradigms, facilitating a more flexible and adaptable learning process. This adaptability is crucial, especially in scenarios where labelled data is scarce or difficult to obtain. A simple example of a one-shot prompting to classify a movie review as positive or negative would be: "*The movie was awful! // NEGATIVE - The movie was fantastic! // *". In this case, a sample review is provided along with the label and the test sample misses the label which is expected to be provided by the model.

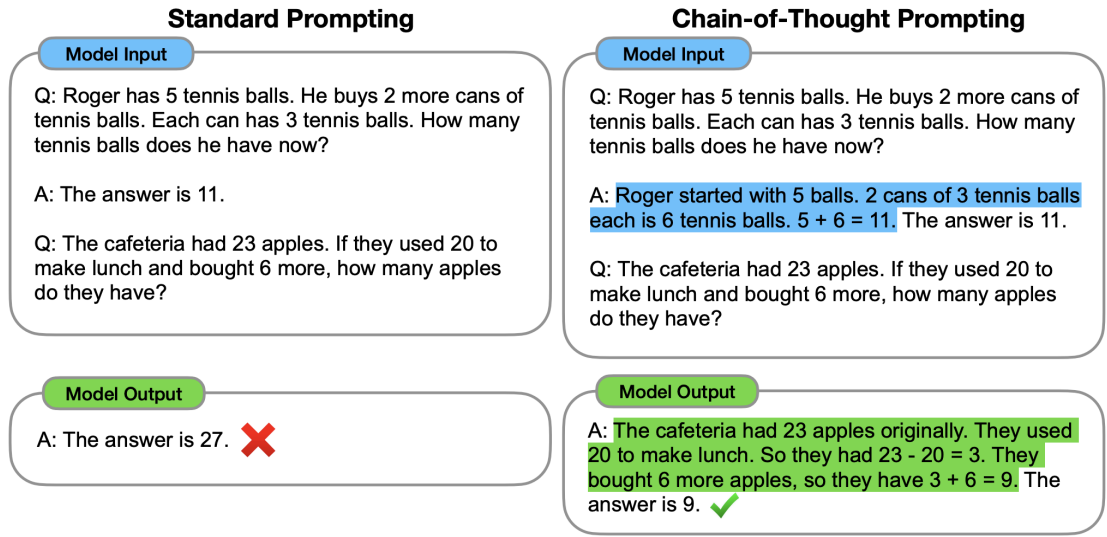

**Figure 5.5:** Example of CoT from the work presented in Wei *et al.*, 2022. Chain-of-thought prompting allows large language models to address intricate tasks involving arithmetic, common-sense reasoning, and symbolic logic. This approach emphasizes the reasoning processes underlying each step.

**Chain-of-Thoughts**   Chain-of-Thoughts (CoT) (Wei *et al.*, 2022) models to generate intermediate reasoning steps, which can facilitate understanding and improve the overall quality of responses. This approach has garnered attention for its capacity to enhance reasoning capabilities in GPT models, allowing them to tackle complex tasks more effectively. The technique was introduced in Wei *et al.*, 2022 and an image from the paper is shown in Figure 5.5.

**Chain-of-Code**   Chain-of-code prompting in large language models (LLMs) is a structured approach where multiple prompts are used sequentially to generate source code for complex tasks Siino *et al.*, 2024a; Lombardo *et al.*, 2024. Instead of requesting a complete solution in a single prompt, this technique breaks down the problem into smaller, manageable steps, guiding the LLM to produce modular and well-structured code. For instance, an initial prompt may define the overall goal, followed by prompts that generate specific functions, optimize performance, or add error handling. This iterative method improves code quality, enhances interpretability, and allows for easier debugging and refinement, making it particularly useful for automating software development tasks (Lombardo *et al.*, 2024; Siino *et al.*, 2024a).

**Retrieval-Augmented Generation (RAG)**   is a technique that enhances the capabilities of language models by integrating external information retrieval systems into the generation process (Lewis *et al.*, 2020). Instead of relying solely on the model's pre-trained knowledge, RAG retrieves relevant documents or pieces of information from an external knowledge base, such as a database or search engine, and incorporates them into the response. This makes it particularly effective for tasks that require up-to-date or domain-specific knowledge. The retrieved content helps the model ground its responses in factual data, improving both accuracy and relevance.

An example would be: "*What were the key events in climate policy during 2025?*". A standard language model might struggle to provide an accurate response if it wasn't trained on recent data. With RAG, the system first retrieves articles or documents summarizing major climate policy decisions from 2025. It then uses this information to generate a coherent and contextually relevant answer, such as: "*In 2025, key climate policy events included the introduction of stricter emission regulations in the EU, the U.S. rejoining the Paris Agreement, and a significant global summit in Tokyo focusing on renewable energy transitions.*"

This approach is particularly valuable for applications like customer support, real-time Q&A systems, and research, where accessing external knowledge ensures the information is both current and reliable.

**Self-Consistency**   (Ahmed and Devanbu, 2023) is a technique used in prompt engineering to improve the reliability of language models when performing tasks that require complex reasoning or multistep problem-solving. Instead of generating a single response, the model is prompted to produce multiple independent reasoning chains for the same query. The final answer is then determined by aggregating these outputs, often by selecting the most common answer or using a heuristic to decide among the generated options. This method helps mitigate errors caused by inconsistencies in individual reasoning paths, ensuring a more robust and accurate output.

Suppose the user asks, "What is the result of 25 multiplied by 13?" Instead of generating one chain of calculations, the model is asked to produce several reasoning paths:

1. "First, calculate $25 \times 10 = 250$, then add $25 \times 3 = 75$, resulting in $250 + 75 = 325$."

2. "Break it into $20 \times 13 = 260$ and $5 \times 13 = 65$. Add them to get $260 + 65 = 325$."

3. "Use direct multiplication: $25 \times 13 = 325$."

The model aggregates the results, and since all reasoning paths converge to 325, it confidently outputs the correct answer.

This technique is particularly useful in mathematical reasoning, logic puzzles, and tasks where intermediate steps can easily lead to errors. By exploring multiple paths and selecting the most consistent result, self-consistency improves the reliability of complex problem-solving processes.

### GPT Models

**GPT2** In 2019, the OpenAI team published GPT2 (Radford *et al.*, 2019), a scaled-up version of GPT. In terms of the location of layer normalization and residual relations, it adds a few minor enhancements over the previous version. There are actually four different GPT2 variants, the smallest of which is identical to GPT, the medium of which is comparable to BERT Large, and the xlarge of which was produced with 1.5B parameters, which is the actual GPT2 standard.

**Llama** (Touvron *et al.*, 2023) is an LLM developed by Meta, designed to handle a wide range of NLP tasks. LLaMa is a collection of foundation language models ranging from 7 billion to 65 billion parameters. These models are trained on trillions of tokens, demonstrating that it is possible to achieve state-of-the-art performance using publicly available datasets exclusively, without relying on proprietary and inaccessible data. Notably, LLaMA-13B outperforms GPT-3 (175B) on most benchmarks, and LLaMA-65B is competitive with top models such as Chinchilla-70B and PaLM-540B. The authors released all their models to the research community. Llama is known for its high performance in understanding and generating human-like text, excelling in tasks such as text completion, translation, and summarization. Llama models come in different sizes, ranging from smaller models with fewer parameters to larger models with billions of parameters.

**Gemini** (Islam and Ahmed, 2024) is a model developed by Google, focusing on multimodal learning. It integrates both textual and visual data to enhance its understanding and generation capabilities. Gemini is trained on a diverse dataset that includes text, images, and other multimedia content. This model is particularly effective in tasks that require a combination of textual and visual information, such as image captioning and visual question answering. Gemini models are designed to be versatile and can be adapted to various applications, including those that require real-time processing.

**Mistral** (Jiang *et al.*, 2023) is a language model developed by Mistral AI, a French startup headquartered in Paris. It is designed to handle a variety of NLP tasks with a focus on efficiency and performance. Mistral models are built on the transformer architecture and are trained on a diverse dataset. The model is known for its ability to generate coherent and contextually relevant text, making it suitable for applications such as chatbots, content generation, and language translation. Mistral models are available in different sizes, allowing for flexibility in deployment based on the specific needs of the application. The authors introduced Mistral 7B v0.1, a 7-billion-parameter language model engineered for superior performance and efficiency. Mistral 7B outperforms Llama 2 13B across all evaluated benchmarks and surpasses Llama 1 34B in reasoning, mathematics, and code generation. The model leverages grouped-query attention (GQA) for faster inference, coupled with sliding window attention (SWA) to effectively handle sequences of arbitrary length with reduced inference costs. Additionally, the authors provide a fine-tuned version, Mistral 7B – Instruct, which follows instructions and outperforms the Llama 2 13B – Chat model on both human and automated benchmarks. These models are released under the Apache 2.0 license.

## Limitations and Ethical Considerations

**Limitations** In the modern Natural Language Generation (NLG) domain, two interconnected challenges persist: neural models often produce linguistically fluent yet inaccurate output, while evaluation metrics primarily focus on fluency rather than accuracy (Siino and Tinnirello, 2024a). This situation leads to the phenomenon known as "hallucinations," where GPT generate output that sounds plausible but deviates from the intended meaning, making automatic detection difficult. Hallucinations are defined as instances where the generated text contains information that is not grounded in the input data or is factually incorrect. This issue is particularly problematic in many NLG applications where the accuracy of the output is crucial. For example, generating translations that diverge from the source text undermines the effectiveness of machine translation systems. Recent survey papers have highlighted that GPTs are especially prone to hallucinations, as evidenced in various studies. To mitigate these challenges, researchers are exploring multiple avenues, including improved training datasets, enhanced model architectures, and the integration of verification mechanisms that cross-check generated outputs against reliable external sources. Furthermore, the implementation of human-in-the-loop systems could help ensure that the outputs align more closely with factual information. By incorporating human oversight, these systems can effectively reduce the rate of hallucinations while allowing for dynamic feedback that can further refine the model's performance. Additionally, developing better metrics for evaluating the factual accuracy of gener-ated outputs is critical for advancing the reliability of LLMs. Current evaluation methods often fall short, lacking the nuance necessary to comprehensively assess the truthfulness of the information presented. This calls for innovative approaches that not only measure factual correctness but also contextual relevance.

**Ethical Considerations** The rapid development and deployment of GPTs raise several ethical considerations that warrant critical examination. One of the foremost concerns is the potential for misuse in generating misleading or harmful content. Additionally, there is the risk of perpetuating biases present in the training data, which can result in outputs that reinforce stereotypes or misinformation. The opacity of these models further complicates accountability, as it is often challenging to trace the origins of specific outputs or evaluate the decision-making processes that lead to their generation (Siino, 2024b). To address these concerns, it is essential to establish robust frameworks for transparency and accountability in the development and deployment of GPTs. This includes implementing guidelines for ethical usage, creating diverse and representative training datasets , and fostering collaboration among stakeholders, including researchers, policymakers, and ethicists. Education on the responsible use of such technologies should also be prioritized to enhance digital literacy among users. Furthermore, ongoing research into the interpretability of AI models will be crucial for understanding their internal mechanics, which, in turn, will aid in building trust between users and AI systems. Continued engagement with interdisciplinary perspectives will enrich this discourse, allowing for more comprehensive approaches to the challenges posed by GPTs. Ultimately, fostering a culture of responsibility and accountability will empower individuals and organizations to harness the potential of GPTs while mitigating risks. This should involve not only strict adherence to ethical standards but also an active pursuit of innovation that respects human values and societal norms. The path forward must be one that encourages collaboration among researchers, industry leaders, and policymakers. By establishing frameworks that prioritize transparency, inclusivity, and ethical considerations, the development of GPTs can be aligned with the broader goals of society. Emphasizing the importance of ongoing education and awareness is crucial in ensuring that all stake-
holders are equipped to navigate the complexities associated with these
technologies. Continuous training and interdisciplinary dialogue will
enhance understanding of GPT capabilities and limitations, enabling
more informed decision-making and fostering public trust. Furthermore,
as the landscape of emerging technologies continues to evolve, it is
imperative that we remain vigilant in our approach to regulation and
governance. Policymakers must stay ahead of the curve, adapting le-
gal frameworks to address new challenges while promoting innovation.
This dynamic relationship between technology and society requires a
collaborative effort among technologists, ethicists, and regulators to
create a holistic strategy that prioritizes ethical considerations alongside
technological advancements. By fostering a culture of transparency and
accountability, we can ensure that the deployment of these systems
aligns with societal values and promotes the common good. Future
research should focus on developing frameworks that facilitate this col-
laboration, examining case studies that illustrate successful partnerships
between these stakeholders. Additionally, ongoing engagement with the
public through education and dialogue will be crucial in demystify-
ing these technologies and empowering individuals to make informed
decisions. This participatory approach will not only enhance trust in
technological innovations but also allow for a more inclusive dialogue on
governance and policy-making. As we navigate this complex landscape,
it is essential to remain agile and responsive to emerging challenges and
opportunities that arise. Policymakers, technologists, and community
leaders must be vigilant in monitoring the impacts of these systems,
adapting strategies as needed to address unforeseen consequences. By
fostering an iterative process of feedback and refinement, we can create
a resilient framework that accommodates the rapid pace of innova-
tion while prioritizing ethical considerations and social well-being. This
proactive stance will encourage collaboration across sectors, stimulating
research and development that aligns with societal values. Furthermore,
engaging diverse stakeholders—from academics to marginalized com-
munities —will ensure that a plurality of perspectives is represented in
the decision-making process. This inclusivity is crucial for identifying
potential biases and inequities that may emerge as these technologies evolve. As we move forward, it is imperative to invest in educational initiatives that equip future generations with the critical skills needed to navigate and influence the landscape of emerging technologies. By fostering digital literacy and ethical reasoning, we empower individuals to critically assess the implications of their choices and the technologies they engage with. Additionally, interdisciplinary research should be encouraged to explore the intersections of technology, society, and ethics comprehensively. Collaborative projects that bring together experts from fields such as computer science and networking, sociology, law, and philosophy will yield richer insights and more robust solutions to the challenges still present (Siino *et al.*, 2025; Siino *et al.*, 2024a; Siino and Tinnirello, 2024c).

## 5.4   Hybrid and others approaches

### 5.4.1   Hybrid approaches

To capture local and global aspects of sentences and documents, many hybrid models that incorporate LSTM and CNN architectures have been developed.

A CNN-RNN model that can capture both global and local textual semantics and, consequently, represent high-order label correlations while having a manageable computational complexity is used by Chen *et al.*, 2017 to perform multi-label text classification.

A Convolutional LSTM (C-LSTM) network is suggested by Zhu *et al.*, 2018. In order to create the sentence representation, C-LSTM uses a CNN to extract a series of higher-level phrase (n-gram) representations. For document modelling, Zhang and Wallace, 2015 suggest using a Dependency Sensitive CNN (DSCNN). The sentence vectors learned by the LSTM in the hierarchical DSCNN model are then supplied to the convolution and max-pooling layers to produce the document representation.

Xiao and Cho, 2016 recommend using character-based convolution and recurrent layers for document encoding, since they see a document as a series of characters rather than words. When compared to word-level models, our model produced equivalent results with a lot less parameters.

Kowsarweet al. suggest a Hierarchical Deep Learning method for text classification in Kowsari *et al.*, 2017. At every level of the document hierarchy, HDLTex uses stacks of hybrid DL model architectures, such as MLP, RNN, and CNN, to give specialized knowledge.

A reliable Stochastic Answer Network (SAN) for multistep reasoning in machine reading comprehension is proposed by Liu *et al.*, 2018. Memory networks, Transformers, BiLSTM, attention networks, and CNN are just a few of the neural network types that are combined in SAN. The context representations for the questions and passages are obtained via the BiLSTM component. A passage representation that is question-aware is derived by its attention mechanism. A second LSTM is then employed to create a working memory for the section. A Gated Recurrent Unit (GRU) based answer module then generates predictions.

For language modelling, Kim *et al.*, 2016 use a highway network with CNN and LSTM over characters. A character embedding lookup is done in the first layer, followed by convolution and max-pooling operations to create a fixed-dimensional representation of the word that is then transferred to the highway network. The output of the highway network serves as the input for a multi-layer LSTM. To extract the distribution across the following word, an affine transformation and a softmax are then applied to the LSTM's hidden representation.

### 5.4.2 Other approaches

The *twin neural network* is another name for the siamese neural network (Chicco, 2021). It works in tandem with two different input vectors and uses equal weights to produce equivalent output vectors. A siamese adaptation of the LSTM network made up of pairs of variable-length sequences is presented by Mueller and Thyagarajan, 2016. The model, which outperforms ANN of higher complexity and painstakingly created features, is used to estimate the semantic similarity between texts. The model also encodes text using neural networks with word vectors as inputs that were separately learned from a sizeable dataset.

Deep learning techniques call for numerous additional hyperparameters, which raises the computational difficulty. In semi-supervised tasks, Virtual Adversarial Training (VAT) Miyato *et al.*, 2018 regularization based on local distributional smoothness can be employed. It simply needs a few hyperparameters and can be directly read as robust optimization. Miyato uses VAT to significantly enhance the model's robustness, generalizability, and word embedding performance.

By increasing the total number of rewards received, Reinforcement Learning (RL) learns the best course of action in a particular situation. Zhang *et al.*, 2018 provide an RL strategy for creating organized sentence representations by teaching the structures relevant to tasks. The model includes representation models for Hierarchical Structured LSTM (HS-LSTM) and Information Distilled LSTM (ID-LSTM). The HS-LSTM is a two-level LSTM for modelling sentence representation, and the ID-LSTM learns the sentence representation by selecting keywords that are pertinent to tasks.

Memory networks (Dai *et al.*, 2019) develop the capacity to integrate the long-term memory and inference components. LweLi and Lam, 2017, who uses two LSTMs with extended memories and neural memory operations to manage the extraction duties of aspects and opinions at once. Latent topic representations indicative of class labels are encoded using Topic Memory Networks (TMN) Zeng *et al.*, 2018, an end-to-end model.

Common-sense acquired outside the country. Authors of Ding *et al.*, 2019 believe that the event extracted from the original text lacked common knowledge, such as the goal and emotion of the event participants, because there was not enough information about the event itself to identify it for the EP task. The model enhances the effectiveness of stock forecasting, EP, and other factors.

The words and their relationships to one another are represented in the quantum language model by fundamental quantum events. In order to learn both the semantic and the sentiment information of subjective writing, Zhang *et al.*, 2019 propose a sentiment representation approach that is quantum-inspired. The model performs better when density matrices are added to the embedding layer.

Notable mention should also be made of integration-based (or ensemble learning) methods, which combine the output of various algorithms to improve performance and interpretation. These contain a number of subcategories, with bagging and boosting being the most well-liked ones. Breiman, 1996 (also known as bootstrap aggregation methods) averages the results of many classifiers without strong dependencies by training each of them separately on a part of the training data (sampling with replacement). Random forests are the most prevalent example of such a method, which increases accuracy and stability.

Interestingly, the model proposed and described in Siino *et al.*, 2022c section is T100. T100 include a logistic regressor model trained on the predictions provided by the first stage of classifiers. The model obtained interesting performance at the challenge hosted at PAN@CLEF2022. There the task was to investigate whether the author of a Twitter feed is likely to spread tweets containing irony and stereotypes. The model consists of a logistic regressor that gets as input the predictions provided by the first stage of classifiers (named *the voters*). The voters are a CNN, an SVM, a Naïve Bayes classifier and a Decision Tree. The training of the model is based on a 5-fold strategy. As a first step, the authors train each voter using the k-training fold. Then they let each voter predict on the corresponding k-validation fold. Then they merge the five sets of predictions on the validation folds. In such a way, a new prediction dataset is generated. In this newly generated predictions dataset, samples consist of voters' predictions and the original corresponding label of the input sample. This new predictions dataset is used to train the logistic regressor that provide the final classification label.

# 6

## Evaluation

To assess the performance of all the classification models discussed in the previous chapter, several metrics have been introduced and used in the literature. In particular, the usually employed ones include accuracy, precision, recall, and F1-score as primary evaluation metrics. Accuracy provides a general measure of how often the classifier is correct, while precision and recall offer insights into the model's ability to correctly identify positive cases and minimize false positives and false negatives, respectively. The F1-score serves as a harmonic mean of precision and recall, providing a single metric that balances both concerns. In this chapter, we will delve into the definition and discussion of these metrics and explore their respective strengths and weaknesses in various contexts. We will also investigate how these metrics can be affected by the distribution of classes within the dataset, particularly in scenarios involving imbalanced classes. Furthermore, we will discuss the implications of relying solely on one metric over another, particularly in cases where high precision might be prioritized at the expense of recall, or vice versa. This can lead to misinterpretations of model performance and potentially result in overlooking critical cases that may influence the overall effectiveness of a predictive system.

## 6.1 Traditional Machine Learning Metrics

The *F1 score* and *accuracy* are two metrics often employed to gauge the effectiveness of text classification models. Later, the assessment metrics are improved due to the complexity of the classification tasks or the existence of some specific activities. Single-label text classification separates samples in one of the categories that are most likely to be used in NLP tasks. It is possible to ignore the relationships between labels in single-label text classification because each text only belongs to one category. Multi-label text classification, as opposed to single-label text classification, breaks the corpus up into various category labels which depend on the task. These metrics were created for single-label text classification and are therefore inappropriate for multi-label jobs. Therefore, some metrics have been created for multi-label text classification. Before introducing the metrics reported in the literature, below we provide the definitions of the terms used in the following equations.

- **True Positive (TP)**. A single prediction provided by a classifier is referred to as a TP when the model *correctly* predicts a positive class.

- **True Negative (TN)**. A single prediction provided by a classifier is referred to as a TN when the model *correctly* predicts a negative class.

- **False Positive (FP)**. A single prediction provided by a classifier is referred to as an FP when the model *incorrectly* predicts a positive class.

- **False Negative (FN)**. A single prediction provided by a classifier is referred to as an FN when the model *incorrectly* predicts a negative class.

In the Table 6.1 is shown a *confusion matrix* (Stehman, 1997). A confusion matrix, also known as an *error matrix*, is a table structure which allows visualizing the performance of an algorithm, often a supervised learning one, in machine learning and, more specifically, the problem of statistical classification — in unsupervised learning it is usually called a matching matrix. Both variations of the matrix, where each column represents instances in the class predicted, and each row represents the actual class instances, are documented in the literature. The name was chosen since it is simple to determine whether the system is conflating two classes (i.e., commonly mislabelling one as another). It is a unique type of contingency table with two dimensions (actual and expected), identical sets of "classes" and two dimensions (each combination of dimension and class is a variable in the contingency table).

|            |          | **Actual** | |
| :--------: | :------: | :------: | :------: |
|            |          | **Positive** | **Negative** |
|            | **Positive** | *#TP* | *#FN* |
| **Predicted** | **Negative** | *#FP* | *#TN* |

**Table 6.1:** Confusion matrix illustrating the performance of a binary classification model. The matrix compares predicted labels to actual labels and contains four outcomes: True Positives (#TP), the number of samples that the model correctly predicts a positive class; False Negatives (#FN), the number of samples that the model incorrectly predicts a negative class for an actual positive; False Positives (#FP), the number of samples that the model incorrectly predicts a positive class for an actual negative; and True Negatives (#TN), the number of samples that the model correctly predicts a negative class.

Given the above definitions, the following are the common metrics used in literature for several text classification tasks.

**Accuracy**. Accuracy is the ratio of correct predictions on the total observations and is given by the Equation 6.1. Accuracy is one way to measure what percentage of predictions are right.

$$Accuracy = \frac{TP + TN}{TP + TN + FP + FN} \tag{6.1}$$

**Error rate**. Closely related to Accuracy is the *Error rate*. The definition is given by the Equation 6.2. The error rate expresses what percentage of predictions are wrong.

$$ErrorRate = 1 - Accuracy = \frac{FP + FN}{TP + TN + FP + FN} \qquad (6.2)$$

Depending on how genuine positives and negatives are defined in a multilabel scenario, the definition of this metric may differ. A prediction is deemed accurate (referred to as "subset accuracy") when the projected labels exactly match the actual labels. Alternately, before the accuracy calculation, predictions can be flattened and condensed to a single-label task.

**Precision**. Equation 6.3 defines *precision* or *sensitivity* as the ratio of true positive (TP) observations to all-around positive predicted values (TP+FP). Precision is the proportion of correctly predicted events among all positively predicted events.

$$Precision = \frac{TP}{TP + FP} \qquad (6.3)$$

**Recall**. Equation 6.4 gives *recall* or *specificity* as the ratio of true positive (TP) observations to all-around actual positive values (TP+FN). Recall is the ratio of right predictions made over all positive predictions that should have been made.

$$Recall = \frac{TP}{TP + FN} \qquad (6.4)$$

For scenarios involving multi-class classification, it is possible to compute the precision and recall for each class label.

**F1 score**. Equation 6.5 illustrates the F1 score, which is the harmonic mean of recall and precision. The maximum precision and recall value of an F1 score is 1, while the lowest value is 0.

$$F1 = 2 \times \frac{Recall \times Precision}{Recall + Precision} \qquad (6.5)$$

**Matthews Correlation Coefficient (MCC)**. The effectiveness of binary classification techniques is also measured by the Matthews Correlation Coefficient (MCC) (Matthews, 1975), which collects all the data in a confusion matrix. MCC can be used to address issues with unequal class sizes and is still regarded as a balanced approach. The MCC scales from -1 to 0. (i.e., the classification is always wrong and always true, respectively). Equation 6.6 provides the formula for MCC.

$$MCC = \frac{TP \times TN - FP \times FN}{\sqrt{(TP + FP)(TP + FN)(TN + FP)(TN + FN)}} \quad (6.6)$$

Finally, some specific metrics related to multilabel tasks are Micro and Macro-F1 Manning *et al.*, 2008, and Precision@k and Normalized Discounted Cumulated Gains Liu *et al.*, 2017.

## 6.2 Linguistic Metrics

When working with the evaluation of the text produced by LLMs, the most popular metrics are BLEU, ROUGE, and METEOR. These metrics focus on the evaluation of n-gram overlap between the generated text and reference text(s). However, these metrics have their limitations, particularly when it comes to capturing semantic similarity and contextual relevance. As such, recent research has begun to explore additional metrics that take into account semantic similarity, such as BERTScore and COMET, which leverage pretrained language models to evaluate text quality based on embeddings rather than n-gram matching. These advanced metrics aim to provide a more nuanced understanding of generated text by considering the contextual meaning of phrases and sentences. Furthermore, they enable the evaluation of generated content in a way that aligns more closely with human judgment, as they can discern subtle differences in meaning that traditional metrics might overlook. The integration of these new evaluation methodologies presents opportunities for refined measurements of text quality and offers a pathway toward improving the generation processes themselves. Furthermore, platforms that incorporate user feedback into the evaluation loop could foster a more dynamic system for continuous improvement. By integrating real-time user reactions and preferences, researchers can adapt and fine-tune generation algorithms to meet evolving standards of quality. This iterative process could also facilitate the development of personalized language models that cater to individual user needs, enhancing the relevance and effectiveness of generated content. Future research should focus on the ethical implications of such personalized systems, ensuring that they respect user privacy and mitigate biases present in training datasets. In the rest of this section, we define the above-mentioned metrics.

The Rouge-1 metric evaluates the overlap of unigrams between generated responses and reference texts, providing a straightforward measure of content similarity. It is essential in assessing the relevance of produced outputs to desired outcomes. The metric is defined as:

$$\text{Rouge-1} = \frac{\sum_{w \in \text{Words}} \text{count}_{\text{matched}}(w)}{\sum_{w \in \text{Words}} \text{count}(w)} \tag{6.7}$$

This equation accurately quantifies the ratio of matched unigrams, reflecting how well the generated text corresponds to expected results. Furthermore, the Rouge-L metric expands the evaluation by considering the longest common subsequence between the generated text and reference texts. This allows for a more nuanced understanding of context and sequence preservation in generated outputs. The Rouge-L metric is particularly useful in tasks where the order of information is crucial, as it emphasizes the importance of maintaining coherence and relevance throughout longer texts.

$$\text{Rouge-L} = \frac{LCS(X, Y)}{\text{length}(Y)} \tag{6.8}$$

where $LCS(X, Y)$ denotes the length of the longest common subsequence between the generated text $X$ and the reference text $Y$. This metric thus highlights how effectively the generated content preserves the structure and intent of the original material, which is particularly valuable in applications such as summarization.

Additionally, metrics such as BLEU can complement these evaluations by assessing n-gram overlaps and precision. BLEU can be defined as:

$$\text{BLEU} = BP \cdot \exp \left( \sum_{n=1}^{N} w_n \log p_n \right) \tag{6.9}$$

where:

- $BP$ is the brevity penalty.

- $N$ is the maximum n-gram order.

- $w_n$ are the weights for each n-gram order.

- $p_n$ is the modified n-gram precision.

The BLEU metric is a widely used evaluation measure for machine translation and other text generation tasks. The equation defines BLEU as the product of the brevity penalty $BP$ and the exponential of the weighted sum of log precisions for n-grams up to order $N$. The brevity penalty $BP$ is included to prevent very short translations from receiving high scores. The modified n-gram precision $p_n$ measures the overlap between the n-grams in the candidate translation and the reference translations, adjusted to avoid penalizing correct but repetitive n-grams. The weights $w_n$ allow for different emphasis on various n-gram orders, providing flexibility in the evaluation. Overall, the BLEU metric provides a balanced assessment of translation quality by considering both precision and recall of n-grams. It is essential to use a combination of these metrics to achieve a comprehensive evaluation framework that captures the multifaceted nature of text generation.

# 7

## Conclusion

### 7.1 Discussion

One of the most relevant challenges in the field of NLP is the text classification. The creation and publication of supervised machine learning methods is becoming increasingly important, especially for text classification as text and document datasets multiply. Determining these methods is necessary to have a better document categorization system for this information. However, the need to have a better understanding of the complete process involved in text classification tasks, models, and algorithms that are already in use could eventually operate more effectively. Currently, a pipeline of this kind can be broadly split in subsequent stages as follows: (I) Present challenges and datasets (II) Applying various strategies and techniques to the raw text during preprocessing, (III) Text representation techniques as Term Frequency-Inverse Document Frequency (TF-IDF), Term Frequency (TF), and Word2Vec, contextualized word representations, Global Vectors for Word Representation (GloVe), and FastText. (IV) Existing classification architectures such as random forest and deep learning models, Transformers, logistic regression, Bayesian classifier, k-nearest neighbor, support vector machine, decision tree classifier, and k-nearest neighbor. (V) Evaluation metrics, (VI) Conclusion and future perspectives on performance and comprehension of the text classification pipeline.

The following are the primary findings and contributions. We listed the prominent dataset used and available in the literature in Chapter 2 along with the current tasks, problems, and applications for text classification. The most popular preprocessing methods for preparing raw text are shown and explored in Chapter 3. In this chapter, we investigated the impact of common preprocessing techniques on a text classification model performance. We discuss that it is also possible to outperform the performance of large pre-trained model using simpler classifiers adopting the proper preprocessing strategy. In Chapter 4, methods for numerically representing text were described, together with a thorough introduction to the attention mechanism. In addition, as a further contribution, we proposed a methodology for an examination of a trained word embedding for a real-case problem and we used the results to improve the model's design. Traditional and contemporary classifiers used for text classification are covered in Chapter 5. The reference materials for a number of contemporary Transformers are listed. Contributions to this chapter regard several cross-experiments on real world datasets and a methodology for a post-hoc analysis of a CNN layers to investigate further the behaviour of a deep learning model and to improve its design. We go over all the evaluation metrics used in text classification in Chapter 6. In Chapter 7 future perspectives are provided along with the conclusions of this work.

We reported that the traditional approach enhances text classification performance primarily by enhancing the classifier design, preprocessing, and text representation scheme. The deep learning model, in contrast, improves performance by enhancing the presentation learning process, the model structure, and the inclusions of new information and data. We can finally say that attention to the very initial stages of the classification pipeline can lead to significant improvements in text classification tasks (i.e., data augmentation, text preprocessing and representation models). The importance of the ensuing stages varies according to the task being considered as well as the dataset involved.

## 7.2   Future perspectives

Two primary paths can be seen on the roadmap for NLP. The first is driven by bigger Transformer Models like GPT-3 and its future relatives. The second important breakthrough will be in dialogue models, where Google, Facebook, and other businesses are investing millions of dollars in R&D. At the moment, in almost every sector, GPT models are sensitively impacting on everyone's life. GPT-3 was created by Open AI, a research company that Elon Musk and other well-known figures like Sam Altman co-founded. A multitasking system called GPT-3 can speak with a human, interpret text, extract text, and, if you're bored, amuse you with its poems. GPT-3 has, nonetheless, developed expertise (and actual utility) in the area of producing computer code. Given the right guidelines, GPT-3 can create full programs in Python, Java, and a number of other languages, opening up interesting new possibilities. Bigger and bigger transformer models, like the GPT-4 or the Chinese variant known as Wu Dao 2.0, are on the horizon.

The second significant development in NLP is the study of dialogue models and conversational AI by Google and Facebook. For instance, Google unveiled a demonstration of the LAMDA conversational AI system. Unlike contemporary chatbots, which are programmed for specific conversations, LAMDA has the advantage of being able to communicate with people on a seemingly limitless range of themes. If LAMDA is effective, it will probably disrupt customer service, help desks, and "whole new types of useful applications," as one Google blog put it.

Text classification is a dynamic field constantly evolving with the advancements in NLP. The emergence of LLMs has ushered in a new era of possibilities, presenting both exciting opportunities and unique challenges. The following are some of the most promising future directions for text classification in the context of Transformers.

### 7.2.1   Enhanced interpretability and explainability

One of the major limitations of current LLMs is their inherent "black-box" nature. Understanding the rationale behind an LLM's classification decisions is crucial for building trust, identifying biases, and improving model robustness. Future research should focus on developing techniques to enhance the interpretability and explainability of LLM-based classifiers. This could involve methods such as:

- **Attention visualization.** Analysing the attention mechanisms within the LLM to identify the parts of the input text that most influenced the classification decision.

- **Feature importance analysis.** Determining the relative importance of different features (words, phrases, or even entire documents) in the classification process.

- **Counterfactual explanations.** Generating "what-if" scenarios to understand how changes to the input text would affect the classification outcome.

### 7.2.2   Addressing bias and fairness

LLMs are trained on massive datasets that may contain inherent biases. These biases can be reflected in the model's predictions, leading to unfair or discriminatory outcomes. Future research should focus on developing techniques to mitigate bias in LLM-based classifiers, such as:

- **Bias detection and mitigation techniques.** Developing methods to identify and quantify biases in LLM training data and in the model's predictions.

- **Fairness-aware training objectives.** Incorporating fairness constraints into the training process to ensure that the model treats different groups of users equitably.

- **De-biasing techniques.** Developing methods to remove or mitigate biases that have already been learned by the model.

- **Regulatory frameworks.** Work towards the development of regulatory frameworks that govern the use of LLMs in text classification. This will help ensure that these models are used responsibly and ethically, while also promoting innovation and progress in the field.

### 7.2.3 Continual learning and adaptation

The real world is constantly changing, and it is essential for text classification systems to adapt to new information and evolving trends. Future research should focus on developing techniques for continual learning in LLM-based classifiers, such as:

- **Incremental learning.** Enabling LLMs to learn new information without forgetting previously learned knowledge.

- **Few-Shot and Zero-Shot learning.** Enabling LLMs to perform well on new classification tasks with limited or no labelled data.

- **Online learning.** Enabling LLMs to adapt to changing data streams in real-time.

### 7.2.4 Cross-lingual text classification

While many LLMs have demonstrated impressive cross-lingual capabilities, further research is needed to improve the performance of LLM-based classifiers on low-resource languages and in multilingual settings. This could involve:

- Developing more effective cross-lingual transfer learning techniques.

- Leveraging multilingual training data to improve model generalization across languages.

- Addressing the challenges of low-resource languages with limited labelled data.

### 7.2.5 Human-in-the-loop systems

Integrating human feedback into the LLM-based classification process can significantly improve model performance and address limitations such as bias and lack of interpretability. Future research should focus on developing effective human-in-the-loop systems, such as:

- **Active learning.** Actively querying human annotators for labels on the most informative data points.

- **Interactive classification systems.** Allowing users to provide feedback and refine the model's predictions in real-time.

- **Explainable AI for human-computer interaction.** Designing interfaces that effectively communicate the model's reasoning to human users.

### 7.2.6   Conclusion

LLMs have already demonstrated the potential to revolutionize the field of text classification, but significant challenges remain. By addressing the issues of interpretability, bias, continual learning, cross-lingual classification, and human-in-the-loop systems, researchers can unlock the full potential of LLMs for a wide range of real-world applications. Continued research and development in these areas will be crucial for advancing the state-of-the-art in text classification and ensuring that these powerful technologies are used responsibly and effectively.

In conclusion, the recent strides in NLP not only render it an appealing investment for professionals and IT enthusiasts but also mark a pivotal moment in its widespread adoption across key sectors such as finance, insurance, and healthcare. The swift expansion of the NLP market as a composite of various technologies underscores the need for practitioners to astutely identify the underlying systems with the utmost commercial potential and strategically time their implementation. Looking forward, the bright future of NLP is unequivocal, characterized by continual enhancements in user experience and the emergence of novel opportunities in unexplored markets. As NLP continues to evolve, its trajectory appears to be one of sustained growth and transformative impact in almost every area of knowledge.

**CRediT Authorship Contribution Statement**

**Marco Siino:** Conceptualization, Formal analysis, Investigation, Methodology, Resources, Software, Validation, Visualization, Writing - Original draft, Writing - review & editing. **Ilenia Tinnirello:** Supervision. **Marco La Cascia:** Supervision.

# Acknowledgements

(From Marco Siino)

This work originates from my PhD thesis conducted under the supervision of Professor Ilenia Tinnirello and Professor Marco La Cascia. First and foremost, I extend my deepest gratitude to Professor Ilenia Tinnirello, whose guidance, dedication, and unwavering support have been fundamental to my research journey. Her mentorship has been an invaluable source of inspiration, not only in the academic domain but also in shaping my ethical and professional principles. She has consistently demonstrated an exceptional commitment to the integrity of public institutions, imparting lessons that extend far beyond the realm of engineering. Her encouragement to continue my research in Palermo, coupled with her tireless efforts to integrate me into the Telecommunications Engineering team, has profoundly influenced my academic path. Without her guidance, this journey would not have been the same.

I also extend my sincere appreciation to Professor Marco La Cascia, my co-supervisor, who has always believed in my abilities and provided invaluable insights throughout my doctoral studies. His support, advice, and thoughtful discussions have been instrumental in refining my approach to research. Beyond academia, he has imparted life lessons that I will cherish, particularly in balancing professional dedication with personal passions.

My gratitude goes as well to the anonymous reviewers of Foundations and Trends® in Information Retrieval, whose constructive feedback and critical insights have significantly improved the quality of this work. Their rigorous evaluations have been crucial in refining the research presented in this paper. Additionally, I would like to sincerely thank Mark de Jongh, whose kindness and professionalism have made the entire editorial process a smooth and enriching experience. His cordiality and attentive approach have been truly appreciated at every stage of this publication process.

My sincere thanks go to Professor Paolo Rosso, whose pioneering work in NLP and shared tasks has been a source of inspiration for much of my research. The time spent collaborating with him and sharing meaningful discussions in Palermo remains a highlight of my academic journey. His sharp wit, infectious enthusiasm, and profound knowledge have left a lasting impression on me and my colleagues, providing insights that extend far beyond the technical aspects of machine learning.

I am deeply grateful to my dad, Emanuele, my entire family, and my closest friends, to whom I owe many apologies for my countless absences due to academic commitments. Despite the time apart, their unwavering love, support, and admiration have been a constant source of strength throughout this journey. I consider myself incredibly fortunate to have them by my side.

A heartfelt thank you to Micaela Di Carlo, who has stood by me throughout the PhD years, always envisioning a future for me in academia. Despite my demanding schedule, we have shared countless unforgettable moments, indulging in our mutual love for music, films, and literature.

Finally, as a tribute to all those who have accompanied me on this journey, I find it fitting to borrow the words once inscribed outside the helmet of the most successful Formula 1 driver of all time—who was also the only hero of my adolescence—words that best express my gratitude to everyone who has been part of this incredible experience.

"*Life is about passion, thanks for sharing mine.*"
(M. Schumacher, Interlagos, 2012)

# References

A. Mullen, L., K. Benoit, O. Keyes, D. Selivanov, and J. Arnold. (2018). "Fast, consistent tokenization of natural language text". *Journal of Open Source Software.* 3(23): 655.

Agarwal, A., B. Xie, I. Vovsha, O. Rambow, and R. Passonneau. (2011). "Sentiment analysis of twitter data". In: *Proceedings of the workshop on language in social media (LSM 2011).* 30–38.

Ahmed, T. and P. T. Devanbu. (2023). "Better Patching Using LLM Prompting, via Self-Consistency". In: *38th IEEE/ACM International Conference on Automated Software Engineering, ASE 2023, Luxembourg, September 11-15, 2023.* IEEE. 1742–1746. DOI: 10.1109/ASE56229.2023.00065.

Akın, A. A. and M. D. Akın. (2007). "Zemberek, an open source NLP framework for Turkic languages". *Structure.* 10: 1–5.

Alam, S. and N. Yao. (2019). "The impact of preprocessing steps on the accuracy of machine learning algorithms in sentiment analysis". *Computational and Mathematical Organization Theory.* 25(3): 319–335.

Albalawi, Y., J. Buckley, and N. S. Nikolov. (2021). "Investigating the impact of pre-processing techniques and pre-trained word embeddings in detecting Arabic health information on social media". *Journal of big Data.* 8(1): 1–29.

Aliakbarzadeh, A., L. Flek, and A. Karimi. (2025). "Exploring Robustness of Multilingual LLMs on Real-World Noisy Data". *arXiv preprint arXiv:2501.08322.*

Aljebreen, A., W. Meng, and E. Dragut. (2021). "Segmentation of tweets with urls and its applications to sentiment analysis". In: *Proceedings of the AAAI Conference on Artificial Intelligence.* Vol. 35. 12480–12488.

Alzahrani, E. and L. Jololian. (2021). "How Different Text-preprocessing Techniques Using The BERT Model Affect The Gender Profiling of Authors". *arXiv preprint arXiv:2109.13890.*

Anandarajan, M., C. Hill, and T. Nolan. (2019). "Text preprocessing". In: *Practical Text Analytics.* Springer. 45–59.

Angiani, G., L. Ferrari, T. Fontanini, P. Fornacciari, E. Iotti, F. Magliani, and S. Manicardi. (2016). "A comparison between preprocessing techniques for sentiment analysis in Twitter". In: *2nd International Workshop on Knowledge Discovery on the WEB (KDWEB).* Vol. 1748.

Araslanov, E., E. Komotskiy, and E. Agbozo. (2020). "Assessing the Impact of Text Preprocessing in Sentiment Analysis of Short Social Network Messages in the Russian Language". In: *2020 International Conference on Data Analytics for Business and Industry: Way Towards a Sustainable Economy (ICDABI).* IEEE. 1–4.

Arief, M. and M. B. M. Deris. (2021). "Text Preprocessing Impact for Sentiment Classification in Product Review". In: *2021 Sixth International Conference on Informatics and Computing (ICIC).* IEEE. 1–7.

Atmadja, A. R. and A. Purwarianti. (2015). "Comparison on the rule based method and statistical based method on emotion classification for Indonesian Twitter text". In: *2015 International Conference on Information Technology Systems and Innovation (ICITSI).* 1–6. DOI: 10.1109/ICITSI.2015.7437692.

Babanejad, N., A. Agrawal, A. An, and M. Papagelis. (2020). "A Comprehensive Analysis of Preprocessing for Word Representation Learning in Affective Tasks". In: *Proceedings of the 58th Annual Meeting of the Association for Computational Linguistics.* Online: Association for Computational Linguistics. 5799–5810. DOI: 10.18653/v1/2020.acl-main.514.

Bahdanau, D., K. Cho, and Y. Bengio. (2015). "Neural Machine Translation by Jointly Learning to Align and Translate". In: *3rd International Conference on Learning Representations, ICLR 2015, San Diego, CA, USA, May 7-9, 2015, Conference Track Proceedings.* Ed. by Y. Bengio and Y. LeCun.

Bakliwal, A., P. Arora, S. Madhappan, N. Kapre, M. Singh, and V. Varma. (2012). "Mining sentiments from tweets". In: *Proceedings of the 3rd Workshop in Computational Approaches to Subjectivity and Sentiment Analysis.* 11–18.

Balahur, A. (2013). "Sentiment analysis in social media texts". In: *Proceedings of the 4th workshop on computational approaches to subjectivity, sentiment and social media analysis.* 120–128.

Bansal, H., G. Shrivastava, G. N. Nguyen, and L.-M. Stanciu. (2018). *Social network analytics for contemporary business organizations.* IGI Global. DOI: 10.4018/978-1-5225-5097-6.

Bao, Y., C. Quan, L. Wang, and F. Ren. (2014). "The role of preprocessing in twitter sentiment analysis". In: *International conference on intelligent computing.* Springer. 615–624.

Barbosa, L. and J. Feng. (2010). "Robust Sentiment Detection on Twitter from Biased and Noisy Data". In: *COLING 2010, 23rd International Conference on Computational Linguistics, Posters Volume, 23-27 August 2010, Beijing, China.* Ed. by C. Huang and D. Jurafsky. Chinese Information Processing Society of China. 36–44.

Bengio, Y., R. Ducharme, and P. Vincent. (2000). "A Neural Probabilistic Language Model". In: *Advances in Neural Information Processing Systems.* Ed. by T. Leen, T. Dietterich, and V. Tresp. Vol. 13. MIT Press.

Benzarti, S. and R. Faiz. (2015). "EgoTR: Personalized tweets recommendation approach". In: *Intelligent Systems in Cybernetics and Automation Theory: Proceedings of the 4th Computer Science Online Conference 2015 (CSOC2015), Vol 2: Intelligent Systems in Cybernetics and Automation Theory.* Springer. 227–238.

Bevendorff, J., B. Chulvi, E. Fersini, A. Heini, M. Kestemont, K. Kredens, M. Mayerl, R. Ortega-Bueno, P. Pkezik, M. Potthast, *et al.* (2022). "Overview of PAN 2022: Authorship Verification, Profiling Irony and Stereotype Spreaders, and Style Change Detection". In: *International Conference of the Cross-Language Evaluation Forum for European Languages.* Springer. 382–394.

Boiy, E., P. Hens, K. Deschacht, and M. Moens. (2007). "Automatic Sentiment Analysis in On-line Text". In: *Openness in Digital Publishing: Awareness, Discovery and Access - Proceedings of the 11th International Conference on Electronic Publishing held in Vienna - ELPUB 2007, Vienna, Austria, June 13-15, 2007. Proceedings.* 349–360.

Bojanowski, P., E. Grave, A. Joulin, and T. Mikolov. (2017). "Enriching Word Vectors with Subword Information". *Transactions of the Association for Computational Linguistics.* 5: 135–146. ISSN: 2307-387X.

Borra, E. and B. Rieder. (2014). "Programmed method: Developing a toolset for capturing and analyzing tweets". *Aslib journal of information management.* 66(3): 262–278.

Bowman, S. R., G. Angeli, C. Potts, and C. D. Manning. (2015). "A large annotated corpus for learning natural language inference". In: *Proceedings of the 2015 Conference on Empirical Methods in Natural Language Processing.* 632–642.

Breiman, L. (1996). "Bagging predictors". *Machine learning.* 24: 123–140.

Bueno, R. O., B. Chulvi, F. Rangel, P. Rosso, and E. Fersini. (2022). "Profiling Irony and Stereotype Spreaders on Twitter (IROSTEREO). Overview for PAN at CLEF 2022". In: *Proceedings of the Working Notes of CLEF 2022 - Conference and Labs of the Evaluation Forum, Bologna, Italy, September 5th - to - 8th, 2022.* Ed. by G. Faggioli, N. Ferro, A. Hanbury, and M. Potthast. Vol. 3180. *CEUR Workshop Proceedings.* CEUR-WS.org. 2314–2343.

Byrd, R. H., P. Lu, J. Nocedal, and C. Zhu. (1995). "A limited memory algorithm for bound constrained optimization". *SIAM Journal on scientific computing.* 16(5): 1190–1208.

Camacho-Collados, J. and M. T. Pilehvar. (2018). "On the Role of Text Preprocessing in Neural Network Architectures: An Evaluation Study on Text Categorization and Sentiment Analysis". In: *Proceedings of the 2018 EMNLP Workshop BlackboxNLP: Analyzing and Interpreting Neural Networks for NLP*. 40–46.

Can, F., S. Kocberber, E. Balcik, C. Kaynak, H. C. Ocalan, and O. M. Vursavas. (2008). "Information retrieval on Turkish texts". *Journal of the American Society for Information Science and Technology*. 59(3): 407–421.

Chang, C.-C. and C.-J. Lin. (2011). "LIBSVM: a library for support vector machines". *ACM transactions on intelligent systems and technology (TIST)*. 2(3): 1–27.

Chen, G., S. Ma, Y. Chen, L. Dong, D. Zhang, J. Pan, W. Wang, and F. Wei. (2021). "Zero-Shot Cross-Lingual Transfer of Neural Machine Translation with Multilingual Pretrained Encoders". In: *Proceedings of the 2021 Conference on Empirical Methods in Natural Language Processing*. 15–26.

Chen, G., D. Ye, Z. Xing, J. Chen, and E. Cambria. (2017). "Ensemble application of convolutional and recurrent neural networks for multi-label text categorization". In: *2017 International joint conference on neural networks (IJCNN)*. IEEE. 2377–2383.

Cheng, J., L. Dong, and M. Lapata. (2016). "Long Short-Term Memory-Networks for Machine Reading". In: *Proceedings of the 2016 Conference on Empirical Methods in Natural Language Processing, EMNLP 2016, Austin, Texas, USA, November 1-4, 2016*. Ed. by J. Su, X. Carreras, and K. Duh. The Association for Computational Linguistics. 551–561. DOI: 10.18653/v1/d16-1053.

Chicco, D. (2021). "Siamese Neural Networks: An Overview". In: *Artificial Neural Networks - Third Edition*. Ed. by H. M. Cartwright. Vol. 2190. *Methods in Molecular Biology*. Springer. 73–94. DOI: 10.1007/978-1-0716-0826-5\_3.

Clark, K., M. Luong, Q. V. Le, and C. D. Manning. (2020). "ELEC-TRA: Pre-training Text Encoders as Discriminators Rather Than Generators". In: *8th International Conference on Learning Representations, ICLR 2020, Addis Ababa, Ethiopia, April 26-30, 2020*. OpenReview.net.

Colas, F. and P. Brazdil. (2006). "Comparison of SVM and some older classification algorithms in text classification tasks". In: *IFIP International Conference on Artificial Intelligence in Theory and Practice.* Springer. 169–178.

Cortes, C. and V. Vapnik. (1995). "Support-vector networks". *Machine learning.* 20(3): 273–297.

Courseault Trumbach, C. and D. Payne. (2007). "Identifying synonymous concepts in preparation for technology mining". *Journal of Information Science.* 33(6): 660–677.

Cover, T. and P. Hart. (1967). "Nearest neighbor pattern classification". *IEEE transactions on information theory.* 13(1): 21–27.

Cover, T. M. and J. A. Thomas. (2001). *Elements of Information Theory.* Wiley. ISBN: 9780471062592. DOI: 10.1002/0471200611.

Croce, D., D. Garlisi, and M. Siino. (2022). "An SVM Ensembler Approach to Detect Irony and Stereotype Spreaders on Twitter". In: *Proceedings of the Working Notes of CLEF 2022 - Conference and Labs of the Evaluation Forum (CLEF)* (Bologna, Italy, Sept. 5–8, 2022). Ed. by G. Faggioli, N. Ferro, A. Hanbury, and M. Potthast. *CEUR Workshop Proceedings.* No. 3180. Aachen. 2426–2432.

Cunha, W., S. Canuto, F. Viegas, T. Salles, C. Gomes, V. Mangaravite, E. Resende, T. Rosa, M. A. Gonçalves, and L. Rocha. (2020). "Extended pre-processing pipeline for text classification: On the role of meta-feature representations, sparsification and selective sampling". *Information Processing & Management.* 57(4): 102263. ISSN: 0306-4573. DOI: https://doi.org/10.1016/j.ipm.2020.102263.

Cunha, W., V. Mangaravite, C. Gomes, S. Canuto, E. Resende, C. Nascimento, F. Viegas, C. França, W. S. Martins, J. M. Almeida, T. Rosa, L. Rocha, and M. A. Gonçalves. (2021). "On the cost-effectiveness of neural and non-neural approaches and representations for text classification: A comprehensive comparative study". *Information Processing & Management.* 58(3): 102481. ISSN: 0306-4573. DOI: https://doi.org/10.1016/j.ipm.2020.102481.

Dai, Z., Z. Yang, Y. Yang, J. G. Carbonell, Q. Le, and R. Salakhutdinov. (2019). "Transformer-XL: Attentive Language Models beyond a Fixed-Length Context". In: *Proceedings of the 57th Annual Meeting of the Association for Computational Linguistics.* 2978–2988.

Dale, R. (2021). "GPT-3: What's it good for?" *Natural Language Engineering.* 27(1): 113–118.

Das, A. S., M. Datar, A. Garg, and S. Rajaram. (2007). "Google news personalization: scalable online collaborative filtering". In: *Proceedings of the 16th international conference on World Wide Web.* 271–280.

Deng, L. and J. Wiebe. (2015). "Mpqa 3.0: An entity/event-level sentiment corpus". In: *Proceedings of the 2015 conference of the North American chapter of the association for computational linguistics: human language technologies.* 1323–1328.

Denny, M. J. and A. Spirling. (2018). "Text preprocessing for unsupervised learning: Why it matters, when it misleads, and what to do about it". *Political Analysis.* 26(2): 168–189.

Devlin, J., M. Chang, K. Lee, and K. Toutanova. (2019). "BERT: Pre-training of Deep Bidirectional Transformers for Language Understanding". In: *Proceedings of the 2019 Conference of the North American Chapter of the Association for Computational Linguistics: Human Language Technologies, NAACL-HLT 2019, Minneapolis, MN, USA, June 2-7, 2019, Volume 1 (Long and Short Papers).* Ed. by J. Burstein, C. Doran, and T. Solorio. Association for Computational Linguistics. 4171–4186. DOI: 10.18653/v1/n19-1423.

Dieng, A. B., C. Wang, J. Gao, and J. W. Paisley. (2017). "TopicRNN: A Recurrent Neural Network with Long-Range Semantic Dependency". In: *5th International Conference on Learning Representations, ICLR 2017, Toulon, France, April 24-26, 2017, Conference Track Proceedings.* OpenReview.net.

Ding, X., K. Liao, T. Liu, Z. Li, and J. Duan. (2019). "Event representation learning enhanced with external commonsense knowledge". *arXiv preprint arXiv:1909.05190.*

Djuric, N., J. Zhou, R. Morris, M. Grbovic, V. Radosavljevic, and N. Bhamidipati. (2015). "Hate speech detection with comment embeddings". In: *Proceedings of the 24th international conference on world wide web.* 29–30.

Dolamic, L. and J. Savoy. (2010). "When stopword lists make the difference". *J. Assoc. Inf. Sci. Technol.* 61(1): 200–203. DOI: 10.1002/asi.21186.

Dolan, W. B. and C. Brockett. (2005). "Automatically constructing a corpus for paraphrase recognition". In: *Proceedings of the Third International Workshop on Paraphrasing (IWP2005)*.

Dong, Q., L. Li, D. Dai, C. Zheng, J. Ma, R. Li, H. Xia, J. Xu, Z. Wu, B. Chang, X. Sun, and Z. Sui. (2024). "A Survey on In-context Learning". In: *Proceedings of the 2024 Conference on Empirical Methods in Natural Language Processing, EMNLP 2024, Miami, FL, USA, November 12-16, 2024.* Ed. by Y. Al-Onaizan, M. Bansal, and Y. Chen. Association for Computational Linguistics. 1107–1128.

Duong, H.-T. and T.-A. Nguyen-Thi. (2021). "A review: preprocessing techniques and data augmentation for sentiment analysis". *Computational Social Networks.* 8(1): 1–16.

Edwards, A. and J. Camacho-Collados. (2024). "Language Models for Text Classification: Is In-Context Learning Enough?" In: *Proceedings of the 2024 Joint International Conference on Computational Linguistics, Language Resources and Evaluation (LREC-COLING 2024)*. Ed. by N. Calzolari, M.-Y. Kan, V. Hoste, A. Lenci, S. Sakti, and N. Xue. Torino, Italia: ELRA and ICCL. 10058–10072.

Emanuel, R. H. K., P. D. Docherty, H. Lunt, and K. Möller. (2024). "The effect of activation functions on accuracy, convergence speed, and misclassification confidence in CNN text classification: a comprehensive exploration". *J. Supercomput.* 80(1): 292–312. DOI: 10.1007/S11227-023-05441-7.

Fields, J., K. Chovanec, and P. Madiraju. (2024a). "A Survey of Text Classification With Transformers: How Wide? How Large? How Long? How Accurate? How Expensive? How Safe?" *IEEE Access.* 12: 6518–6531. DOI: 10.1109/ACCESS.2024.3349952.

Fields, J., K. Chovanec, and P. Madiraju. (2024b). "A survey of text classification with transformers: How wide? how large? how long? how accurate? how expensive? how safe?" *IEEE Access.* 12: 6518–6531.

Flood, B. J. (1999). "Historical Note: The Start of a Stop List at Biological Abstracts". *J. Am. Soc. Inf. Sci.* 50(12): 1066. DOI: 10.1002/(SICI)1097-4571(1999)50:12\<1066::AID-ASI5\>3.0.CO;2-A.

Gao, J., B. Peng, C. Li, J. Li, S. Shayandeh, L. Liden, and H.-Y. Shum. (2020). "Robust Conversational AI with Grounded Text Generation". *arXiv e-prints*: arXiv–2009.

Garg, N. and K. Sharma. (2022). "Text pre-processing of multilingual for sentiment analysis based on social network data." *International Journal of Electrical & Computer Engineering (2088-8708)*. 12(1).

Garrido-Merchan, E. C., R. Gozalo-Brizuela, and S. Gonzalez-Carvajal. (2023). "Comparing BERT against traditional machine learning models in text classification". *Journal of Computational and Cognitive Engineering*. 2(4): 352–356.

Gemci, F. and K. A. Peker. (2013). "Extracting Turkish tweet topics using LDA". In: *2013 8th International Conference on Electrical and Electronics Engineering (ELECO)*. IEEE. 531–534.

Genkin, A., D. D. Lewis, and D. Madigan. (2007). "Large-Scale Bayesian Logistic Regression for Text Categorization". *Technometrics*. 49(3): 291–304. DOI: 10.1198/004017007000000245.

Gerlach, M., H. Shi, and L. A. N. Amaral. (2019). "A universal information theoretic approach to the identification of stopwords". *Nature Machine Intelligence*. 1(12): 606–612.

González, J. Á., L.-F. Hurtado, and F. Pla. (2020). "Transformer based contextualization of pre-trained word embeddings for irony detection in Twitter". *Information Processing & Management*. 57(4): 102262. ISSN: 0306-4573. DOI: https://doi.org/10.1016/j.ipm.2020.102262.

Granik, M. and V. Mesyura. (2017). "Fake news detection using naive Bayes classifier". In: *2017 IEEE first Ukraine conference on electrical and computer engineering (UKRCON)*. IEEE. 900–903.

Gupta, V. and G. S. Lehal. (2011). "Punjabi language stemmer for nouns and proper names". In: *Proceedings of the 2nd Workshop on South Southeast Asian Natural Language Processing (WSSANLP)*. 35–39.

Guzman, E. and W. Maalej. (2014). "How Do Users Like This Feature? A Fine Grained Sentiment Analysis of App Reviews". In: *2014 IEEE 22nd International Requirements Engineering Conference (RE)*. 153–162. DOI: 10.1109/RE.2014.6912257.

HaCohen-Kerner, Y., D. Miller, and Y. Yigal. (2020). "The influence of preprocessing on text classification using a bag-of-words representation". *PloS one*. 15(5): e0232525.

Haddi, E., X. Liu, and Y. Shi. (2013). "The Role of Text Pre-processing in Sentiment Analysis". In: *Proceedings of the First International Conference on Information Technology and Quantitative Management, ITQM 2013, Dushu Lake Hotel, Sushou, China, 16-18 May, 2013*. Ed. by Y. Shi, Y. Xi, P. Wolcott, Y. Tian, J. Li, D. Berg, Z. Chen, E. Herrera-Viedma, G. Kou, H. Lee, Y. Peng, and L. Yu. Vol. 17. *Procedia Computer Science*. Elsevier. 26–32. DOI: 10.1016/j.procs.2013.05.005.

Hair Zaki, U. H., R. Ibrahim, S. Abd Halim, and I. I. Kamsani. (2022). "Text Detergent: The Systematic Combination of Text Preprocessing Techniques for Social Media Sentiment Analysis". In: *International Conference of Reliable Information and Communication Technology*. Springer. 50–61.

Hassler, M. and G. Fliedl. (2006). "Text preparation through extended tokenization". *WIT Transactions on Information and Communication Technologies*. 37.

Hernández Farías, D. I., R. M. Ortega-Mendoza, and M. Montes-y-Gómez. (2019). "Exploring the use of psycholinguistic information in author profiling". In: *Mexican Conference on Pattern Recognition*. Springer. 411–421.

Hickman, L., S. Thapa, L. Tay, M. Cao, and P. Srinivasan. (2022). "Text preprocessing for text mining in organizational research: Review and recommendations". *Organizational Research Methods*. 25(1): 114–146.

Hinton, G. E., A. Krizhevsky, and S. D. Wang. (2011). "Transforming auto-encoders". In: *International conference on artificial neural networks*. Springer. 44–51.

Hinton, G. E., S. Sabour, and N. Frosst. (2018). "Matrix capsules with EM routing". In: *International conference on learning representations*.

Ho, T. K. (1998). "The random subspace method for constructing decision forests". *IEEE transactions on pattern analysis and machine intelligence*. 20(8): 832–844.

Hogenboom, A., D. Bal, F. Frasincar, M. Bal, F. De Jong, and U. Kaymak. (2013). "Exploiting emoticons in sentiment analysis". In: *Proceedings of the 28th annual ACM symposium on applied computing.* 703–710.

Indra, S., L. Wikarsa, and R. Turang. (2016). "Using logistic regression method to classify tweets into the selected topics". In: *2016 international conference on advanced computer science and information systems (icacsis).* IEEE. 385–390.

Islam, R. and I. Ahmed. (2024). "Gemini-the most powerful LLM: Myth or Truth". In: *2024 5th Information Communication Technologies Conference (ICTC).* IEEE. 303–308.

Iyyer, M., V. Manjunatha, J. Boyd-Graber, and H. Daumé III. (2015). "Deep unordered composition rivals syntactic methods for text classification". In: *Proceedings of the 53rd annual meeting of the association for computational linguistics and the 7th international joint conference on natural language processing (volume 1: Long papers).* 1681–1691.

Jiang, A. Q., A. Sablayrolles, A. Mensch, C. Bamford, D. S. Chaplot, D. d. l. Casas, F. Bressand, G. Lengyel, G. Lample, L. Saulnier, *et al.* (2023). "Mistral 7B". *arXiv preprint arXiv:2310.06825.*

Jiang, S., G. Pang, M. Wu, and L. Kuang. (2012). "An improved K-nearest-neighbor algorithm for text categorization". *Expert Systems with Applications.* 39(1): 1503–1509.

Jianqiang, Z. and G. Xiaolin. (2017). "Comparison research on text pre-processing methods on twitter sentiment analysis". *IEEE Access.* 5: 2870–2879.

Jin, D., Z. Jin, J. T. Zhou, and P. Szolovits. (2020). "Is BERT Really Robust? A Strong Baseline for Natural Language Attack on Text Classification and Entailment". In: *The Thirty-Fourth AAAI Conference on Artificial Intelligence, AAAI 2020, The Thirty-Second Innovative Applications of Artificial Intelligence Conference, IAAI 2020, The Tenth AAAI Symposium on Educational Advances in Artificial Intelligence, EAAI 2020, New York, NY, USA, February 7-12, 2020.* AAAI Press. 8018–8025.

Joachims, T. (1998). "Text categorization with support vector machines: Learning with many relevant features". In: *European conference on machine learning*. Springer. 137–142.

Joachims, T. (1999). "Transductive Inference for Text Classification using Support Vector Machines". In: *Proceedings of the Sixteenth International Conference on Machine Learning (ICML 1999), Bled, Slovenia, June 27 - 30, 1999*. Ed. by I. Bratko and S. Dzeroski. Morgan Kaufmann. 200–209.

Joachims, T. (2002). "A statistical learning model of text classification for SVMs". In: *Learning to Classify Text Using Support Vector Machines*. Springer. 45–74.

Johnson, D. E., F. J. Oles, T. Zhang, and T. Goetz. (2002). "A decision-tree-based symbolic rule induction system for text categorization". *IBM Systems Journal*. 41(3): 428–437.

Johnson, R. and T. Zhang. (2016). "Supervised and semi-supervised text categorization using LSTM for region embeddings". In: *International Conference on Machine Learning*. PMLR. 526–534.

Joulin, A., E. Grave, P. Bojanowski, M. Douze, H. Jégou, and T. Mikolov. (2016). "Fasttext. zip: Compressing text classification models". *arXiv preprint arXiv:1612.03651*.

Joyce, J. M. (2011). "Kullback-leibler divergence". In: *International encyclopedia of statistical science*. Springer. 720–722.

Jurafsky, D. and J. H. Martin. (2009). *Speech and language processing: an introduction to natural language processing, computational linguistics, and speech recognition, 2nd Edition. Prentice Hall series in artificial intelligence*. Prentice Hall, Pearson Education International. ISBN: 9780135041963.

Kadhim, A. I. (2018). "An Evaluation of Preprocessing Techniques for Text Classification". *International Journal of Computer Science and Information Security (IJCSIS)*. 16(6).

Kathuria, A., A. Gupta, and R. Singla. (2021). "A Review of Tools and Techniques for Preprocessing of Textual Data". *Computational Methods and Data Engineering*: 407–422.

Kenny, E. M., C. Ford, M. Quinn, and M. T. Keane. (2021). "Explaining black-box classifiers using post-hoc explanations-by-example: The effect of explanations and error-rates in XAI user studies". *Artificial Intelligence.* 294: 103459.

Ketsbaia, L., B. Issac, and X. Chen. (2020). "Detection of hate tweets using machine learning and deep learning". In: *2020 IEEE 19th International Conference on Trust, Security and Privacy in Computing and Communications (TrustCom).* IEEE. 751–758.

Kim, Y. (2014). "Convolutional Neural Networks for Sentence Classification". In: *Proceedings of the 2014 Conference on Empirical Methods in Natural Language Processing (EMNLP).* Doha, Qatar: Association for Computational Linguistics. 1746–1751. DOI: 10.3115/v1/D14-1181.

Kim, Y., Y. Jernite, D. Sontag, and A. M. Rush. (2016). "Character-aware neural language models". In: *Thirtieth AAAI conference on artificial intelligence.*

Kipf, T. N. and M. Welling. (2017). "Semi-Supervised Classification with Graph Convolutional Networks". In: *5th International Conference on Learning Representations, ICLR 2017, Toulon, France, April 24-26, 2017, Conference Track Proceedings.* OpenReview.net.

Kong, A., S. Zhao, H. Chen, Q. Li, Y. Qin, R. Sun, X. Zhou, E. Wang, and X. Dong. (2024). "Better Zero-Shot Reasoning with Role-Play Prompting". In: *Proceedings of the 2024 Conference of the North American Chapter of the Association for Computational Linguistics: Human Language Technologies (Volume 1: Long Papers), NAACL 2024, Mexico City, Mexico, June 16-21, 2024.* Ed. by K. Duh, H. Gómez-Adorno, and S. Bethard. Association for Computational Linguistics. 4099–4113. DOI: 10.18653/V1/2024.NAACL-LONG.228.

Koopman, C. and A. Wilhelm. (2020). "The effect of preprocessing on short document clustering". *Archives of Data Science, Series A.* 6(1): 01.

Kouloumpis, E., T. Wilson, and J. Moore. (2011). "Twitter sentiment analysis: The good the bad and the omg!" In: *Proceedings of the international AAAI conference on web and social media.* Vol. 5. 538–541.

Kowsari, K., D. E. Brown, M. Heidarysafa, K. J. Meimandi, M. S. Gerber, and L. E. Barnes. (2017). "Hdltex: Hierarchical deep learning for text classification". In: *2017 16th IEEE international conference on machine learning and applications (ICMLA)*. IEEE. 364–371.

Kowsari, K., K. Jafari Meimandi, M. Heidarysafa, S. Mendu, L. Barnes, and D. Brown. (2019). "Text classification algorithms: A survey". *Information*. 10(4): 150.

Kudo, T. (2018). "Subword Regularization: Improving Neural Network Translation Models with Multiple Subword Candidates". In: *Proceedings of the 56th Annual Meeting of the Association for Computational Linguistics (Volume 1: Long Papers)*. 66–75.

Kumar, P. and L. Dhinesh Babu. (2019). "Novel text preprocessing framework for sentiment analysis". In: *Smart intelligent computing and applications*. Springer. 309–317.

Kunilovskaya, M. and A. Plum. (2021). "Text Preprocessing and its Implications in a Digital Humanities Project". In: *Proceedings of the Student Research Workshop Associated with RANLP 2021*. 85–93.

Kurniasih, A. and L. P. Manik. (2022). "On the Role of Text Preprocessing in BERT Embedding-based DNNs for Classifying Informal Texts". *International Journal of Advanced Computer Science and Applications*. 13(6): 927–934. DOI: 10.14569/IJACSA.2022.01306109.

Kuznetsov, I. and I. Gurevych. (2018). "From text to lexicon: Bridging the gap between word embeddings and lexical resources". In: *Proceedings of the 27th International Conference on Computational Linguistics*. 233–244.

Lample, G., M. Ballesteros, S. Subramanian, K. Kawakami, and C. Dyer. (2016). "Neural architectures for named entity recognition". In: *Proceedings of NAACL-HLT*.

Lan, Z., M. Chen, S. Goodman, K. Gimpel, P. Sharma, and R. Soricut. (2020). "ALBERT: A Lite BERT for Self-supervised Learning of Language Representations". In: *8th International Conference on Learning Representations, ICLR 2020, Addis Ababa, Ethiopia, April 26-30, 2020*. OpenReview.net.

Landauer, T. K. and S. T. Dumais. (1997). "A solution to Plato's problem: The latent semantic analysis theory of acquisition, induction, and knowledge." *Psychological review*. 104(2): 211.

Le, Q. and T. Mikolov. (2014). "Distributed representations of sentences and documents". In: *International conference on machine learning*. PMLR. 1188–1196.

Lee, K., L. He, M. Lewis, and L. Zettlemoyer. (2017). "End-to-end neural coreference resolution". In: *Proceedings of the 2017 Conference on Empirical Methods in Natural Language Processing*. 188–197.

Lehmann, J., R. Isele, M. Jakob, A. Jentzsch, D. Kontokostas, P. N. Mendes, S. Hellmann, M. Morsey, P. Van Kleef, S. Auer, *et al.* (2015). "Dbpedia–a large-scale, multilingual knowledge base extracted from wikipedia". *Semantic web*. 6(2): 167–195.

Leopold, E. and J. Kindermann. (2002). "Text categorization with support vector machines. How to represent texts in input space?" *Machine Learning*. 46(1): 423–444.

Leslie, C., E. Eskin, and W. S. Noble. (2001). "The spectrum kernel: A string kernel for SVM protein classification". In: *Biocomputing 2002*. World Scientific. 564–575.

Lewis, P. S. H., E. Perez, A. Piktus, F. Petroni, V. Karpukhin, N. Goyal, H. Küttler, M. Lewis, W. Yih, T. Rocktäschel, S. Riedel, and D. Kiela. (2020). "Retrieval-Augmented Generation for Knowledge-Intensive NLP Tasks". In: *Advances in Neural Information Processing Systems 33: Annual Conference on Neural Information Processing Systems 2020, NeurIPS 2020, December 6-12, 2020, virtual*. Ed. by H. Larochelle, M. Ranzato, R. Hadsell, M. Balcan, and H. Lin.

Li, Q., H. Peng, J. Li, C. Xia, R. Yang, L. Sun, P. S. Yu, and L. He. (2020). "A survey on text classification: From shallow to deep learning". *arXiv preprint arXiv:2008.00364*.

Li, X. and Y. Guo. (2013). "Active Learning with Multi-Label SVM Classification". In: *IJCAI 2013, Proceedings of the 23rd International Joint Conference on Artificial Intelligence, Beijing, China, August 3-9, 2013*. Ed. by F. Rossi. IJCAI/AAAI. 1479–1485.

Li, X. and W. Lam. (2017). "Deep multi-task learning for aspect term extraction with memory interaction". In: *Proceedings of the 2017 conference on empirical methods in natural language processing*. 2886–2892.

Lin, C. and Y. He. (2009). "Joint sentiment/topic model for sentiment analysis". In: *Proceedings of the 18th ACM conference on Information and knowledge management.* 375–384.

Lison, P. and A. Kutuzov. (2017). "Redefining Context Windows for Word Embedding Models: An Experimental Study". In: *Proceedings of the 21st Nordic Conference on Computational Linguistics.* 284–288.

Liu, H., Z. Zhao, J. Wang, H. Kamarthi, and B. A. Prakash. (2024a). "LSTPrompt: Large Language Models as Zero-Shot Time Series Forecasters by Long-Short-Term Prompting". In: *Findings of the Association for Computational Linguistics, ACL 2024, Bangkok, Thailand and virtual meeting, August 11-16, 2024.* Ed. by L. Ku, A. Martins, and V. Srikumar. Association for Computational Linguistics. 7832–7840. DOI: 10.18653/V1/2024.FINDINGS-ACL.466.

Liu, J., W.-C. Chang, Y. Wu, and Y. Yang. (2017). "Deep learning for extreme multi-label text classification". In: *Proceedings of the 40th international ACM SIGIR conference on research and development in information retrieval.* 115–124.

Liu, N. F., K. Lin, J. Hewitt, A. Paranjape, M. Bevilacqua, F. Petroni, and P. Liang. (2024b). "Lost in the middle: How language models use long contexts". *Transactions of the Association for Computational Linguistics.* 12: 157–173. DOI: 10.1162/TACL\_A\_00638.

Liu, P., X. Qiu, X. Chen, S. Wu, and X.-J. Huang. (2015). "Multi-timescale long short-term memory neural network for modelling sentences and documents". In: *Proceedings of the 2015 conference on empirical methods in natural language processing.* 2326–2335.

Liu, P., X. Qiu, and X. Huang. (2016). "Recurrent Neural Network for Text Classification with Multi-Task Learning". In: *Proceedings of the Twenty-Fifth International Joint Conference on Artificial Intelligence, IJCAI 2016, New York, NY, USA, 9-15 July 2016.* Ed. by S. Kambhampati. IJCAI/AAAI Press. 2873–2879.

Liu, X., Y. Shen, K. Duh, and J. Gao. (2018). "Stochastic Answer Networks for Machine Reading Comprehension". In: *Proceedings of the 56th Annual Meeting of the Association for Computational Linguistics, ACL 2018, Melbourne, Australia, July 15-20, 2018, Volume 1: Long Papers*. Ed. by I. Gurevych and Y. Miyao. Association for Computational Linguistics. 1694–1704. DOI: 10.18653/v1/P18-1157.

Liu, Y., M. Ott, N. Goyal, J. Du, M. Joshi, D. Chen, O. Levy, M. Lewis, L. Zettlemoyer, and V. Stoyanov. (2019). "Roberta: A robustly optimized bert pretraining approach". *arXiv preprint arXiv:1907.11692*.

Liu, Z., X. Lv, K. Liu, and S. Shi. (2010). "Study on SVM compared with the other text classification methods". In: *2010 Second international workshop on education technology and computer science*. Vol. 1. IEEE. 219–222.

Lo, R. T.-W., B. He, and I. Ounis. (2005). "Automatically building a stopword list for an information retrieval system". In: *Journal on Digital Information Management: Special Issue on the 5th Dutch-Belgian Information Retrieval Workshop (DIR)*. Vol. 5. 17–24.

Lombardo, A., G. Morabito, S. Quattropani, C. Ricci, M. Siino, and I. Tinnirello. (2024). "AI-GeneSI: Exploiting generative AI for autonomous generation of the southbound interface in the IoT". In: *2024 IEEE 10th World Forum on Internet of Things (WF-IoT)*. 1–7. DOI: 10.1109/WF-IoT62078.2024.10811300.

Lomonaco, F., G. Donabauer, and M. Siino. (2022). "COURAGE at CheckThat! 2022: Harmful Tweet Detection using Graph Neural Networks and ELECTRA". In: *Proceedings of the Working Notes of CLEF 2022 - Conference and Labs of the Evaluation Forum (CLEF)* (Bologna, Italy, Sept. 5–8, 2022). Ed. by G. Faggioli, N. Ferro, A. Hanbury, and M. Potthast. *CEUR Workshop Proceedings*. No. 3180. Aachen. 573–583.

Lovins, J. B. (1968). "Development of a stemming algorithm." *Mechanical Translation and Computational Linguistics*. 11(1-2): 22–31.

Loza Mencía, E. and J. Fürnkranz. (2008). "Efficient pairwise multilabel classification for large-scale problems in the legal domain". In: *Joint European Conference on Machine Learning and Knowledge Discovery in Databases*. Springer. 50–65.

Luhn, H. P. (1960). "Key word-in-context index for technical literature (kwic index)". *American documentation.* 11(4): 288–295.

Maas, A. L., R. E. Daly, P. T. Pham, D. Huang, A. Y. Ng, and C. Potts. (2011). "Learning Word Vectors for Sentiment Analysis". In: *Proceedings of the 49th Annual Meeting of the Association for Computational Linguistics: Human Language Technologies.* Portland, Oregon, USA: Association for Computational Linguistics. 142–150.

Magerman, D. M. (1995). "Statistical Decision-Tree Models for Parsing". In: *33rd Annual Meeting of the Association for Computational Linguistics.* 276–283.

Makrehchi, M. and M. S. Kamel. (2008). "Automatic extraction of domain-specific stopwords from labeled documents". In: *Advances in Information Retrieval: 30th European Conference on IR Research, ECIR 2008, Glasgow, UK, March 30-April 3, 2008. Proceedings 30.* Springer. 222–233.

Mangione, S., M. Siino, and G. Garbo. (2022). "Improving Irony and Stereotype Spreaders Detection using Data Augmentation and Convolutional Neural Network". In: *Proceedings of the Working Notes of CLEF 2022 - Conference and Labs of the Evaluation Forum (CLEF)* (Bologna, Italy, Sept. 5–8, 2022). Ed. by G. Faggioli, N. Ferro, A. Hanbury, and M. Potthast. *CEUR Workshop Proceedings.* No. 3180. Aachen. 2585–2593.

Manning, C. D., P. Raghavan, and H. Schütze. (2008). *Introduction to information retrieval.* Cambridge University Press. ISBN: 978-0-521-86571-5. DOI: 10.1017/CBO9780511809071.

Manning, C. D., H. Schütze, and G. Weikurn. (2002). "Foundations of Statistical Natural Language Processing". *SIGMOD Record.* 31(3): 37–38.

Marcus, G. and E. Davis. (2019). *Rebooting AI: Building artificial intelligence we can trust.* Vintage.

Matthews, B. W. (1975). "Comparison of the predicted and observed secondary structure of T4 phage lysozyme". *Biochimica et Biophysica Acta (BBA)-Protein Structure.* 405(2): 442–451.

McCallum, A. and K. Nigam. (1998). "A comparison of event models for naive bayes text classification". In: *AAAI-98 workshop on learning for text categorization.* Vol. 752. Citeseer. 41–48.

McCann, B., J. Bradbury, C. Xiong, and R. Socher. (2017). "Learned in Translation: Contextualized Word Vectors". In: *Advances in Neural Information Processing Systems 30: Annual Conference on Neural Information Processing Systems 2017, December 4-9, 2017, Long Beach, CA, USA*. Ed. by I. Guyon, U. von Luxburg, S. Bengio, H. M. Wallach, R. Fergus, S. V. N. Vishwanathan, and R. Garnett. 6294–6305.

McNamee, P. and J. Mayfield. (2004). "Character n-gram tokenization for European language text retrieval". *Information retrieval.* 7(1): 73–97.

Melamud, O., J. Goldberger, and I. Dagan. (2016). "context2vec: Learning Generic Context Embedding with Bidirectional LSTM". In: *Proceedings of the 20th SIGNLL Conference on Computational Natural Language Learning.* Berlin, Germany: Association for Computational Linguistics. 51–61. DOI: 10.18653/v1/K16-1006.

Miculicich, L., D. Ram, N. Pappas, and J. Henderson. (2018). "Document-Level Neural Machine Translation with Hierarchical Attention Networks". In: *Proceedings of the 2018 Conference on Empirical Methods in Natural Language Processing, Brussels, Belgium, October 31 - November 4, 2018.* Ed. by E. Riloff, D. Chiang, J. Hockenmaier, and J. Tsujii. Association for Computational Linguistics. 2947–2954. DOI: 10.18653/v1/d18-1325.

Mihalcea, R. and P. Tarau. (2004). "Textrank: Bringing order into text". In: *Proceedings of the 2004 conference on empirical methods in natural language processing.* 404–411.

Mikolov, T., K. Chen, G. Corrado, and J. Dean. (2013a). "Efficient Estimation of Word Representations in Vector Space". In: *1st International Conference on Learning Representations, ICLR 2013, Scottsdale, Arizona, USA, May 2-4, 2013, Workshop Track Proceedings.* Ed. by Y. Bengio and Y. LeCun.

Mikolov, T., I. Sutskever, K. Chen, G. S. Corrado, and J. Dean. (2013b). "Distributed Representations of Words and Phrases and their Compositionality". In: *Advances in Neural Information Processing Systems 26: 27th Annual Conference on Neural Information Processing Systems 2013. Proceedings of a meeting held December 5-8, 2013, Lake Tahoe, Nevada, United States.* Ed. by C. J. C. Burges, L. Bottou, Z. Ghahramani, and K. Q. Weinberger. 3111–3119.

Miller, G. A. (1995). "WordNet: a lexical database for English". *Communications of the ACM.* 38(11): 39–41.

Minaee, S., N. Kalchbrenner, E. Cambria, N. Nikzad, M. Chenaghlu, and J. Gao. (2021). "Deep learning–based text classification: a comprehensive review". *ACM Computing Surveys (CSUR).* 54(3): 1–40.

Mitra, V., C.-J. Wang, and S. Banerjee. (2007). "Text classification: A least square support vector machine approach". *Applied soft computing.* 7(3): 908–914.

Miyato, T., S.-i. Maeda, M. Koyama, and S. Ishii. (2018). "Virtual adversarial training: a regularization method for supervised and semi-supervised learning". *IEEE transactions on pattern analysis and machine intelligence.* 41(8): 1979–1993.

Mohammad, F. (2018). "Is preprocessing of text really worth your time for online comment classification?" *arXiv preprint arXiv:1806.02908.*

Moral, C., A. de Antonio, R. Imbert, and J. Ramírez. (2014). "A survey of stemming algorithms in information retrieval." *Information Research: An International Electronic Journal.* 19(1).

Mubarok, M. S., Adiwijaya, and M. D. Aldhi. (2017). "Aspect-based sentiment analysis to review products using Naı̈ve Bayes". In: *AIP Conference Proceedings.* Vol. 1867. AIP Publishing LLC. 020060.

Mueller, J. and A. Thyagarajan. (2016). "Siamese recurrent architectures for learning sentence similarity". In: *Proceedings of the AAAI conference on artificial intelligence.* Vol. 30.

Mullen, T. and R. Malouf. (2006). "A Preliminary Investigation into Sentiment Analysis of Informal Political Discourse." In: *AAAI spring symposium: computational approaches to analyzing weblogs.* 159–162.

Naseem, U., I. Razzak, and P. W. Eklund. (2021). "A survey of pre-processing techniques to improve short-text quality: a case study on hate speech detection on twitter". *Multimedia Tools and Applications*. 80(28): 35239–35266.

Nowak, J., A. Taspinar, and R. Scherer. (2017). "LSTM recurrent neural networks for short text and sentiment classification". In: *International Conference on Artificial Intelligence and Soft Computing*. Springer. 553–562.

Paice, C. D. (1990). "Another Stemmer". *SIGIR Forum*. 24(3): 56–61. ISSN: 0163-5840. DOI: 10.1145/101306.101310.

Palmer, D. D. (1997). "A trainable rule-based algorithm for word segmentation". In: *35th Annual Meeting of the Association for Computational Linguistics and 8th Conference of the European Chapter of the Association for Computational Linguistics*. 321–328.

Pang, B., L. Lee, and S. Vaithyanathan. (2002). "Thumbs up? Sentiment Classification using Machine Learning Techniques". In: *Proceedings of the 2002 Conference on Empirical Methods in Natural Language Processing (EMNLP 2002)*. 79–86.

Pardo, F. M. R., A. Giachanou, B. Ghanem, and P. Rosso. (2020). "Overview of the 8th Author Profiling Task at PAN 2020: Profiling Fake News Spreaders on Twitter". In: *Working Notes of CLEF 2020 - Conference and Labs of the Evaluation Forum, Thessaloniki, Greece, September 22-25, 2020*. Ed. by L. Cappellato, C. Eickhoff, N. Ferro, and A. Névéol. Vol. 2696. *CEUR Workshop Proceedings*. CEUR-WS.org.

Pecar, S., M. Simko, and M. Bielikova. (2018). "Sentiment Analysis of Customer Reviews: Impact of Text Pre-Processing". In: *2018 World Symposium on Digital Intelligence for Systems and Machines (DISA)*. 251–256. DOI: 10.1109/DISA.2018.8490619.

Peng, H., J. Li, Y. He, Y. Liu, M. Bao, L. Wang, Y. Song, and Q. Yang. (2018). "Large-scale hierarchical text classification with recursively regularized deep graph-cnn". In: *Proceedings of the 2018 world wide web conference*. 1063–1072.

Peng, H., J. Li, S. Wang, L. Wang, Q. Gong, R. Yang, B. Li, S. Y. Philip, and L. He. (2019). "Hierarchical taxonomy-aware and attentional graph capsule RCNNs for large-scale multi-label text classification". *IEEE Transactions on Knowledge and Data Engineering.* 33(6): 2505–2519.

Peng, T., W. Zuo, and F. He. (2008). "SVM based adaptive learning method for text classification from positive and unlabeled documents". *Knowledge and Information Systems.* 16(3): 281–301.

Pennington, J., R. Socher, and C. D. Manning. (2014). "Glove: Global vectors for word representation". In: *Proceedings of the 2014 conference on empirical methods in natural language processing (EMNLP).* 1532–1543.

Pereira, J. M., M. Basto, and A. F. da Silva. (2016). "The logistic lasso and ridge regression in predicting corporate failure". *Procedia Economics and Finance.* 39: 634–641.

Pérez-Almendros, C., L. E. Anke, and S. Schockaert. (2020). "Don't Patronize Me! An Annotated Dataset with Patronizing and Condescending Language towards Vulnerable Communities". In: *Proceedings of the 28th International Conference on Computational Linguistics.* 5891–5902.

Pérez-Almendros, C., L. E. Anke, and S. Schockaert. (2022). "SemEval-2022 task 4: Patronizing and condescending language detection". In: *Proceedings of the 16th International Workshop on Semantic Evaluation (SemEval-2022).* Association for Computational Linguistics. 298–307.

Peters, B. and A. F. Martins. (2024). "Did Translation Models Get More Robust Without Anyone Even Noticing?" *arXiv preprint arXiv:2403.03923.*

Peters, M. E., M. Neumann, M. Iyyer, M. Gardner, C. Clark, K. Lee, and L. Zettlemoyer. (2018). "Deep Contextualized Word Representations". In: *Proceedings of the 2018 Conference of the North American Chapter of the Association for Computational Linguistics: Human Language Technologies, NAACL-HLT 2018, New Orleans, Louisiana, USA, June 1-6, 2018, Volume 1 (Long Papers).* Ed. by M. A. Walker, H. Ji, and A. Stent. Association for Computational Linguistics. 2227–2237. DOI: 10.18653/v1/n18-1202.

Peters, M., M. Neumann, M. Iyyer, M. Gardner, C. Clark, K. Lee, and L. Zettlemoyer. (1802). "Deep contextualized word representations. arXiv 2018". *arXiv preprint arXiv:1802.05365*. 12.

Petrović, D. and M. Stanković. (2019). "The influence of text preprocessing methods and tools on calculating text similarity". *Facta Universitatis, Series: Mathematics and Informatics*. 34: 973–994.

Porter, M. F. (1980). "An algorithm for suffix stripping". *Program: electronic library and information systems*. 14(3): 130–137.

Pouyanfar, S., S. Sadiq, Y. Yan, H. Tian, Y. Tao, M. P. Reyes, M.-L. Shyu, S.-C. Chen, and S. S. Iyengar. (2018). "A survey on deep learning: Algorithms, techniques, and applications". *ACM Computing Surveys (CSUR)*. 51(5): 1–36.

Pradha, S., M. N. Halgamuge, and N. T. Q. Vinh. (2019). "Effective text data preprocessing technique for sentiment analysis in social media data". In: *2019 11th international conference on knowledge and systems engineering (KSE)*. IEEE. 1–8.

*Proceedings of the Working Notes of CLEF 2022 - Conference and Labs of the Evaluation Forum (CLEF)*. (2022) (Bologna, Italy, Sept. 5–8, 2022). Ed. by G. Faggioli, N. Ferro, A. Hanbury, and M. Potthast. *CEUR Workshop Proceedings*. No. 3180. Aachen.

Qu, Z., X. Song, S. Zheng, X. Wang, X. Song, and Z. Li. (2018). "Improved Bayes method based on TF-IDF feature and grade factor feature for chinese information classification". In: *2018 IEEE International Conference on Big Data and Smart Computing (BigComp)*. IEEE. 677–680.

Quinlan, J. R. (1986). "Induction of decision trees". *Machine learning*. 1(1): 81–106.

Radford, A., J. Wu, R. Child, D. Luan, D. Amodei, I. Sutskever, *et al.* (2019). "Language models are unsupervised multitask learners". *OpenAI blog*. 1(8): 9.

Raffel, C., N. Shazeer, A. Roberts, K. Lee, S. Narang, M. Matena, Y. Zhou, W. Li, and P. J. Liu. (2020). "Exploring the Limits of Transfer Learning with a Unified Text-to-Text Transformer". *Journal of Machine Learning Research*. 21(140): 1–67.

Rangel, F., G. L. D. la Peña Sarracén, B. Chulvi, E. Fersini, and P. Rosso. (2021). "Profiling Hate Speech Spreaders on Twitter Task at PAN 2021". In: *Proceedings of the Working Notes of CLEF 2021 - Conference and Labs of the Evaluation Forum, Bucharest, Romania, September 21st - to - 24th, 2021.* Ed. by G. Faggioli, N. Ferro, A. Joly, M. Maistro, and F. Piroi. Vol. 2936. *CEUR Workshop Proceedings.* CEUR-WS.org. 1772–1789.

Raschka, S. (2014). "Naive bayes and text classification i-introduction and theory". *arXiv preprint arXiv:1410.5329.*

Rastogi, R. and K. Shim. (2000). "PUBLIC: A decision tree classifier that integrates building and pruning". *Data Mining and Knowledge Discovery.* 4(4): 315–344.

Rathje, S., D.-M. Mirea, I. Sucholutsky, R. Marjieh, C. E. Robertson, and J. J. Van Bavel. (2024). "GPT is an effective tool for multilingual psychological text analysis". *Proceedings of the National Academy of Sciences.* 121(34): e2308950121.

Resyanto, F., Y. Sibaroni, and A. Romadhony. (2019). "Choosing the most optimum text preprocessing method for sentiment analysis: Case: iPhone Tweets". In: *2019 Fourth International Conference on Informatics and Computing (ICIC).* IEEE. 1–5.

Rijsbergen, C. J. van. (1979). "Information Retrieval".

Rosid, M. A., A. S. Fitrani, I. R. I. Astutik, N. I. Mulloh, and H. A. Gozali. (2020). "Improving text preprocessing for student complaint document classification using sastrawi". In: *IOP Conference Series: Materials Science and Engineering.* Vol. 874. IOP Publishing. 012017.

Sabour, S., N. Frosst, and G. E. Hinton. (2017). "Dynamic routing between capsules". *Advances in neural information processing systems.* 30.

Sagolla, D. (2009). *140 characters: A style guide for the short form.* John Wiley & Sons.

Saif, H., M. Fernandez, Y. He, and H. Alani. (2014). "On Stopwords, Filtering and Data Sparsity for Sentiment Analysis of Twitter". In: *Proceedings of the Ninth International Conference on Language Resources and Evaluation (LREC'14).* 810–817.

Sanh, V., L. Debut, J. Chaumond, and T. Wolf. (2019). "DistilBERT, a distilled version of BERT: smaller, faster, cheaper and lighter". *arXiv preprint arXiv:1910.01108.*

Schlag, I., P. Smolensky, R. Fernandez, N. Jojic, J. Schmidhuber, and J. Gao. (2019). "Enhancing the transformer with explicit relational encoding for math problem solving". *arXiv preprint arXiv:1910.06611.*

Schuster, M. and K. Nakajima. (2012). "Japanese and korean voice search". In: *2012 IEEE international conference on acoustics, speech and signal processing (ICASSP).* IEEE. 5149–5152.

Senette, C., M. Siino, and M. Tesconi. (2024). "User Identity Linkage on Social Networks: A Review of Modern Techniques and Applications". *IEEE Access.* 12: 171241–171268. DOI: 10.1109/ACCESS.2024.3500374.

Sennrich, R., B. Haddow, and A. Birch. (2016). "Neural Machine Translation of Rare Words with Subword Units". In: *Proceedings of the 54th Annual Meeting of the Association for Computational Linguistics (Volume 1: Long Papers).* 1715–1725.

Shah, K., H. Patel, D. Sanghvi, and M. Shah. (2020). "A comparative analysis of logistic regression, random forest and KNN models for the text classification". *Augmented Human Research.* 5(1): 1–16.

Siino, M. (2024a). "All-Mpnet at SemEval-2024 Task 1: Application of Mpnet for Evaluating Semantic Textual Relatedness". In: *Proceedings of the 18th International Workshop on Semantic Evaluation (SemEval-2024).* Ed. by A. K. Ojha, A. S. Doğruöz, H. Tayyar Madabushi, G. Da San Martino, S. Rosenthal, and A. Rosá. Mexico City, Mexico: Association for Computational Linguistics. 379–384. DOI: 10.18653/v1/2024.semeval-1.59. URL: https://aclanthology.org/2024.semeval-1.59/.

Siino, M. (2024b). "BadRock at SemEval-2024 Task 8: DistilBERT to Detect Multigenerator, Multidomain and Multilingual Black-Box Machine-Generated Text". In: *Proceedings of the 18th International Workshop on Semantic Evaluation (SemEval-2024).* Ed. by A. K. Ojha, A. S. Doğruöz, H. Tayyar Madabushi, G. Da San Martino, S. Rosenthal, and A. Rosá. Mexico City, Mexico: Association for Computational Linguistics. 239–245. DOI: 10.18653/v1/2024.semeval-1.37.

Siino, M. (2024c). "T5-Medical at SemEval-2024 Task 2: Using T5 Medical Embedding for Natural Language Inference on Clinical Trial Data". In: *Proceedings of the 18th International Workshop on Semantic Evaluation (SemEval-2024)*. Ed. by A. K. Ojha, A. S. Doğruöz, H. Tayyar Madabushi, G. Da San Martino, S. Rosenthal, and A. Rosá. Mexico City, Mexico: Association for Computational Linguistics. 40–46. DOI: 10.18653/v1/2024.semeval-1.7.

Siino, M., E. Di Nuovo, I. Tinnirello, and M. La Cascia. (2021). "Detection of hate speech spreaders using convolutional neural networks". In: *Proceedings of the Working Notes of CLEF 2021 - Conference and Labs of the Evaluation Forum, Bucharest, Romania, September 21st - to - 24th, 2021*. Ed. by G. Faggioli, N. Ferro, A. Joly, M. Maistro, and F. Piroi. Vol. 2936. *CEUR Workshop Proceedings*. CEUR-WS.org. 2126–2136.

Siino, M., E. Di Nuovo, I. Tinnirello, and M. La Cascia. (2022a). "Fake News Spreaders Detection: Sometimes Attention Is Not All You Need". *Information*. 13(9): 426.

Siino, M., M. Falco, D. Croce, and P. Rosso. (2025). "Exploring LLMs Applications in Law: A Literature Review on Current Legal NLP Approaches". *IEEE Access*. 13: 18253–18276. DOI: 10.1109/ACCESS.2025.3533217.

Siino, M., F. Giuliano, and I. Tinnirello. (2024a). "LLM Application for Knowledge Extraction from Networking Log Files". In: *2024 4th International Conference on Electrical, Computer, Communications and Mechatronics Engineering (ICECCME)*. 01–06. DOI: 10.1109/ICECCME62383.2024.10796967.

Siino, M., M. La Cascia, and I. Tinnirello. (2020). "WhoSNext: Recommending Twitter Users to Follow Using a Spreading Activation Network Based Approach". In: *2020 International Conference on Data Mining Workshops (ICDMW)*. IEEE. 62–70.

Siino, M., M. La Cascia, and I. Tinnirello. (2022b). "McRock at SemEval-2022 Task 4: Patronizing and Condescending Language Detection using Multi-Channel CNN, Hybrid LSTM, DistilBERT and XLNet". In: *Proceedings of the 16th International Workshop on Semantic Evaluation (SemEval-2022)*. Seattle, United States: Association for Computational Linguistics. 409–417. DOI: 10.18653/v1/2022.semeval-1.55.

Siino, M., F. Lomonaco, and P. Rosso. (2024b). "Backtranslate what you are saying and I will tell who you are". *Expert Systems*. 41(8): e13568. DOI: https://doi.org/10.1111/exsy.13568.

Siino, M. and I. Tinnirello. (2023). "XLNet with Data Augmentation to Profile Cryptocurrency Influencers". In: *Working Notes of the Conference and Labs of the Evaluation Forum (CLEF 2023), Thessaloniki, Greece, September 18th to 21st, 2023*. Ed. by M. Aliannejadi, G. Faggioli, N. Ferro, and M. Vlachos. Vol. 3497. *CEUR Workshop Proceedings*. CEUR-WS.org. 2763–2771.

Siino, M. and I. Tinnirello. (2024a). "GPT Hallucination Detection Through Prompt Engineering". In: *Working Notes of the Conference and Labs of the Evaluation Forum (CLEF 2024), Grenoble, France, 9-12 September, 2024*. Ed. by G. Faggioli, N. Ferro, P. Galuscáková, and A. G. S. de Herrera. Vol. 3740. *CEUR Workshop Proceedings*. CEUR-WS.org. 712–721.

Siino, M. and I. Tinnirello. (2024b). "GPT Prompt Engineering for Scheduling Appliances Usage for Energy Cost Optimization". In: *2024 IEEE International Symposium on Measurements & Networking (M&N), Rome, Italy, July 2-5, 2024*. IEEE. 1–6. DOI: 10.1109/MN60932.2024.10615758.

Siino, M. and I. Tinnirello. (2024c). "Prompt engineering for identifying sexism using GPT Mistral 7B". In: *Working Notes of the Conference and Labs of the Evaluation Forum (CLEF 2024), Grenoble, France, 9-12 September, 2024*. Ed. by G. Faggioli, N. Ferro, P. Galuscáková, and A. G. S. de Herrera. Vol. 3740. *CEUR Workshop Proceedings*. CEUR-WS.org. 1228–1236.

Siino, M., I. Tinnirello, and M. L. Cascia. (2022c). "T100: A modern classic ensembler to profile irony and stereotype spreaders". In: *Proceedings of the Working Notes of CLEF 2022 - Conference and Labs of the Evaluation Forum (CLEF)* (Bologna, Italy, Sept. 5–8, 2022). Ed. by G. Faggioli, N. Ferro, A. Hanbury, and M. Potthast. *CEUR Workshop Proceedings*. No. 3180. Aachen. 2666–2674.

Siino, M., I. Tinnirello, and M. La Cascia. (2024c). "Is text preprocessing still worth the time? A comparative survey on the influence of popular preprocessing methods on Transformers and traditional classifiers". *Information Systems*. 121: 102342. ISSN: 0306-4379. DOI: https://doi.org/10.1016/j.is.2023.102342.

Siino, M., I. Tinnirello, and M. La Cascia. (2024d). "Is text preprocessing still worth the time? A comparative survey on the influence of popular preprocessing methods on Transformers and traditional classifiers". *Information Systems*. 121: 102342.

Singh, A., N. Singh, and S. Vatsal. (2024). "Robustness of llms to perturbations in text". *arXiv preprint arXiv:2407.08989*.

Singh, T. and M. Kumari. (2016). "Role of text pre-processing in twitter sentiment analysis". *Procedia Computer Science*. 89: 549–554.

Smelyakov, K., D. Karachevtsev, D. Kulemza, Y. Samoilenko, O. Patlan, and A. Chupryna. (2020). "Effectiveness of preprocessing algorithms for natural language processing applications". In: *2020 IEEE International Conference on Problems of Infocommunications. Science and Technology (PIC S&T)*. IEEE. 187–191.

Socher, R., A. Perelygin, J. Wu, J. Chuang, C. D. Manning, A. Y. Ng, and C. Potts. (2013). "Recursive deep models for semantic compositionality over a sentiment treebank". In: *Proceedings of the 2013 conference on empirical methods in natural language processing*. 1631–1642.

Soucy, P. and G. W. Mineau. (2001). "A simple KNN algorithm for text categorization". In: *Proceedings 2001 IEEE international conference on data mining*. IEEE. 647–648.

Srividhya, V. and R. Anitha. (2010). "Evaluating preprocessing techniques in text categorization". *International journal of computer science and application*. 47(11): 49–51.

Stehman, S. V. (1997). "Selecting and interpreting measures of thematic classification accuracy". *Remote sensing of Environment.* 62(1): 77–89.

Symeonidis, S., D. Effrosynidis, and A. Arampatzis. (2018). "A comparative evaluation of pre-processing techniques and their interactions for twitter sentiment analysis". *Expert Systems with Applications.* 110: 298–310.

Tai, K. S., R. Socher, and C. D. Manning. (2015). "Improved Semantic Representations From Tree-Structured Long Short-Term Memory Networks". In: *Proceedings of the 53rd Annual Meeting of the Association for Computational Linguistics and the 7th International Joint Conference on Natural Language Processing of the Asian Federation of Natural Language Processing, ACL 2015, July 26-31, 2015, Beijing, China, Volume 1: Long Papers.* The Association for Computer Linguistics. 1556–1566. DOI: 10.3115/v1/p15-1150.

Taira, H. and M. Haruno. (1999). "Feature Selection in SVM Text Categorization". In: *Proceedings of the Sixteenth National Conference on Artificial Intelligence and Eleventh Conference on Innovative Applications of Artificial Intelligence, July 18-22, 1999, Orlando, Florida, USA.* Ed. by J. Hendler and D. Subramanian. AAAI Press / The MIT Press. 480–486.

Tan, L., H. Zhang, C. Clarke, and M. Smucker. (2015). "Lexical comparison between wikipedia and twitter corpora by using word embeddings". In: *Proceedings of the 53rd Annual Meeting of the Association for Computational Linguistics and the 7th International Joint Conference on Natural Language Processing (Volume 2: Short Papers).* 657–661.

Tan, S. (2005). "Neighbor-weighted k-nearest neighbor for unbalanced text corpus". *Expert Systems with Applications.* 28(4): 667–671.

Thelwall, M. (2017). "The Heart and soul of the web? Sentiment strength detection in the social web with SentiStrength". In: *Cyberemotions.* Springer. 119–134.

Toman, M., R. Tesar, and K. Jezek. (2006). "Influence of word normalization on text classification". *Proceedings of InSciT.* 4: 354–358.

Touvron, H., T. Lavril, G. Izacard, X. Martinet, M.-A. Lachaux, T. Lacroix, B. Rozière, N. Goyal, Y. Jernite, E. Grave, and G. Lample. (2023). "LLaMA: Open and Efficient Foundation Language Models". *arXiv preprint arXiv:2302.13971.* URL: https://arxiv.org/abs/2302.13971.

Uysal, A. K. and S. Gunal. (2014). "The impact of preprocessing on text classification". *Information processing & management.* 50(1): 104–112.

Vaswani, A., N. Shazeer, N. Parmar, J. Uszkoreit, L. Jones, A. N. Gomez, L. Kaiser, and I. Polosukhin. (2017). "Attention is All you Need". In: *Advances in Neural Information Processing Systems 30: Annual Conference on Neural Information Processing Systems 2017, December 4-9, 2017, Long Beach, CA, USA.* Ed. by I. Guyon, U. von Luxburg, S. Bengio, H. M. Wallach, R. Fergus, S. V. N. Vishwanathan, and R. Garnett. 5998–6008.

Vateekul, P. and M. Kubat. (2009). "Fast induction of multiple decision trees in text categorization from large scale, imbalanced, and multi-label data". In: *2009 IEEE International Conference on Data Mining Workshops.* IEEE. 320–325.

Vijayarani, S., M. J. Ilamathi, and M. Nithya. (2015). "Preprocessing techniques for text mining-an overview". *International Journal of Computer Science & Communication Networks.* 5(1): 7–16.

Vijayarani, S. and R. Janani. (2016). "Text mining: open source tokenization tools-an analysis". *Advanced Computational Intelligence: An International Journal (ACII).* 3(1): 37–47.

Virmani, D. and S. Taneja. (2019). "A text preprocessing approach for efficacious information retrieval". In: *Smart innovations in communication and computational sciences.* Springer. 13–22.

Wan, S., Y. Lan, J. Guo, J. Xu, L. Pang, and X. Cheng. (2016). "A deep architecture for semantic matching with multiple positional sentence representations". In: *Proceedings of the AAAI Conference on Artificial Intelligence.* Vol. 30.

Wang, H. and J. A. Castanon. (2015). "Sentiment expression via emoticons on social media". In: *2015 ieee international conference on big data (big data).* IEEE. 2404–2408.

Wang, S. I. and C. D. Manning. (2012). "Baselines and bigrams: Simple, good sentiment and topic classification". In: *Proceedings of the 50th Annual Meeting of the Association for Computational Linguistics (Volume 2: Short Papers)*. 90–94.

Wang, Z., W. Hamza, and R. Florian. (2017). "Bilateral Multi-Perspective Matching for Natural Language Sentences". In: *Proceedings of the Twenty-Sixth International Joint Conference on Artificial Intelligence, IJCAI 2017, Melbourne, Australia, August 19-25, 2017*. Ed. by C. Sierra. ijcai.org. 4144–4150. DOI: 10.24963/ijcai.2017/579.

Wei, J., X. Wang, D. Schuurmans, M. Bosma, B. Ichter, F. Xia, E. H. Chi, Q. V. Le, and D. Zhou. (2022). "Chain-of-Thought Prompting Elicits Reasoning in Large Language Models". In: *Advances in Neural Information Processing Systems 35: Annual Conference on Neural Information Processing Systems 2022, NeurIPS 2022, New Orleans, LA, USA, November 28 - December 9, 2022*. Ed. by S. Koyejo, S. Mohamed, A. Agarwal, D. Belgrave, K. Cho, and A. Oh.

Wiegmann, M., B. Stein, and M. Potthast. (2020). "Overview of the Celebrity Profiling Task at PAN 2020". In: *Working Notes of CLEF 2020 - Conference and Labs of the Evaluation Forum, Thessaloniki, Greece, September 22-25, 2020*. Ed. by L. Cappellato, C. Eickhoff, N. Ferro, and A. Névéol. Vol. 2696. *CEUR Workshop Proceedings*. CEUR-WS.org.

Xiao, Y. and K. Cho. (2016). "Efficient character-level document classification by combining convolution and recurrent layers". *arXiv preprint arXiv:1602.00367*.

Yamaguchi, H. and K. Tanaka-Ishii. (2012). "Text segmentation by language using minimum description length". In: *Proceedings of the 50th Annual Meeting of the Association for Computational Linguistics (Volume 1: Long Papers)*. 969–978.

Yang, M., W. Zhao, J. Ye, Z. Lei, Z. Zhao, and S. Zhang. (2018). "Investigating Capsule Networks with Dynamic Routing for Text Classification". In: *Proceedings of the 2018 Conference on Empirical Methods in Natural Language Processing, Brussels, Belgium, October 31 - November 4, 2018*. Ed. by E. Riloff, D. Chiang, J. Hockenmaier, and J. Tsujii. Association for Computational Linguistics. 3110–3119. DOI: 10.18653/v1/d18-1350.

Yang, Z., Z. Dai, Y. Yang, J. Carbonell, R. R. Salakhutdinov, and Q. V. Le. (2019). "Xlnet: Generalized autoregressive pretraining for language understanding". *Advances in neural information processing systems.* 32.

Yang, Z., D. Yang, C. Dyer, X. He, A. Smola, and E. Hovy. (2016). "Hierarchical attention networks for document classification". In: *Proceedings of the 2016 conference of the North American chapter of the association for computational linguistics: human language technologies.* 1480–1489.

Yao, L., C. Mao, and Y. Luo. (2019). "Graph convolutional networks for text classification". In: *Proceedings of the AAAI conference on artificial intelligence.* Vol. 33. 7370–7377.

Ye, X. and G. Durrett. (2022). "The Unreliability of Explanations in Few-shot Prompting for Textual Reasoning". In: *Advances in Neural Information Processing Systems 35: Annual Conference on Neural Information Processing Systems 2022, NeurIPS 2022, New Orleans, LA, USA, November 28 - December 9, 2022.* Ed. by S. Koyejo, S. Mohamed, A. Agarwal, D. Belgrave, K. Cho, and A. Oh.

Yu, H., Z. Yang, K. Pelrine, J. F. Godbout, and R. Rabbany. (2023). "Open, Closed, or Small Language Models for Text Classification?" *arXiv preprint arXiv:2308.10092.* URL: https://arxiv.org/abs/2308.10092.

Zeng, D., K. Liu, S. Lai, G. Zhou, and J. Zhao. (2014). "Relation classification via convolutional deep neural network". In: *Proceedings of COLING 2014, the 25th International Conference on Computational Linguistics: Technical Papers.* 2335–2344.

Zeng, J., J. Li, Y. Song, C. Gao, M. R. Lyu, and I. King. (2018). "Topic Memory Networks for Short Text Classification". In: *Proceedings of the 2018 Conference on Empirical Methods in Natural Language Processing, Brussels, Belgium, October 31 - November 4, 2018.* Ed. by E. Riloff, D. Chiang, J. Hockenmaier, and J. Tsujii. Association for Computational Linguistics. 3120–3131. DOI: 10.18653/v1/d18-1351.

Zhang, T., M. Huang, and L. Zhao. (2018). "Learning structured representation for text classification via reinforcement learning". In: *Proceedings of the AAAI Conference on Artificial Intelligence.* Vol. 32.

Zhang, X., J. Zhao, and Y. LeCun. (2015). "Character-level convolutional networks for text classification". *Advances in neural information processing systems*. 28.

Zhang, Y., D. Song, P. Zhang, X. Li, and P. Wang. (2019). "A quantum-inspired sentiment representation model for twitter sentiment analysis". *Applied Intelligence*. 49: 3093–3108.

Zhang, Y. and B. Wallace. (2015). "A sensitivity analysis of (and practitioners' guide to) convolutional neural networks for sentence classification". *arXiv preprint arXiv:1510.03820*.

Zhou, P., Z. Qi, S. Zheng, J. Xu, H. Bao, and B. Xu. (2016). "Text Classification Improved by Integrating Bidirectional LSTM with Two-dimensional Max Pooling". In: *COLING 2016, 26th International Conference on Computational Linguistics, Proceedings of the Conference: Technical Papers, December 11-16, 2016, Osaka, Japan*. Ed. by N. Calzolari, Y. Matsumoto, and R. Prasad. ACL. 3485–3495.

Zhu, X., P. Sobihani, and H. Guo. (2015). "Long short-term memory over recursive structures". In: *International Conference on Machine Learning*. PMLR. 1604–1612.

Zhu, Y., X. Gao, W. Zhang, S. Liu, and Y. Zhang. (2018). "A bidirectional LSTM-CNN model with attention for aspect-level text classification". *Future Internet*. 10(12): 116.

Zong, C., R. Xia, and J. Zhang. (2021). "Data Annotation and Preprocessing". In: *Text Data Mining*. Singapore: Springer Singapore. 15–31. ISBN: 978-981-16-0100-2. DOI: 10.1007/978-981-16-0100-2_2. URL: https://doi.org/10.1007/978-981-16-0100-2_2.